\documentclass[10pt,twocolumn,letterpaper]{article}

\usepackage[pagenumbers]{cvpr} 

\usepackage{graphicx}
\usepackage{amsmath}
\usepackage{amssymb}
\usepackage{booktabs}
\usepackage{xcolor}
\usepackage{xargs}
\usepackage{rotating}
\usepackage{array}
\usepackage[ruled]{algorithm2e} 
\usepackage{algorithmic}
\usepackage{multirow}

\usepackage{caption}
\usepackage{float}
\usepackage{subcaption}
\usepackage{enumitem}

\SetAlFnt{\small}
\SetAlCapFnt{\small}
\SetAlCapNameFnt{\small}
\SetAlCapHSkip{0pt}

\usepackage[pagebackref,breaklinks,colorlinks]{hyperref}

\usepackage[capitalize]{cleveref}
\crefname{section}{Sec.}{Secs.}
\Crefname{section}{Section}{Sections}
\Crefname{table}{Table}{Tables}
\crefname{table}{Tab.}{Tabs.}

\def\cvprPaperID{*****} 
\def\confName{CVPR}
\def\confYear{2023}

\begin{document}

\title{Video Colorization with Pre-trained Text-to-Image Diffusion Models}

\author{
\begin{tabular}{ccccc}
Hanyuan Liu\textsuperscript{\rm 1} &
Minshan Xie\textsuperscript{\rm 1} &
Jinbo Xing\textsuperscript{\rm 1} &
Chengze Li\textsuperscript{\rm 2} &
Tien-Tsin Wong\textsuperscript{\rm 1}\thanks{Corresponding author.}
 \end{tabular}\\
\begin{tabular}{cc}
\textsuperscript{\rm 1} The Chinese University of Hong Kong & 
\textsuperscript{\rm 2} Caritas Institute of Higher Education\\
 \end{tabular}\\
\begin{tabular}{cc}
{\tt\small \{liuhy, msxie, jbxing, ttwong\}@cse.cuhk.edu.hk} &
{\tt\small czli@cihe.edu.hk}
 \end{tabular}
}
\addtocounter{figure}{-1}
\twocolumn[{%
\renewcommand\twocolumn[1][]{#1}%
\maketitle
\begin{center}
    \captionsetup{type=figure}
\def\myhighreswidth{0.17}
\def\myhighresoffset{0.00}
\centering
\renewcommand{\arraystretch}{1.0}
\setlength{\tabcolsep}{0pt}
\begin{tabular}[c]{cccccc}
        \begin{subfigure}[c]{\myhighreswidth\textwidth}\includegraphics[width=\textwidth]{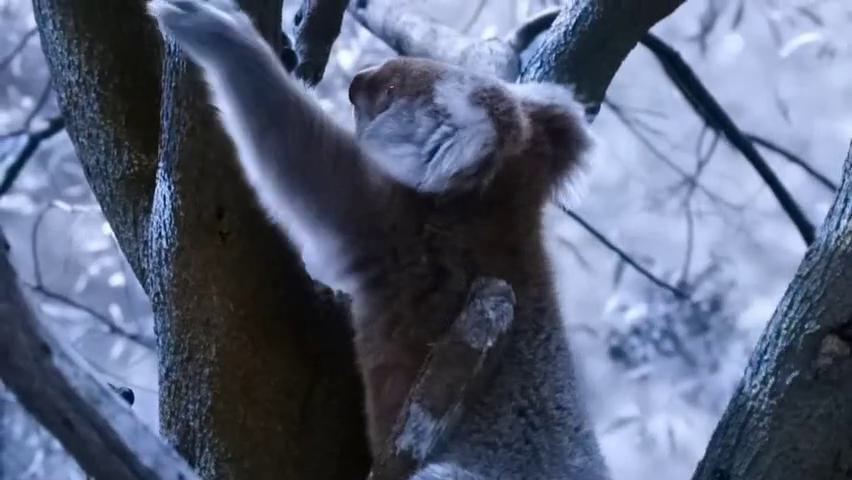}\end{subfigure}
        &
        \begin{subfigure}[c]{\myhighreswidth\textwidth}\includegraphics[width=\textwidth]{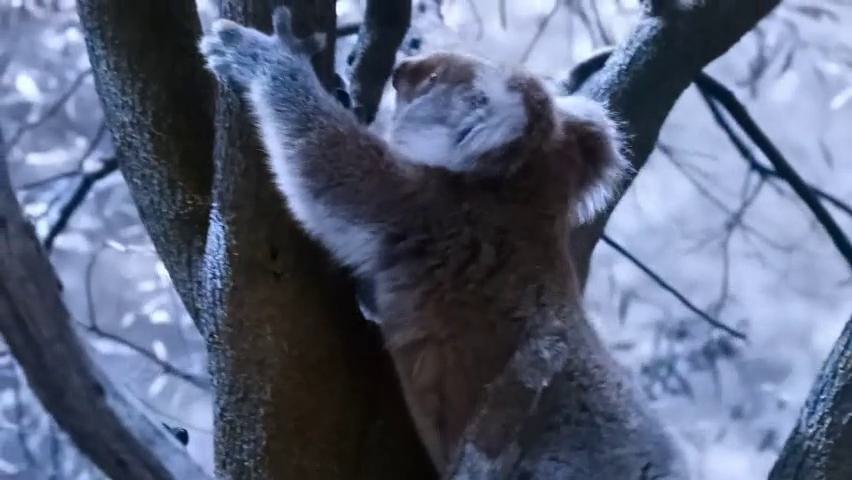}\end{subfigure}
        &
        \begin{subfigure}[c]{\myhighreswidth\textwidth}\includegraphics[width=\textwidth]{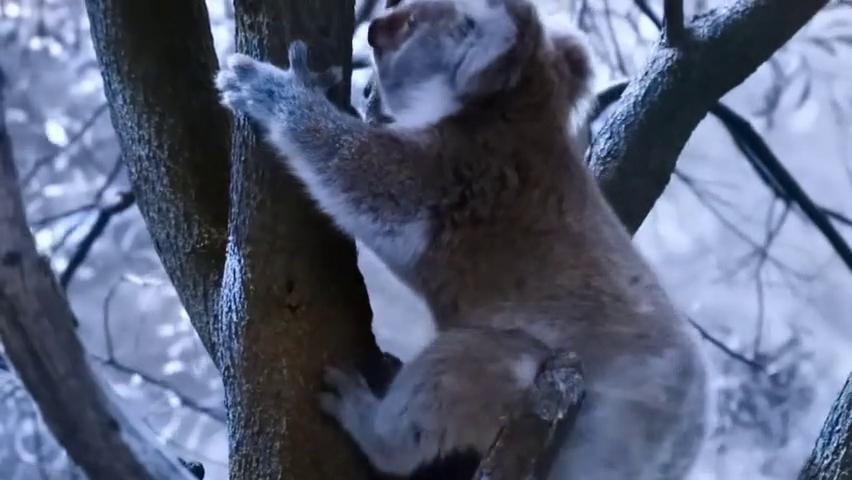}\end{subfigure}
        &
        \begin{subfigure}[c]{\myhighreswidth\textwidth}\includegraphics[width=\textwidth]{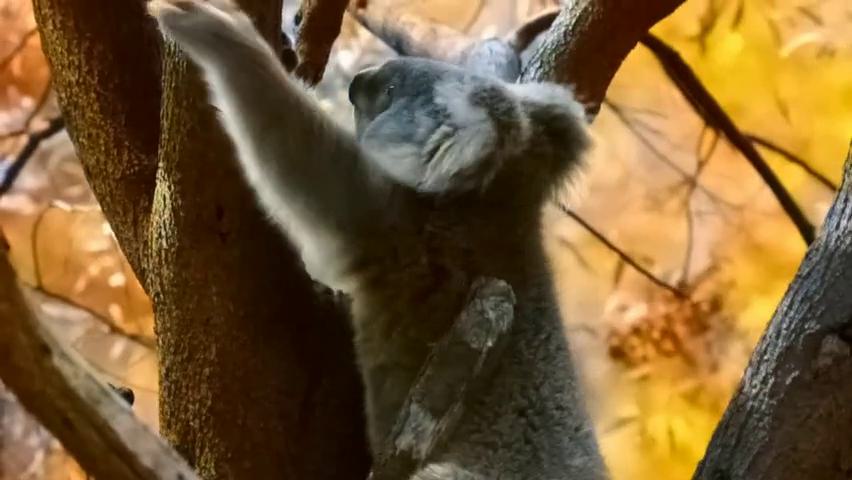}\end{subfigure}
        &
        \begin{subfigure}[c]{\myhighreswidth\textwidth}\includegraphics[width=\textwidth]{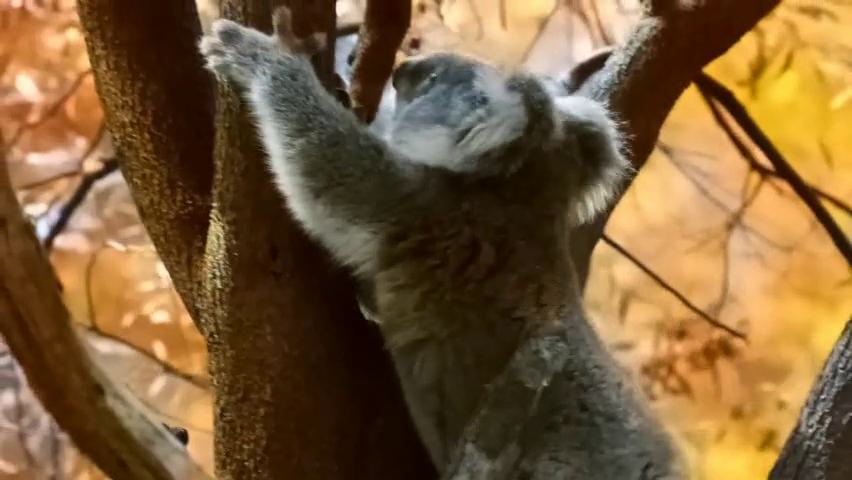}\end{subfigure}
        &
        \begin{subfigure}[c]{\myhighreswidth\textwidth}\includegraphics[width=\textwidth]{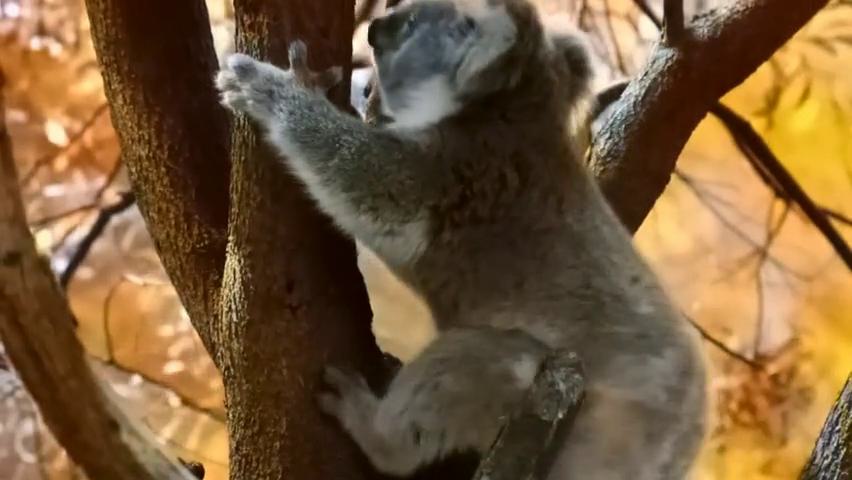}\end{subfigure}
        \\
        \multicolumn{3}{c}{\textit{"koala bear in the winter season"}} &         \multicolumn{3}{c}{\textit{"koala bear in the autumn season"}}\\
        \begin{subfigure}[c]{\myhighreswidth\textwidth}\includegraphics[width=\textwidth]{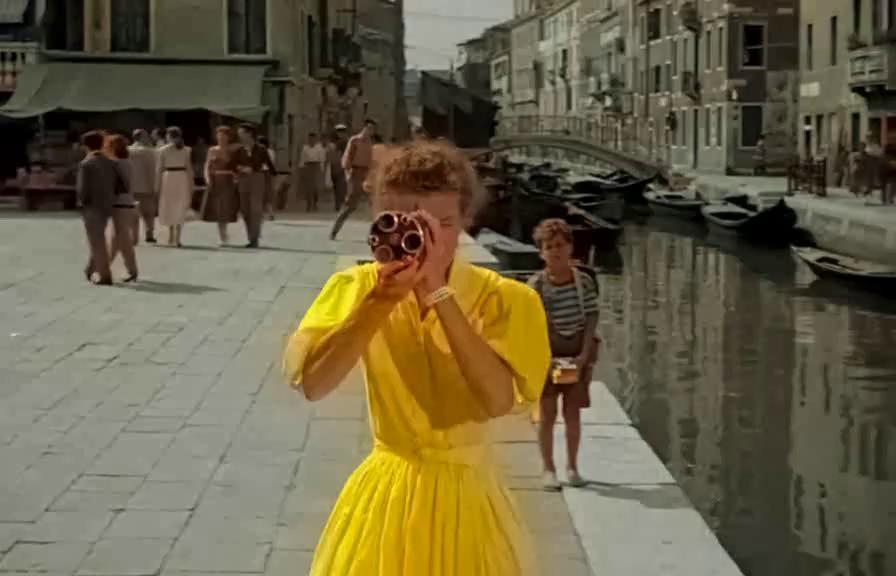}\end{subfigure}
        &
        \begin{subfigure}[c]{\myhighreswidth\textwidth}\includegraphics[width=\textwidth]{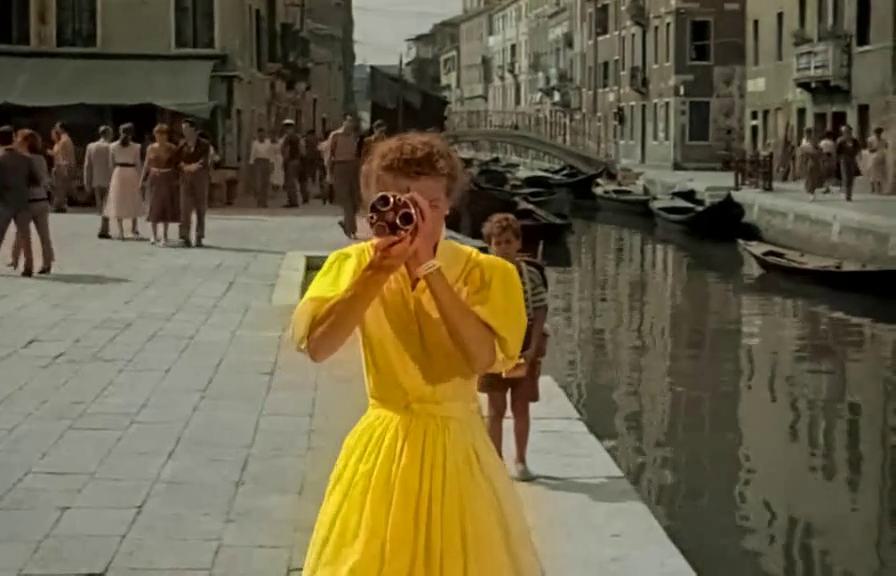}\end{subfigure}
        &
        \begin{subfigure}[c]{\myhighreswidth\textwidth}\includegraphics[width=\textwidth]{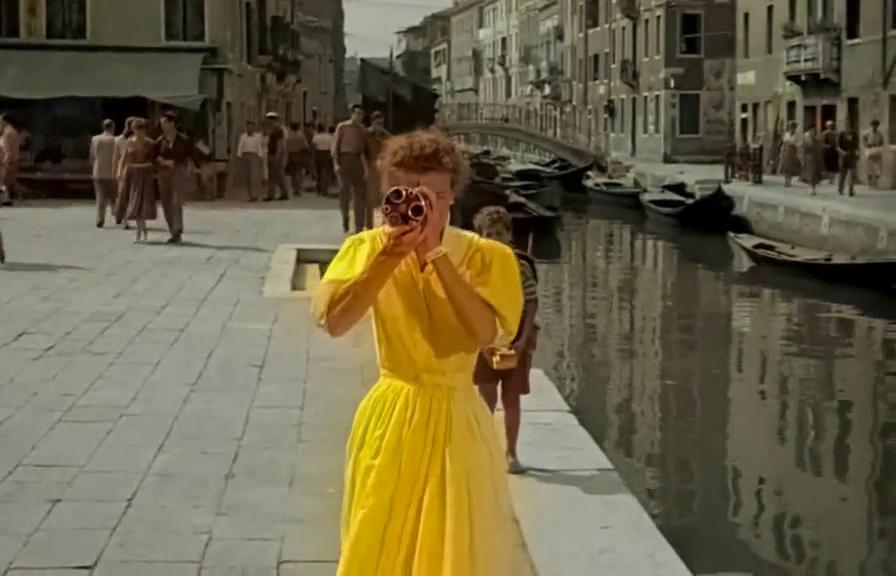}\end{subfigure}
        &
        \begin{subfigure}[c]{\myhighreswidth\textwidth}\includegraphics[width=\textwidth]{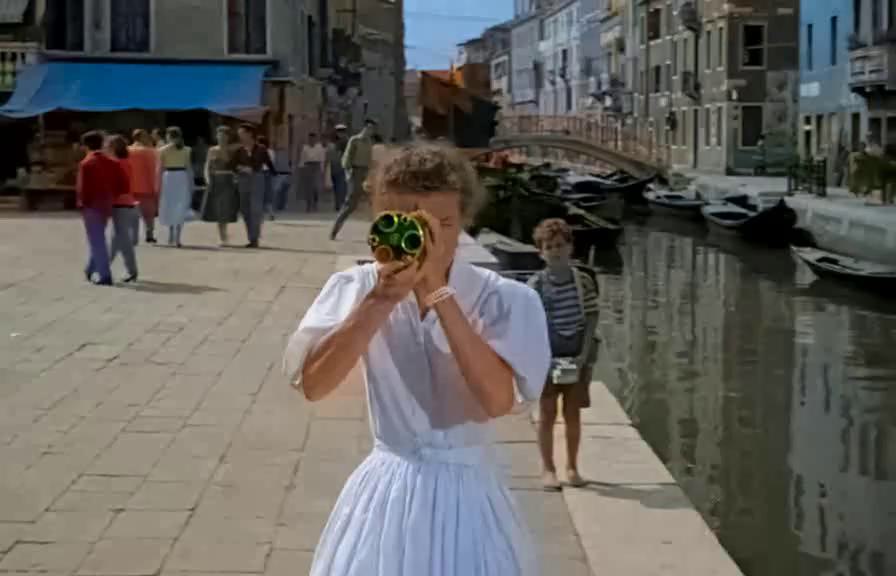}\end{subfigure}
        &
        \begin{subfigure}[c]{\myhighreswidth\textwidth}\includegraphics[width=\textwidth]{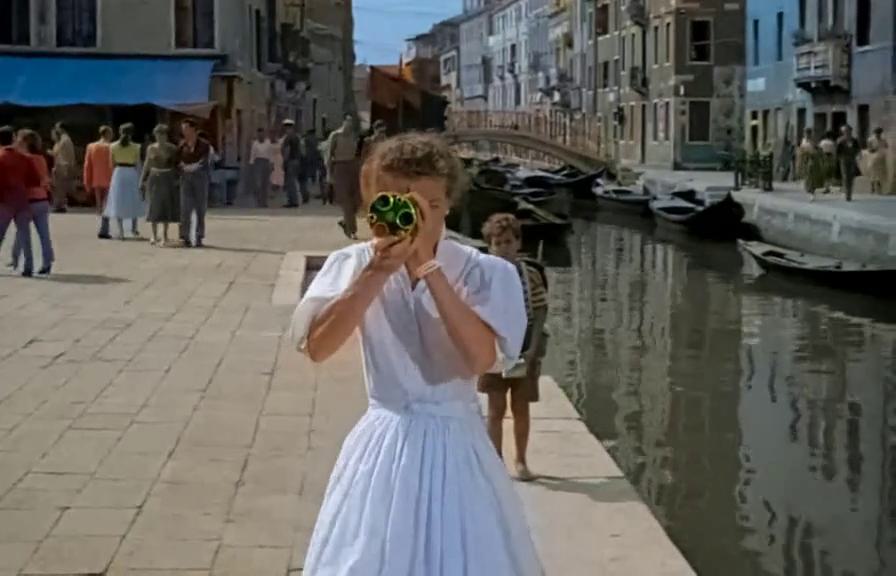}\end{subfigure}
        &
        \begin{subfigure}[c]{\myhighreswidth\textwidth}\includegraphics[width=\textwidth]{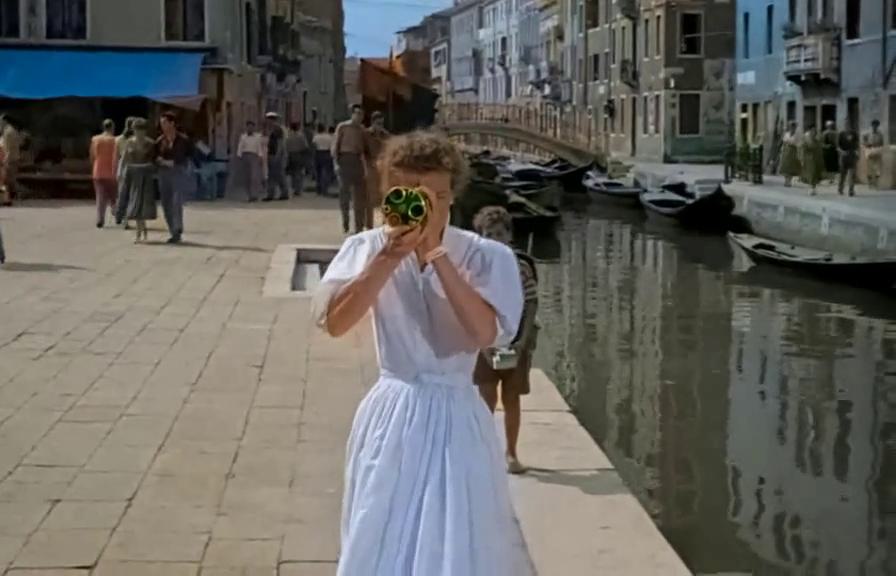}\end{subfigure}
        \\
        \multicolumn{3}{c}{\textit{"yellow skirt"}} &         \multicolumn{3}{c}{\textit{"vivid street view"}}\\
        \begin{subfigure}[c]{\myhighreswidth\textwidth}\includegraphics[width=\textwidth]{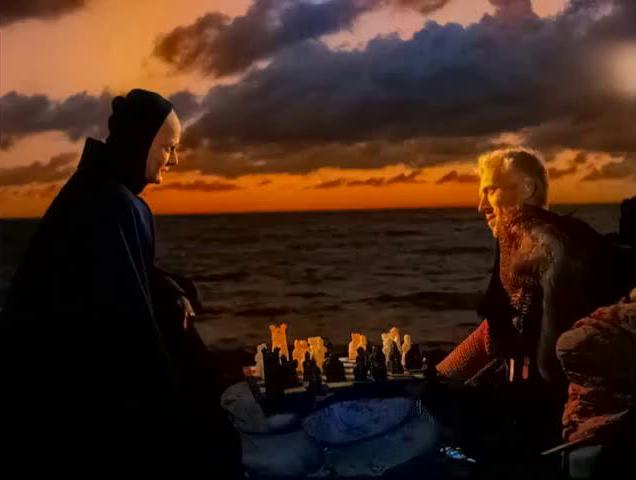}\end{subfigure}
        &
        \begin{subfigure}[c]{\myhighreswidth\textwidth}\includegraphics[width=\textwidth]{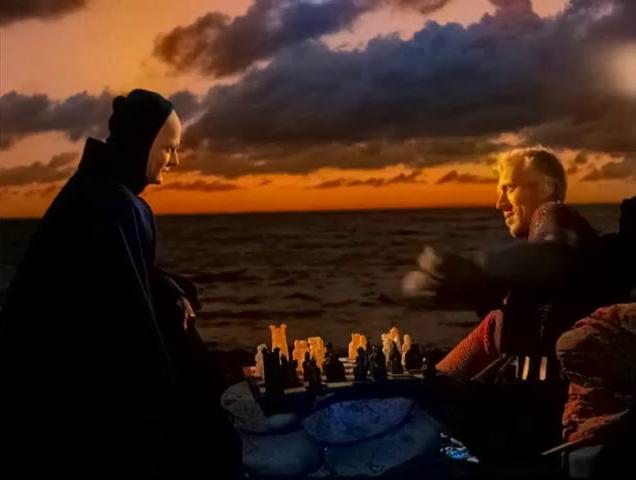}\end{subfigure}
        &
        \begin{subfigure}[c]{\myhighreswidth\textwidth}\includegraphics[width=\textwidth]{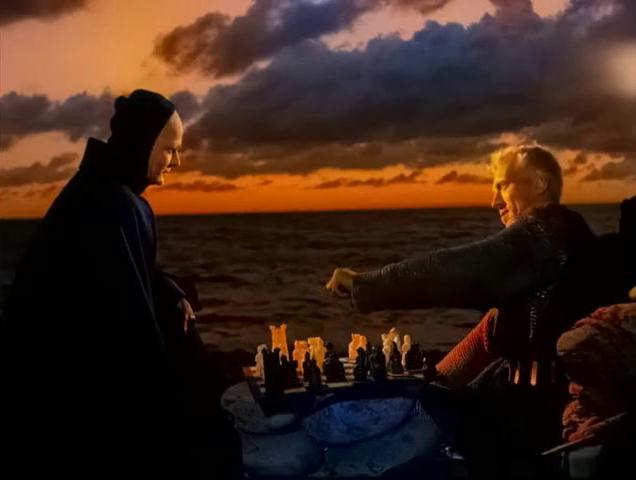}\end{subfigure}
        &
        \begin{subfigure}[c]{\myhighreswidth\textwidth}\includegraphics[width=\textwidth]{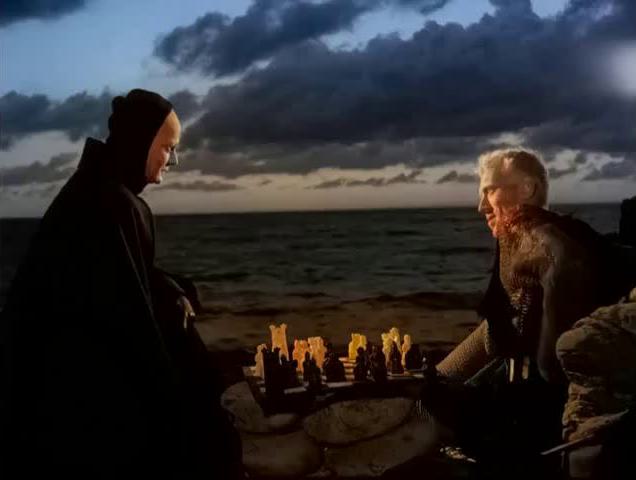}\end{subfigure}
        &
        \begin{subfigure}[c]{\myhighreswidth\textwidth}\includegraphics[width=\textwidth]{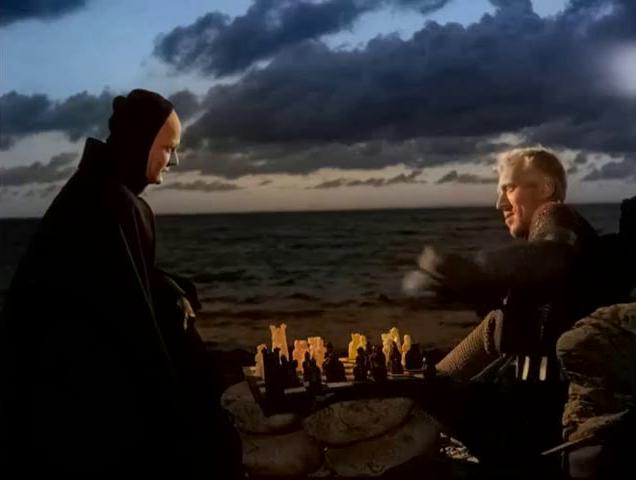}\end{subfigure}
        &
        \begin{subfigure}[c]{\myhighreswidth\textwidth}\includegraphics[width=\textwidth]{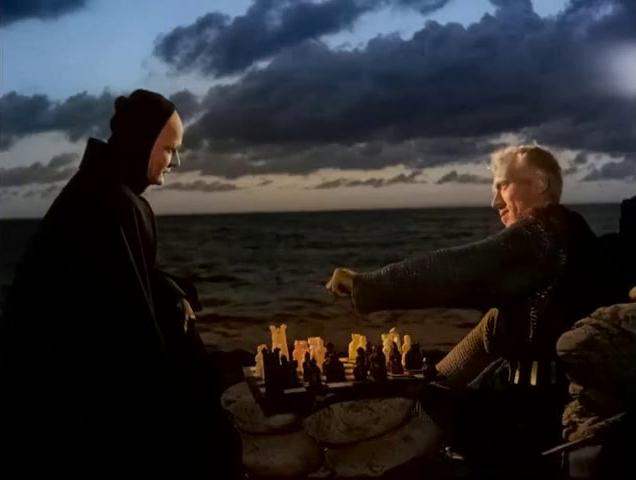}\end{subfigure}
        \\
        \multicolumn{3}{c}{\textit{"sunsetting"}} &         \multicolumn{3}{c}{\textit{"two men playing chess in front of the sea"}}\\

\end{tabular}
\renewcommand{\arraystretch}{1.0}

    \captionof{figure}{Our framework can generate high quality video colorization results based on textual descriptions.}
\end{center}%
}]

\begin{abstract}
  Video colorization is a challenging task that involves inferring plausible and temporally consistent colors for grayscale frames. In this paper, we present ColorDiffuser, an adaptation of a pre-trained text-to-image latent diffusion model for video colorization. With the proposed adapter-based approach, we repropose the pre-trained text-to-image model to accept input grayscale video frames, with the optional text description, for video colorization. To enhance the temporal coherence and maintain the vividness of colorization across frames, we propose two novel techniques: the {\em Color Propagation Attention} and {\em Alternated Sampling Strategy}. Color Propagation Attention enables the model to refine its colorization decision based on a reference latent frame, while Alternated Sampling Strategy captures spatiotemporal dependencies by using the next and previous adjacent latent frames alternatively as reference during the generative diffusion sampling steps. This encourages bidirectional color information propagation between adjacent video frames, leading to improved color consistency across frames. We conduct extensive experiments on benchmark datasets, and the results demonstrate the effectiveness of our proposed framework. The evaluations show that ColorDiffuser achieves state-of-the-art performance in video colorization, surpassing existing methods in terms of color fidelity, temporal consistency, and visual quality. 
\end{abstract}

\newcommand{\vardbtilde}[1]{\tilde{\raisebox{0pt}[0.85\height]{$\tilde{#1}$}}}
\newcommand{\defeq}{\coloneqq}
\newcommand{\shortcite}{\cite}
\newcommand{\grad}{\nabla}
\newcommand{\E}{\mathbb{E}}
\newcommand{\Var}{\mathrm{Var}}
\newcommand{\Cov}{\mathrm{Cov}}
\newcommand{\Ea}[1]{\E\left[#1\right]}
\newcommand{\Eb}[2]{\E_{#1}\!\left[#2\right]}
\newcommand{\Vara}[1]{\Var\left[#1\right]}
\newcommand{\Varb}[2]{\Var_{#1}\left[#2\right]}
\newcommand{\kl}[2]{D_{\mathrm{KL}}\!\left(#1 ~ \| ~ #2\right)}
\newcommand{\pdata}{{p_\mathrm{data}}}
\newcommand{\bA}{\mathbf{A}}
\newcommand{\bI}{\mathbf{I}}
\newcommand{\bJ}{\mathbf{J}}
\newcommand{\bH}{\mathbf{H}}
\newcommand{\bL}{\mathbf{L}}
\newcommand{\bM}{\mathbf{M}}
\newcommand{\bQ}{\mathbf{Q}}
\newcommand{\bR}{\mathbf{R}}
\newcommand{\bzero}{\mathbf{0}}
\newcommand{\bone}{\mathbf{1}}
\newcommand{\bb}{\mathbf{b}}
\newcommand{\bu}{\mathbf{u}}
\newcommand{\bv}{\mathbf{v}}
\newcommand{\bw}{\mathbf{w}}
\newcommand{\bx}{\mathbf{x}}
\newcommand{\by}{\mathbf{y}}
\newcommand{\bz}{\mathbf{z}}
\newcommand{\bxh}{\hat{\mathbf{x}}}
\newcommand{\btheta}{{\boldsymbol{\theta}}}
\newcommand{\bphi}{{\boldsymbol{\phi}}}
\newcommand{\bepsilon}{{\boldsymbol{\epsilon}}}
\newcommand{\bmu}{{\boldsymbol{\mu}}}
\newcommand{\bnu}{{\boldsymbol{\nu}}}
\newcommand{\bSigma}{{\boldsymbol{\Sigma}}}

\newcommand*{\approxident}{%
  \mathrel{\vcenter{\offinterlineskip
  \hbox{$\sim$}\vskip-.35ex\hbox{$\sim$}\vskip-.35ex\hbox{$\sim$}}}}
\newcommand{\RR}{\mathbb{R}}
\newcommand{\expec}{\mathbb{E}}
\newcommand{\prior}{\mathcal{N}(0,1)}

\newcommand{\PriorModel}{diffusion coordinator model}
\newcommand{\ColorizationDecoder}{video colorization VQVAE model}

\newcommand{\xpixel}{x_\text{pixel}}
\newcommand{\xpixelrec}{\tilde{x}_\text{pixel}}
\newcommand{\hpixel}{H}
\newcommand{\wpixel}{W}
\newcommand{\cpixel}{3}
\newcommand{\xrec}{\tilde{x}}

\newcommand{\latent}{z_0}
\newcommand{\hlatent}{h}
\newcommand{\wlatent}{w}
\newcommand{\clatent}{c}

\newcommand{\zt}[1]{z_{#1}}
\newcommand{\enc}{E}
\newcommand{\ppixel}{p_\text{pixel}}
\newcommand{\pzt}[1]{p_{\zt{#1}}}
\newcommand{\qzt}[1]{q_{\zt{#1}}}
\newcommand{\q}{q}
\newcommand{\R}{\mathbb{R}}

\newcommand{\LPIPS}{\text{LPIPS}}
\newcommand{\KL}{\mathbb{KL}}
\newcommand{\expect}{\mathbb{E}}
\newcommand{\pmodel}[1]{p^{#1}_{\theta}}
\newcommand{\pchain}{p_{\theta}}
\newcommand{\qchain}{q}
\newcommand{\qmodel}[1]{q_{#1}}

\newcommand{\qenc}{q_{\phi}}
\newcommand{\pdec}{p_{\phi}}
\newcommand{\dec}{G_{\phi}}

\newcommand{\disc}{D_{\psi}}

\newcommand{\lrec}{L_{rec}}
\newcommand{\ladv}{L_{adv}}
\newcommand{\lreg}{L_{reg}}
\newcommand{\lcomp}{L_{cm}}
\newcommand{\lsimple}{L_{DM}}
\newcommand{\lsimpleldm}{L_{LDM}}
\newcommand{\lsimplelcm}{L_{LDM}}
\newcommand{\lcolor}{L}

\newcommand{\pretrainedmodel}{\epsilon_{p}}
\newcommand{\model}{\epsilon_\theta}
\newcommand{\conditioner}{\tau_\theta}

\newcommand{\encoder}{\mathcal{E}}
\newcommand{\GrayEncoder}{{\mathcal{G}}}
\newcommand{\decoder}{\mathcal{D}}

\newcommand{\cond}{y}

\newcommand{\guider}{\mathcal{F}}

\def\headline#1{\hbox to \hsize{\hrulefill\quad\lower.3em\hbox{#1}\quad\hrulefill}}

\section{Introduction}

Colorization of grayscale images and videos enables the transformation of historical contents, enhancing visual aesthetics, and aiding in the delivery of the video contents. Various approaches have been proposed for this purpose, ranging from hand-crafted techniques~\cite{levin2004colorization,endo2016deepprop,qu2006manga,yatziv2006fast} to data-driven deep learning methods~\shortcite{endo2016deepprop,zhang2017real,he2019progressive, xu2020stylization}, and recent advances in generative models~\cite{gen-color-prior, kumar2021colorization}. While the existing methods manage to output high-quality color propagation on still images, video colorization remains challenging due to the simultaneous fulfillment over multiple visual aspects including temporal coherence, spatial and semantics consistency, as well as color richness and faithfulness over the contents (especially on colorizing artificial content). Naturally, training over large-scale dataset can achieve high-quality colorization, but the computational resource required is prohibitively large.

In this paper, instead of relying on large-scale data training solely for colorization purposes, we re-propose a pre-trained text-to-image diffusion model for our video colorization purpose. The text-to-image diffusion framework, originally designed for generating realistic images from textual descriptions, presents a promising potential for video colorization due to its ability to incorporate high-level semantic cues and the rich color variety learned from its large-scale dataset. To exploit the color information within such a pre-trained text-to-image model, we build an adapter-based model to redirect the pre-trained text-to-image diffusion model to generate color image that matches our input grayscale frame in terms of luminance and structure. In this way, the redirected text-to-image model not just synthesizes the colors, but also provides a diverse color variety due to its generative nature. Note that the pre-trained text-to-image model only generates static images, and hence, it does not guarantee temporal coherence across all video frames. In this paper, instead of relying on the unstable optical flow, typically used in existing methods, we propose a novel {\em color propagation attention module} to consistently colorize video frames over time. Such colorization is {\em not} a post-processing step, but a step {\em during the color inference}.

To evaluate the effectiveness of our proposed approach, we conduct extensive experiments on benchmark video colorization datasets and compare our results against several baseline methods. The experimental analysis demonstrates the superiority of our adapted text-to-image diffusion model in terms of color accuracy, temporal coherence, and controllability.

Our contributions can be summarized as follows:
\begin{itemize}[nolistsep]
    \item We propose a novel video colorization method, ColorDiffuser, that can colorize monochrome videos with diverse and consistent colors through multiple frames based on optional textual descriptions.
    \item We propose the color propagation attention and alternated sampling strategy, enabling video colorization with a conditional text-to-image diffusion model. 
\end{itemize}
\section{Related Work}
\subsection{Image Colorization}
\paragraph{\textbf{Semi-Automatic Colorization.}}
Early colorization methods focus on using local user hints, such as user scribbles~\cite{levin2004colorization}, and global hints, such as color palette~\cite{chang2015palette} or text~\cite{l-code}, to colorize images. 
These color hints are then propagated to the entire image and optimized based on hand-crafted low-level features~\cite{levin2004colorization,endo2016deepprop,qu2006manga,yatziv2006fast,cho2018text2colors}. 
Recently, Zhu et al.~\shortcite{zhang2017real} proposed a deep-learning-based method to propagate the sparse color hints by incorporating semantic information, and achieve real-time performance and remarkable quality. 
But, these methods usually require intensive manual annotations to generate plausible colorful images, which poses challenges to users.

Another category of work colorizes the grayscale image by transferring color information from a reference image with similar content. These methods transfer the color information to corresponding regions by matching low-level hand-crafted features~\cite{welsh2002transferring,tai2005local,bugeau2013variational,liu2008intrinsic,chia2011semantic}. But, these correspondence are not robust to objects with complex appearance as they do not capture high-level semantic information. To establish semantic correspondences, some researchers~\cite{he2019progressive, xu2020stylization,he2018deep} made use of deep semantic features extracted by a pre-trained neural network and generate plausible results. 
However, these methods may fail to provide robust and automatic colorization when content-related images are not available for reference. 
Instead of relying on user-provided references, our method exploits the color priors within pre-trained T2I model trained on large-scale datasets to generate diverse and vivid results. Furthermore, our method also accepts textual descriptions, a more convenient global hint, for generating controllable results.

\paragraph{\textbf{Fully Automatic Colorization.}}
With the advances of deep learning techniques, several automatic colorization methods attempted to learn the grayscale-to-color mapping from large-scale datasets~\cite{cheng2015deep,deshpande2015learning,isola2017image,zhang2016colorful}. These methods predict the color by considering both low and high-level semantics to achieve compelling colorized results~\cite{larsson2016learning,iizuka2016let,larsson2017pixcolor,zhao2020pixelated,su2020instance}. But, these methods lack the modeling of color ambiguity (especially on coloring artificial objects) and thus cannot generate diverse results. 
To alleviate this obstacle, diverse colorization methods have been proposed using generative models~\cite{vitoria2020chromagan,isola2017image,deshpande2015learning} and transformer models~\cite{kumar2021colorization,ji2022colorformer,color-token}. 
Some follow-up works achieve controllable and diverse colorization by predicting sparse guidance, such as color palette~\cite{wang2022palgan}, local color hints~\cite{disco-color}. 
However, these automatic methods are prone to produce visual artifacts like unnatural and incoherent colors when colorizing objects with color uncertainty. 

Recently, some approaches~\cite{pan2020exploiting,kim2022bigcolor} attempt to achieve diverse colorization by exploiting generative priors of pre-trained GANs~\cite{karras2019style,karras2020analyzing}. Wu et al.~\shortcite{wu2021towards} utilize GAN inversion techniques~\cite{zhu2016generative,zhu2020domain} to obtain a content-related image as a reference and then warp the color features into the grayscale image. 
However, the colorized results highly depend on the quality of images generated by GAN inversion. 
In contrast, we build a adapter-based model to redirect the powerful T2I diffusion model to generate diverse colorized results. In addition, it also supports conditional video colorization using different textual descriptions to specify the desired color composition and object colors.

\subsection{Video Colorization}
Video colorization considers temporal constraints on top of image colorization. Existing video colorization can be classified into three categories. 
The first is to perform post-processing to impose the temporal coherence in order to suppress the flickers of per-frame colorization with a general temporal filter~\cite{bonneel2015blind,lai2018learning}, but these works tend to wash out the colors. 
Another category on video colorization is mainly exemplar-guided, including propagating the user scribbles~\cite{yatziv2006fast,levin2004colorization}, attaching the colors from colorized frames~\cite{jampani2017video}, or given images~\cite{reinhard2001color} to the rest of frames. 
These approaches rely on the optical flow to propagate colors in videos either from scribbles or fully colored frames~\cite{levin2004colorization,yatziv2006fast}. However, inaccuracies in optical flow may lead to color artifacts with the accumulated errors over time.
Some other work uses one colored frame as an example and colorizes the following frames in sequence. While conventional methods rely on hand-crafted low-level features to find the temporal correspondence~\cite{jacob2009colorization,ben2015approximate,xia2016robust}, deep learning methods further improve colorization quality by taking advantage of semantics to learn the temporal propagation~\cite{jampani2017video,vondrick2018tracking,liu2018switchable}. 
Zhang et al.~\shortcite{zhang2019deep} colorized frames by considering both the matched features of the reference image and the propagated color from previous frames. 
However, the color propagation of these methods will be problematic when the scene disparity of examples and grayscale frames cannot be ignored. 
In contrast, our method does not require user to provide any reference, instead it exploits the color information hidden inside the pretrained T2I model trained on the large-scale data set. Our generative nature allows use to generate diverse colorization results for content without definitive colors. Conditional video colorization can also be achieved by feeding texts. The temporal coherence is achieved in our method via an innovative {\em frame-alternating cross attention} approach. 

\subsection{Diffusion Models}
\textit{Diffusion Models} \cite{diffusion-model} are probabilistic models designed to learn a data distribution $p(x)$ by gradually denoising a normally distributed variable. This denoising process or the generation process corresponds to learning the reverse process of a fixed Markov Chain of length $T$ \cite{ddpm, openai-diffusion, sr3-diffusion}. For image synthesis, the most successful models rely on a reweighted variant of the variational lower bound on $p(x)$ based on denoising score-matching \cite{score-sde}. These models can be interpreted as an equally weighted sequence of denoising autoencoders $\model(x_{t},t);\, t=1\dots T$, which are trained to predict a denoised variant of their input $x_t$, where $x_t$ is a noisy version of the input $x$. Instead of directly learning the target data distribution in the high-dimensional pixel space, \textit{Latent Diffusion Models} \cite{latent-diffusion} leverage perceptual compression with an autoencoder $\encoder$ and $\decoder$ for efficient low-dimensional representation features. The model learns $p(z|y)$, in which $\encoder(x)=z, \encoder(\decoder(x))\approx x$ for reconstruction:
\begin{equation}
\lsimplelcm = \expec_{\encoder(x), y, \epsilon \sim \mathcal{N}(0, 1), t }\Big[ \Vert \epsilon - \model(z_{t},t, y) \Vert_{2}^{2}\Big]
\label{eq:cond_loss}
\end{equation}
with $t$ uniformly sampled from $\{1, \dots, T\}$. The neural backbone $\model(\circ, t, \conditioner(y))$ is generally realized as a denoising U-Net~\cite{unet} with cross-attention conditioning mechanisms \cite{attention} to accept additional conditions. The research community also proposes instruction fine-tuning \cite{brooks2023instructpix2pix}, adapter-based modification \cite{t2iadapter,control-net} as extra conditioning mechanisms for a pre-trained text-to-image latent diffusion model. Recent works \cite{make-a-video,runway-gen-1} also extend the pre-trained text-to-image latent diffusion model for text-to-video generation. These works are trained on large scale video data, requiring extensive computational resources. 
\section{Method}
\subsection{Overview}
\begin{figure*}
   \centering
   \includegraphics[width=1.0\linewidth]{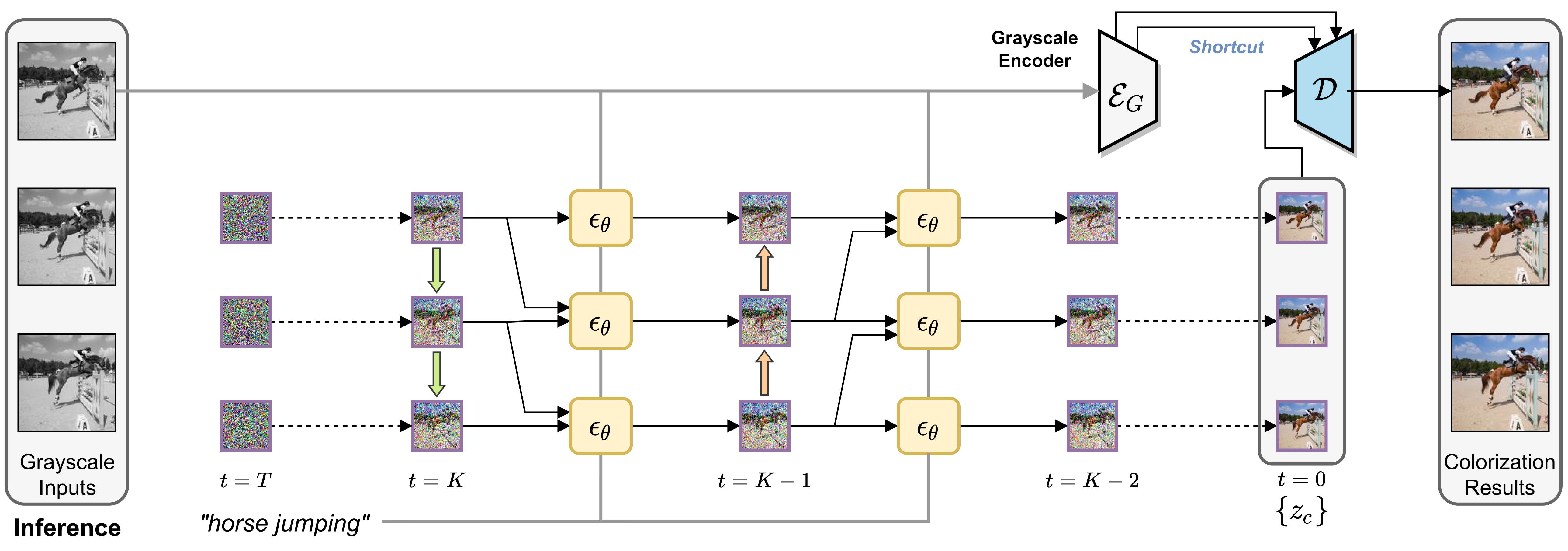}
\end{figure*}
\begin{figure}
   \centering
   \includegraphics[width=1.0\linewidth]{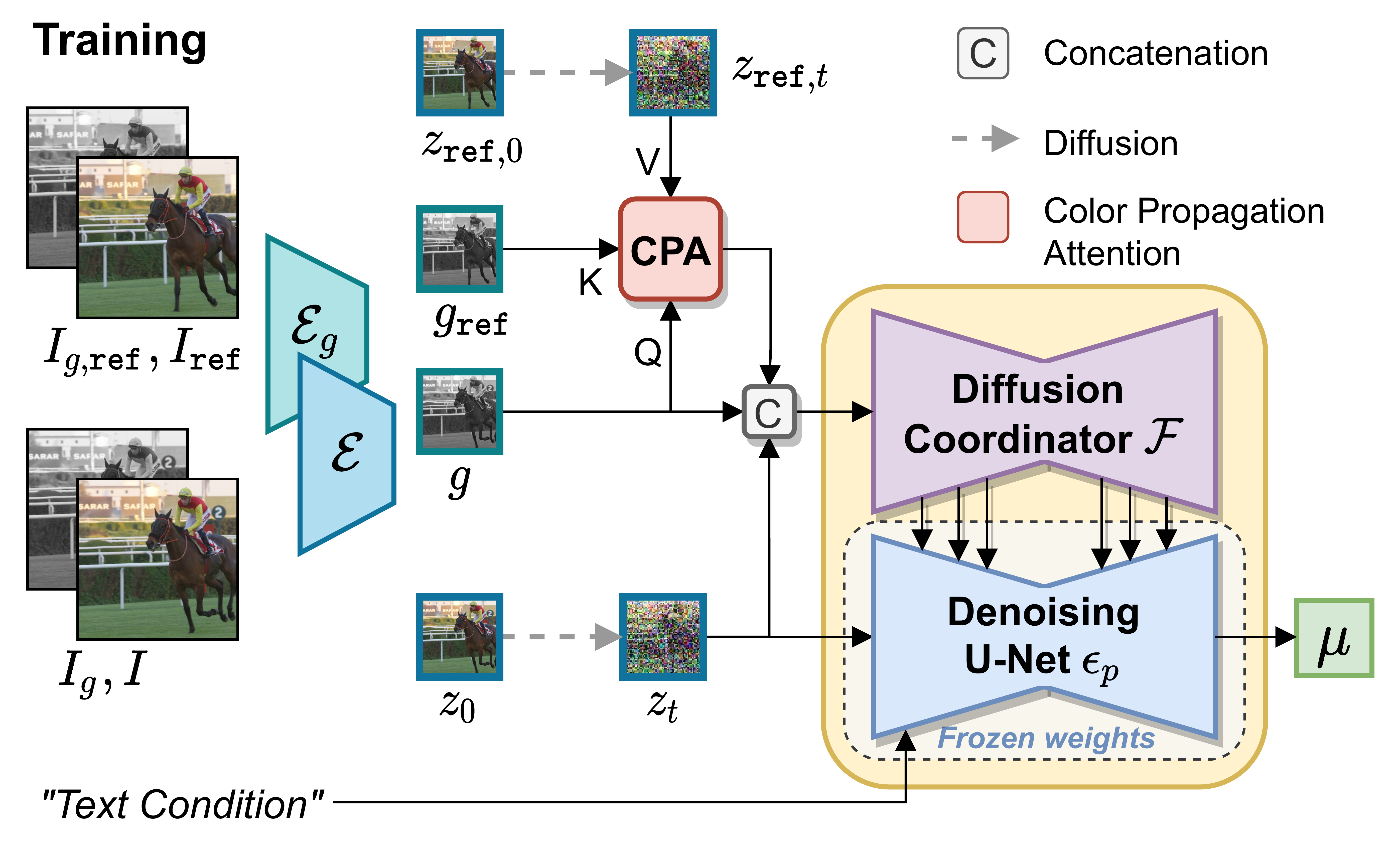}
      \caption{Overview. Our framework consists of three components: a \PriorModel, a pretrained stable diffusion model and a \ColorizationDecoder. The \PriorModel extends the pretrained T2I diffusion model into a conditional T2I diffsuion model. With the grayscale image $I_g$ and the optional text input $\texttt{text}$ and a reference color frame latent $z_{\texttt{ref}}$, the \PriorModel~guides the pretrained Stable Diffusion model to generate a ``colorized" latent $z_c$ through the generative diffusion process. The \ColorizationDecoder~then uses $z_c$ as the latent color prior and incorporates the grayscale information of $I_g$ to produce a pixel-aligned colorization result.}
   \label{fig:overview}
\end{figure}

In this work, we propose a framework to produce a high-quality and diverse colorization, optionally conditioned on user-input textual descriptions for grayscale videos. Specifically, the framework is built upon a pre-trained text-to-image (T2I) latent diffusion model, Stable Diffusion model, to exploit its capability in visual semantic understanding and image synthesis. As shown in Figure~\ref{fig:overview}, we extend the pre-trained Stable Diffusion model to a reference-based frame colorization model using Diffusion Coordinator $\mathcal{F}$. With the adapter-based mechanism \cite{control-net}, we obtain a conditional text-to-image diffusion model that leverages the power of the pre-trained Stable Diffusion model to render colors in the latent space $z_c$, according to the visual semantics of the grayscale input, the text input, and the reference color latent. 

To generate video outputs with coherent colorization, we introduce two key components: the Color Propagation Attention and the Alternated Sampling Strategy. For each frame in the input grayscale video, we perform a parallel sampling process. Each sampling step for a particular frame is conditioned on the latent information from the previous sampling step of an adjacent frame. Essentially, the Color Propagation Attention and the Alternated Sampling Strategy coordinate the reverse diffusion process and enable bidirectional propagation of color information between adjacent frames to ensure consistency in colorization over time. At last, we design the novel \ColorizationDecoder~$\decoder$ that fuses the compressed VQ color prior $\{z_c\}$ and $I_g$ to reconstruct the final colorization $\{I_c\}$ that precisely aligns with the structure and texture of $\{I_g\}$ to mitigate information loss of the latent diffusion. 

\subsection{Adapting text-to-image latent diffusion model towards video colorization}

The recent state-of-the-art text-to-image diffusion models \cite{latent-diffusion} ) have demonstrated successful learning of the conditional distribution, $p(z\vert{\texttt{text}})$, using large-scale text-image datasets \cite{laion-dataset}. 
These models exhibit a deep understanding of the intrinsic structure of natural images and the semantic relationship that exists between textual descriptions and corresponding images. Consequently, these models can serve as effective generative priors. Building upon this insight, we propose a framework that harnesses the capabilities of pre-trained text-to-image latent diffusion models as a color prior for video colorization. 

\subsubsection{Conditional generation}

\begin{figure}[h]
   \centering
   \includegraphics[width=1.0\linewidth]{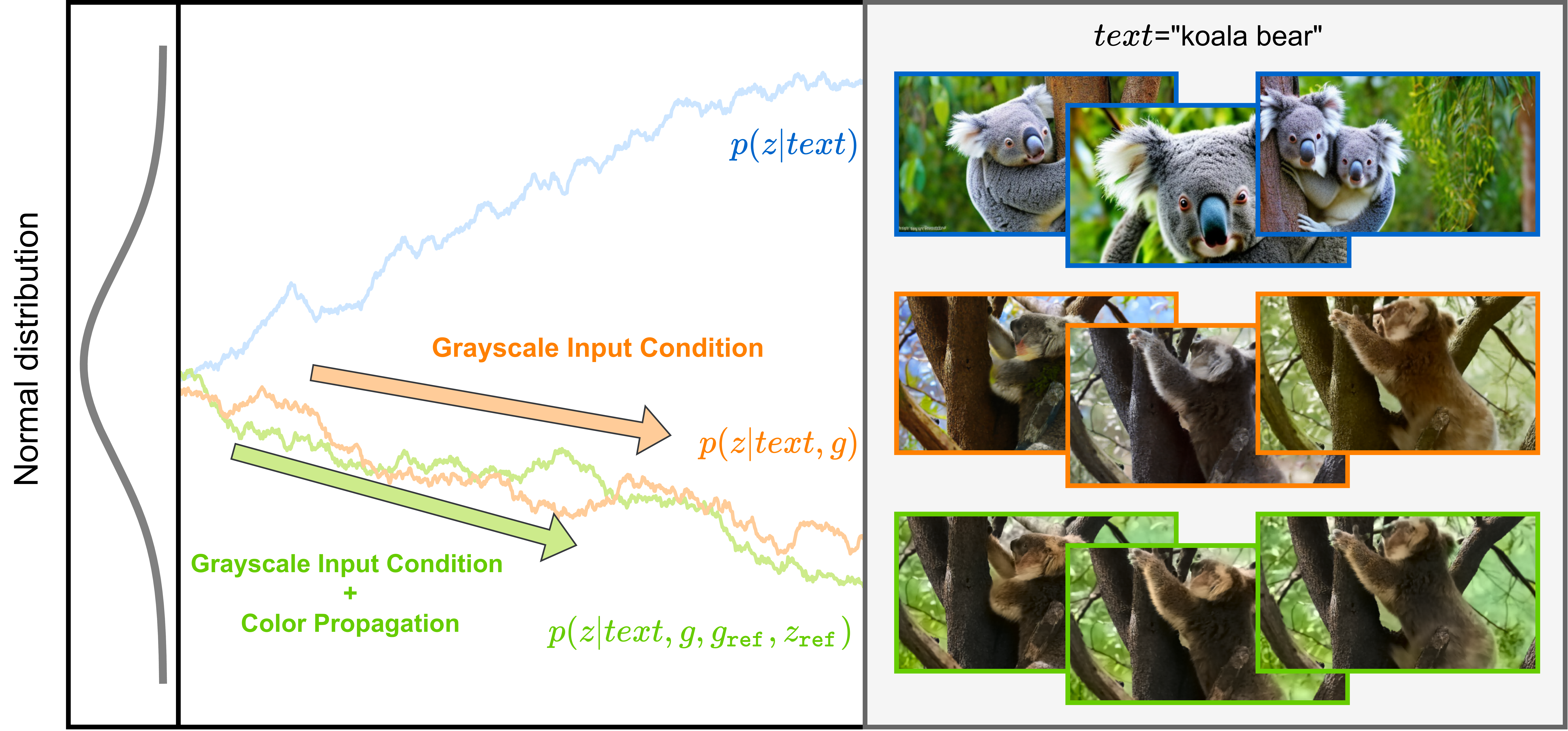}
      \caption{Given a sequence of grayscale frames, the diffusion coordinator redirects the SDE trajectories towards the target conditional data distribution $p(z\vert{\texttt{text}, g, g_{\texttt{ref}}, z_{\texttt{ref}}})$.}
   \label{fig:redirect-sde}
\end{figure}

Specifically, given a pre-trained text-to-image latent diffusion model $\pretrainedmodel$ for $p(z\vert{\texttt{text}})$, we propose to leverage adapter-based approach \cite{control-net,t2iadapter,piggyback-color} to extend $\pretrainedmodel$ to a conditional text-to-image latent diffusion model $\model$ for $p(z\vert{\texttt{text}, g, g_{\texttt{ref}}, z_{\texttt{ref}}})$. 
Instead of extending the pre-trained text-to-image diffusion model $\pretrainedmodel$ into a video diffusion model like \cite{make-a-video,runway-gen-1} which directly operates on an entire sequence of video frames and requires extensive computational resources for training and inference, we design to keep $\model$ as a conditional text-to-image diffusion model for synthesizing each video frame in a coordinated manner. 

We design a Diffusion Coordinator module $\mathcal{F}_{\theta}(\epsilon_{p}, g, g_{\texttt{ref}}, z_{\texttt{ref}})$ that can guide the generative diffusion process of a pre-trained diffusion model towards a certain subspace, in which $\texttt{rgb2gray}(\decoder(z_c)) \approx I_g$ and $\decoder(z_c)$ should have the same color distribution of $\decoder(z_{\texttt{ref}})$, as shown in Figure~\ref{fig:redirect-sde}. 
The diffusion coordinator shares a similar architecture of the denoising U-Net $\pretrainedmodel(\circ, t, \texttt{text})$ except that it contains additional encoders for the grayscale input images and the reference color latents. As shown in Figure~\ref{fig:overview}, we take the feature map $F_\mathcal{F}$ from each downsampling/upsampling layer of Diffusion Coordinator $\mathcal{F}$ and inject them into the corresponding layer of $\pretrainedmodel$:
\begin{equation}
F_{{\pretrainedmodel}_i}' = \texttt{Conv}_{1x1}(\texttt{Concat}(F_{{\pretrainedmodel}_i}, \texttt{Conv}_{1x1}(F_{\mathcal{F}_i})))
\end{equation}
The Diffusion Coordinator takes the grayscale images and a reference color latent as input and learns to adjust the inner features of $\pretrainedmodel$ to generate a modified score, which leads the generation result conforming to the grayscale image and reference color latent. In this way, we can consider $\mathcal{F}_{\theta}(\pretrainedmodel, g, g_{\texttt{ref}}, z_{\texttt{ref}})$ as a large diffusion model $\model$ conditioned on $(g, g_{\texttt{ref}}, z_{\texttt{ref}})$.

\subsubsection{Color Propagation Attention}

The Diffusion Coordinator $\mathcal{F}$ takes the grayscale input frame, grayscale version of the reference frame, and the latent version of reference frame as inputs and guides the pretrained text-to-image latent diffusion $\pretrainedmodel$ to generate the colorized latent $z_c$ for the grayscale input frame $g$. 

We propose the {\em color propagation attention} module to propagate the color information from the reference frame to the current frame generation. This {\em reference} frame can be a {\em neighboring frame in time}. Here, we modify the cross-attention mechanism \cite{vaswani2017attention} to achieve this propagation. 
As illustrated in Figure~\ref{fig:overview}, instead of deriving the key $K$ and the value $V$ from the same feature map as in typical usages, we project the features of the {\em current grayscale} frame, the features of {\em the reference grayscale} frame, and the features of the {\em reference color} frame to  $Q$,  $K$, and  $V$, respectively. 
With this design, our module effectively pairs features from the reference grayscale and reference color frames, by exploiting the locality of latent representations for the reference grayscale and the reference color frames, i.e. the fact that the features at the same spatial location in $K$ and $V$ are likely representing the same patch from the original pixel space. Then, this cross-attention mechanism
works like a ``table lookup'' operation for each query feature $Q$ (from the current grayscale frame), into the ``table'' of $K$ (from the reference grayscale frame) and $V$ (from reference color frame).  Under such design, our module not just determines the latent color features for the current grayscale frame, but also simultaneously propagates the latent color features across frames in the temporal domain:
\begin{equation}
   {F}_{c} = \texttt{Attention}(Q,K,V) = \texttt{softmax}(\frac{QK^T}{\sqrt{d_k}})V
\end{equation}

\subsubsection{Alternating Sampling Strategy}

With the color propagation attention module, our diffusion model can generate colorized latent based on the neighboring reference frame, e.g. the previous frame. If we simply use the framework in an auto-regressive manner, the temporal coherence is only limited in the forward-time direction. Consequently, the color propagation may degrade through the long-term generation, especially when new content appears in the subsequent frames. To mitigate this issue, we propose to exploit the sampling steps of the diffusion sampling process to ensure long-term information propagation. 

Recall that we train $\model$ conditioned on noisy version of reference latents, we do not need to restrict the reference frame to be the previous frame throughout the {\em whole} diffusion process. Instead, during the generative diffusion sampling process, $\model$ can take the intermediate sampling results as the reference latents from either the {\em previous} or the {\em next} frames in an {\em alternating} fashion. 
Practically, for a certain frame $(z_{t,i}, g_i)$, we use $\model(z_{t,i}, t, \texttt{text}, g_i, g_{i-1}, z_{t,i-1})$ at odd steps and use $\model(z_{t,i}, t, \texttt{text}, g_i, g_{i+1}, z_{t,i+1})$ at even steps as the score function. By doing so, we gradually propagate the color information in a {\em bidirectional} manner, through the diffusion sampling steps and thereby enable the video colorization even we only have a text-to-image model. We show the detailed algorithm in Algorithm~\ref{alg:inference}. 

\subsubsection{Loss Function} Treating $\model$ as a conditional diffusion model, the loss function can be written as follows:
\begin{equation}
\resizebox{.9\hsize}{!}{$
    \lcolor = \expec_{\encoder(x), \epsilon \sim \mathcal{N}(0, 1),  t}\Big[ \Vert \epsilon - \model(z_{t} ,t, \texttt{text}, g, g_{\texttt{ref}}, z_{t,\texttt{ref}}) \Vert_{2}^{2}\Big] \, $}
    \label{eq:colorloss}
\end{equation}
To reduce the learning difficulty and fully utilize diffusion prior, we freeze the weights of the original diffusion model $\pretrainedmodel$ during the training. 

\renewcommand{\algorithmicrequire}{\textbf{Input:}}
\begin{algorithm}[hbt!]
   \caption{Inference}\label{alg:inference}
   \begin{algorithmic}[1]
   \REQUIRE grayscale frame features $\{g_i\}_{N}$
       \FOR{$i=1, \dotsc, N$}
         \STATE  $z_{T,i} \sim \mathcal{N}(\bzero, \bI)$
       \ENDFOR
       \FOR{$t=T, \dotsc, 1$}
        \FOR{$i=1, \dotsc, N$}
        \STATE $\mu = \model(z_{t,i}, t, \texttt{text}, g_i, g_{i-1}, z_{t,i-1})$ if $t$ is even, else $\mu = \model(z_{t,i}, t, \texttt{text}, g_i, g_{i+1}, z_{t,i+1})$
        \STATE $z_{t-1,i} = \texttt{update}(\mu, z_{t,i})$
       \ENDFOR
       \ENDFOR
       \STATE \textbf{return} $\{\decoder(z_0)\}_N$
   \end{algorithmic}
   \end{algorithm}

\subsection{Video Colorization VQVAE Decoder}

With our conditional latent diffusion model $\model$, we manage to achieve video colorization in the latent space. However, the visual quality and temporal consistency may not be guaranteed in the pixel space as we model the conditional data distribution in a compressed vector quantized space \cite{vq-vae, vq-gan}. The output of the original latent decoder $\decoder$ may exhibit significant distortion and artifacts because the precise structure and texture information are only approximated during the vector quantization process. Consequently, the generated output $\{z_c\}$ is only an estimation for the color frames $\tilde{I_c}=\decoder(z_c)$, where the intensity of $\tilde{I_c}$ closely resembles $I_g$ through the relationship $\texttt{rgb2gray}(\tilde{I_c}) \approx I_g$. This approximation cannot be directly used as colorization results, as it fails to align the structure and texture accurately with the original grayscale input frames $\{I_g\}$.

Unfortunately, we are constrained to the latent space of the pre-trained latent diffusion model in order to exploit its capabilities for image generation. Although the resulting $\tilde{I_c}$ falls short in terms of visual quality,  it is noteworthy that the generated latent representation ${z_c}$ has already produced commendable colorization outcomes for the input grayscale frames, both at a global and local level, as depicted in Figure~\ref{fig:vae-out}. 

In order to address the aforementioned artifacts and generate high-quality video colorization results, we propose a solution that combines the generated latents $z_c$ with pixel-level details extracted from the input grayscale frames. This combination enables precise alignment of the colorization output with the original frame structure. Our approach introduces the \ColorizationDecoder, which consists of a replica of the original latent diffusion VQVAE decoder and a grayscale encoder module $\GrayEncoder$. Instead of relying solely on the decoder $\decoder$ to reconstruct the compressed structure and texture details, we incorporate the grayscale encoder to shortcut and inject grayscale features from $I_g$ into the decoder. This integration provides essential cues to facilitate accurate reconstruction at the pixel level during the decoding process. To further enhance the temporal coherence, we extend $\decoder$ by incorporating pseudo 3D convolution \cite{pseudo-3d, make-a-video}. This is achieved by adding an additional Dirac-initialized temporal convolutional layer at the end of each residual block. Through this extension, we obtain a video VQVAE decoder capable of sharing information between a sequence of frames, thus improving the temporal consistency of the colorization results.

Motivated by the concept of shortcut learning \cite{shortcut-learning}, the grayscale encoder $\GrayEncoder$ closely resembles the original VQVAE encoder, with the exception that we extract the intermediate features $\{F_{\GrayEncoder_i}\}=\GrayEncoder(I_g)$ at each downsampling layer as the encoder's output. These processed grayscale features are then added to the corresponding feature maps in the upsampling layers of the VQVAE decoder $\decoder$.
\begin{equation}
F_{\decoder_i}' = F_{\decoder_i} + \texttt{Conv}_{1x1}(F_{\GrayEncoder_i})
\end{equation}
To promote the effective utilization of grayscale features by the \ColorizationDecoder, we adopt a strategy where we freeze the original weights of the VQVAE decoder and solely optimize $\GrayEncoder$ and newly added temporal convolutional layers during the training. In this way, we consider $\texttt{Conv}_{1x1}(F_{\GrayEncoder_i})$ as learnable residuals \cite{resnet}. Consequently, the architecture of the \ColorizationDecoder ~closely resembles a UNet \cite{unet} but is disconnected in the middle layers.

\noindent\textbf{Loss function.} To enhance the quality of reconstruction, we use a combination of $L_1$ loss, perceptual loss \cite{perceptual-loss}, and discriminator loss \cite{vq-gan}. The loss function is defined as:
\begin{equation}
L=\mathcal{L}_1+\lambda_{p}\mathcal{L}_p+\lambda_{d}\mathcal{L}_d
\end{equation}
where the coefficient $\lambda_{p}$ is set to 0.1 and $\lambda_{d}$ is an adaptive weight determined using the strategy proposed in \cite{vq-gan}. 

\section{Experiments}

\begin{table*}[!ht]
\caption{Quantitative results on the evaluation datasets from different methods. The best items and second best items are highlighted in \textbf{bold} and \underline{underline} respectively. Ours\textsuperscript{\textdagger} uses generated captions from BLIP captioning model \cite{li2022blip} as additional inputs. Ours\textsuperscript{\textdaggerdbl} uses prompt \textit{"high quality color photo"} and negative prompt \textit{"color bleeding"} as additional inputs.}
\label{tab:quantitative_eval}

\setlength\tabcolsep{1pt}
\centering
\resizebox{1\textwidth}{!}{
  \begin{tabular}{c|cccccc|cccccc}
    \toprule
    \multirow{2}*{Method} & \multicolumn{6}{c|}{DAVIS30 (medium frame length)}&\multicolumn{6}{c}{Videvo20 (long frame length)}  \\
    ~ & FID $\downarrow$ &Colorfulness $\uparrow$ & PSNR $\uparrow$ & SSIM $\uparrow$ & LPIPS $\downarrow$ & CDC $\downarrow$  & FID $\downarrow$ &Colorfulness $\uparrow$ & PSNR $\uparrow$ & SSIM $\uparrow$ & LPIPS $\downarrow$ & CDC $\downarrow$ \\
    \midrule
    \midrule
AutoColor    &           { 83.05 } &           { 14.14 } & \underline{ 24.41 } & { 0.915 } & { 0.264 } & { 0.003734 } & { 76.28 } & { 13.23 } & \textbf{ 25.90 } & { 0.925 } & { 0.277 } & { 0.001668 }\\
Deoldify     &           { 76.21 } &           { 25.47 } & { 23.99 } & { 0.885 } & { 0.306 } & { 0.004901 } & { 66.89 } & \underline{ 22.05 } & { 24.31 } & { 0.895 } & { 0.325 } & { 0.003134 }\\
DeepExemplar &           { 77.26 } & \underline{ 28.82 } & { 21.78 } & { 0.846 } & { 0.325 } & { 0.004006 } & { 76.63 } & \textbf{ 32.44 } & { 20.63 } & { 0.831 } & { 0.348 } & { 0.002011 }\\
DeepRemaster &           { 97.54 } &           { 25.66 } & { 21.95 } & { 0.848 } & { 0.354 } & { 0.005098 } & { 86.23 } & { 28.72 } & { 21.88 } & { 0.856 } & { 0.358 } & { 0.003607 }\\
TCVC         &           { 74.94 } &           { 21.72 } & \textbf{ 25.17 } & \underline{ 0.921 } & \underline{ 0.239 } & \underline{ 0.003649 } & { 76.02 } & { 18.89 } & { 25.18 } & \underline{ 0.929 } & \underline{ 0.273 } & \underline{ 0.001629 }\\
VCGAN        & \underline{ 70.29 } &           { 15.89 } & { 23.90 } & { 0.910 } & { 0.247 } & { 0.005303 } & \textbf{ 63.83 } & { 14.90 } & { 24.67 } & { 0.919 } & { 0.276 } & { 0.002998 }\\
Ours         &    \textbf{ 69.51 } &    \textbf{ 29.13 } & { 23.73 } & \textbf{ 0.939 } & \textbf{ 0.213 } & \textbf{ 0.003607 } & \underline{ 66.11 } & { 20.73 } & \underline{ 25.27 } & \textbf{ 0.951 } & \textbf{ 0.205 } & \textbf{ 0.001591 }\\
    \midrule
    \midrule
    Ours\textsuperscript{\textdagger}     & 63.06 & 32.00 & 23.12 & 0.943 & 0.219 & 0.003963 & 61.69 & 33.44 & 23.23 & 0.922 & 0.235 & 0.002029\\
    Ours\textsuperscript{\textdaggerdbl}  & 70.98 & 36.03 & 22.33 & 0.931 & 0.233 & 0.003819 & 63.50 & 34.37 & 23.24 & 0.938 & 0.226 & 0.001974 \\
    \bottomrule
  \end{tabular}
  }

\end{table*}

\subsection{Implementation and Training}

\noindent\textbf{Implementation details}. We implement our framework in JAX and Flax based on the open-source codebase of diffusers \cite{diffusers}. We use the pre-trained \texttt{miniSD} \cite{mini-sd}, a fine-tuned Stable Diffusion v1.4 model \cite{latent-diffusion} on image size 256x256 as the latent diffusion prior inside our framework. 

\noindent\textbf{Data preparation}. To train the \PriorModel, we utilize the WebVid-2M video-text dataset \cite{webvid}. For each video in the dataset, we sample pairs of adjacent frames at random frame rates to create the training data. In order to enable unconditional colorization, we randomly replace the original caption with empty string at a chance of 50 percent and randomly set the reference frame to zeros at a chance of 10 percent. We use spatial resolution of $256\times256$ for training. For frames with an original caption, we first resize the frame to $454\times256$ and perform a center crop. For frames with a replaced caption, we apply a random crop. Additionally, the \ColorizationDecoder~is trained on the WebVid-2M video-text dataset \cite{webvid}.

\noindent\textbf{Training setup}. For both models, we utilize the AdamW optimizer \cite{adamw} with a learning rate of $lr=0.00001$. The \PriorModel~is trained with a per-device batch size of 8 for 200,000 steps on a TPUv3-32. On the other hand, the \ColorizationDecoder~is trained with a per-device batch size of 6 for 100,000 steps on a TPUv3-32.
To reduce the learning burden, we employ a two-stage training strategy for the \PriorModel: in the first stage, we use the unnoised version of reference frame latent for training; in the second stage, we use the noisy version of reference frame latent for training. 

\subsection{Evaluations}

\noindent\textbf{Dataset.} In this section, we assess the performance of our framework along with existing representative works using two widely-used evaluation benchmarks: the DAVIS dataset \cite{davis} and the Videvo dataset \cite{videvo}. We follow the evaluation protocols established by previous studies~\cite{tcvc-color, vcgan} and conduct evaluations on specific subsets of these datasets. Specifically, we evaluate all methods on the validation split of the DAVIS dataset and the test split of the Videvo dataset. 

\noindent\textbf{Evaluation metrics.} Evaluation of video colorization involves assessing perceptual realism, color vividness, and temporal consistency. To quantitatively measure the perceptual realism of the colorized videos, we employ the Fréchet Inception Score (FID)~\cite{heusel2017gans}. FID measures the distribution similarity between the predicted colors and the ground truth, thus providing an indication of perceptual realism.
For evaluating color vividness, we utilize the Colorfulness metric~\cite{hasler2003measuring}, which closely aligns with human visual perception. This metric enables us to assess the richness and vibrancy of colors in the colorized videos.
To evaluate temporal consistency, we utilize the Color Distribution Consistency (CDC) index \cite{tcvc-color}. This index is derived from the Jensen-Shannon divergence of the color distribution between consecutive frames, providing a measure of how consistent the color transitions are across frames in the colorized videos.
In addition to the aforementioned metrics, we also report evaluation results using PSNR, SSIM, and LPIPS~\cite{zhang2018unreasonable}. These metrics provide further insights into the perceptual quality of the colorized videos. 

\begin{figure*}[hbtp]
    \centering
    \captionsetup[subfigure]{labelformat=empty}
\def\myhighreswidth{0.12}
\def\myhighresoffset{-0.002}
\centering
\renewcommand{\arraystretch}{0.1}
\setlength{\tabcolsep}{0pt}
\begin{tabular}[c]{ccccccccc}
\begin{sideways}{\tiny Grayscale Input}\end{sideways} & \begin{subfigure}[b]{\myhighreswidth\textwidth}\includegraphics[width=\textwidth]{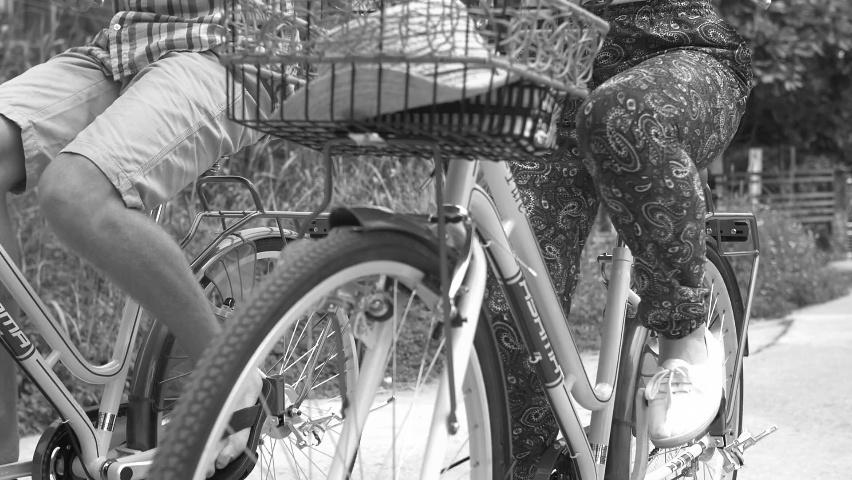}\end{subfigure} & \begin{subfigure}[b]{\myhighreswidth\textwidth}\includegraphics[width=\textwidth]{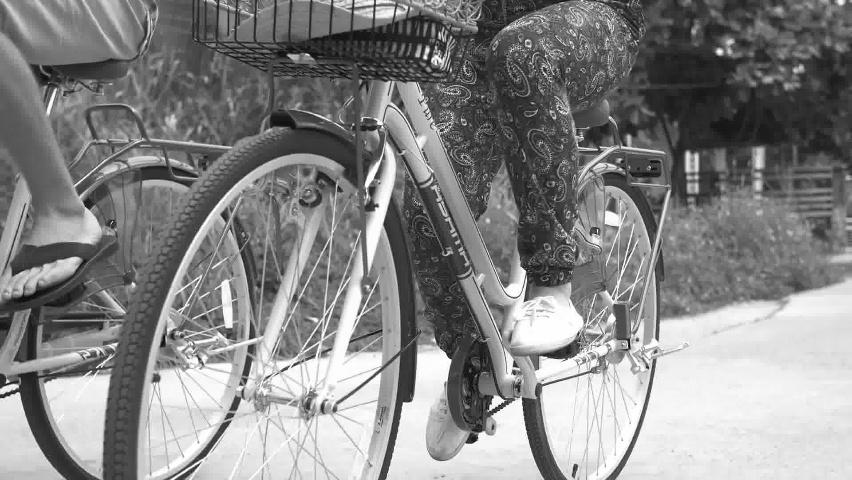}\end{subfigure} & \begin{subfigure}[b]{\myhighreswidth\textwidth}\includegraphics[width=\textwidth]{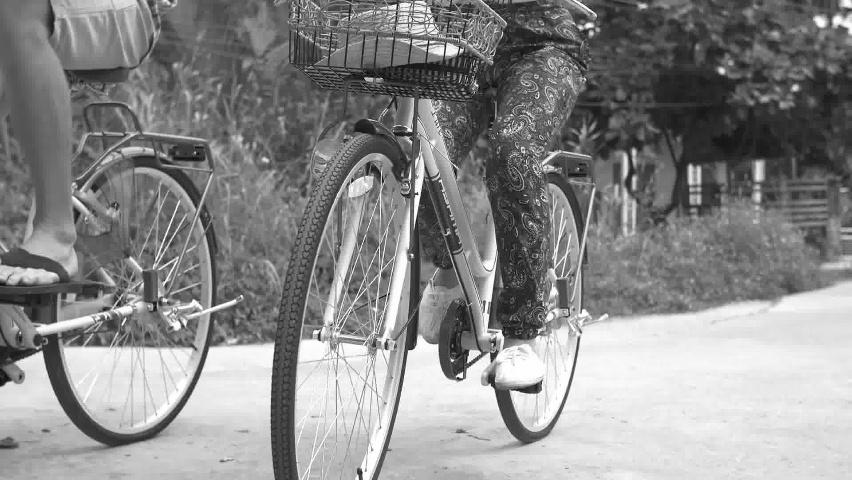}\end{subfigure} & \begin{subfigure}[b]{\myhighreswidth\textwidth}\includegraphics[width=\textwidth]{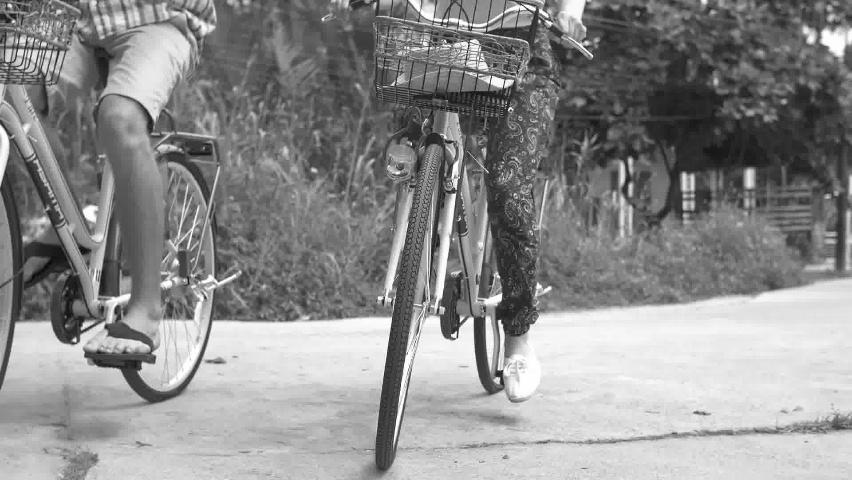}\end{subfigure} & \begin{subfigure}[b]{\myhighreswidth\textwidth}\includegraphics[width=\textwidth]{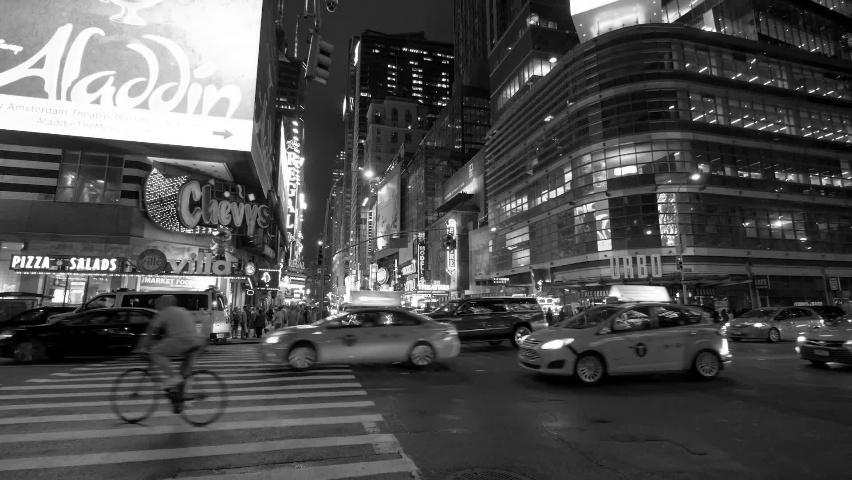}\end{subfigure} & \begin{subfigure}[b]{\myhighreswidth\textwidth}\includegraphics[width=\textwidth]{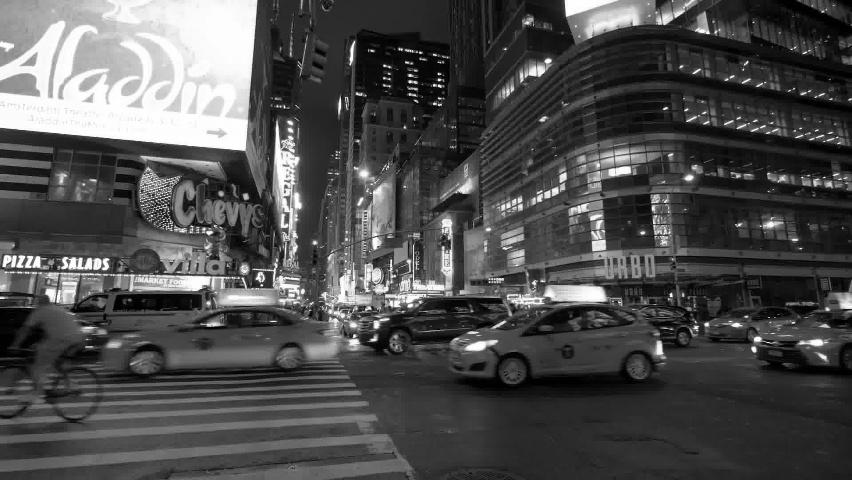}\end{subfigure} & \begin{subfigure}[b]{\myhighreswidth\textwidth}\includegraphics[width=\textwidth]{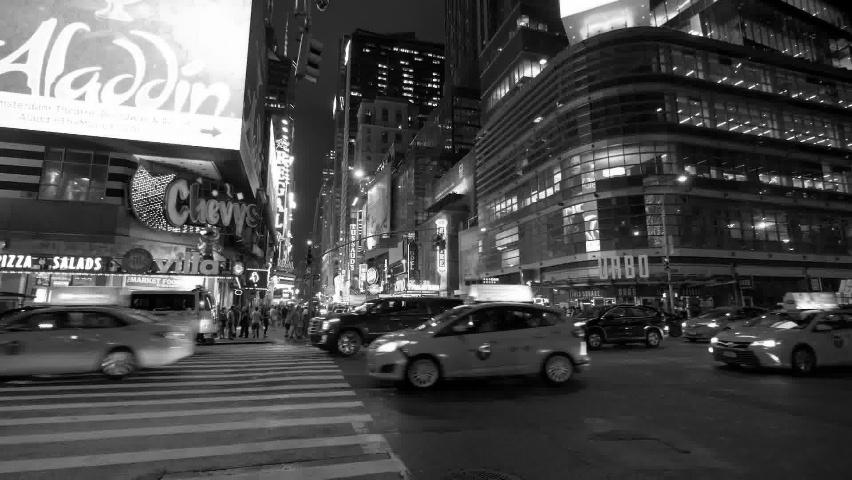}\end{subfigure} & \begin{subfigure}[b]{\myhighreswidth\textwidth}\includegraphics[width=\textwidth]{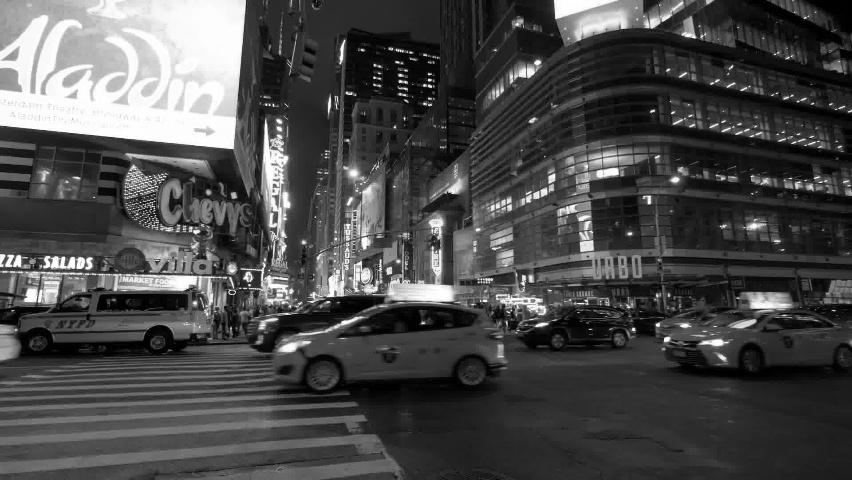}\end{subfigure}\\
\begin{sideways}{\tiny Ground Truth}\end{sideways} & \begin{subfigure}[b]{\myhighreswidth\textwidth}\includegraphics[width=\textwidth]{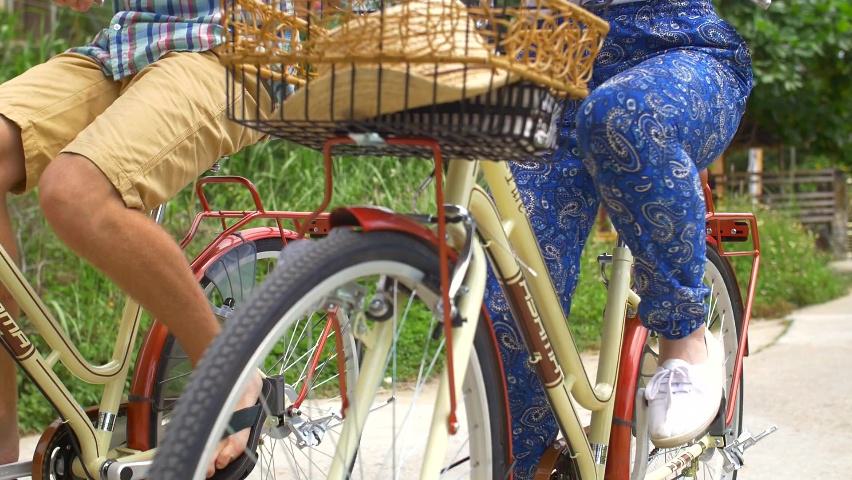}\end{subfigure} & \begin{subfigure}[b]{\myhighreswidth\textwidth}\includegraphics[width=\textwidth]{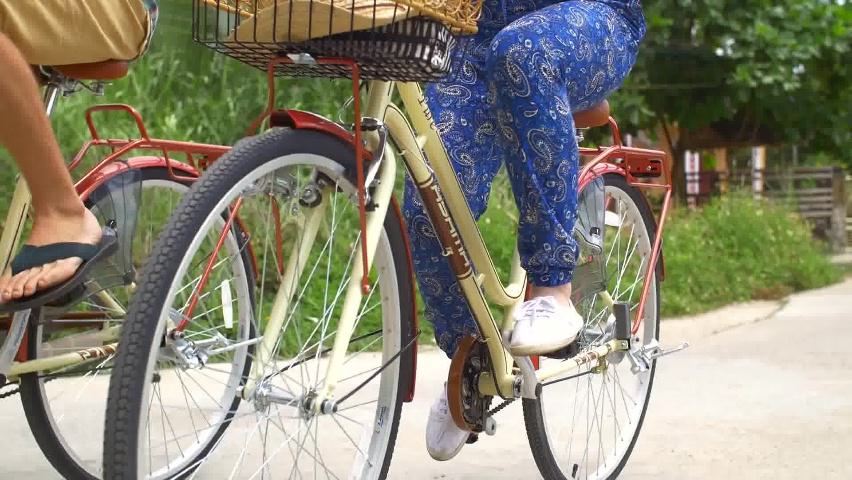}\end{subfigure} & \begin{subfigure}[b]{\myhighreswidth\textwidth}\includegraphics[width=\textwidth]{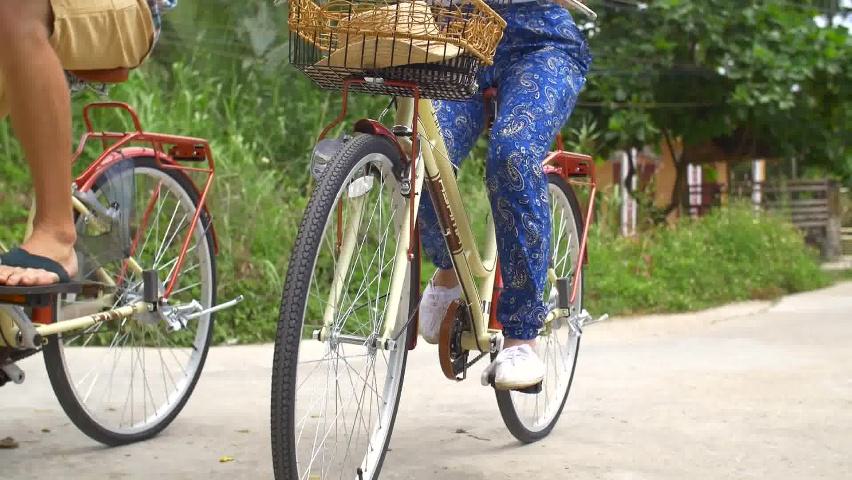}\end{subfigure} & \begin{subfigure}[b]{\myhighreswidth\textwidth}\includegraphics[width=\textwidth]{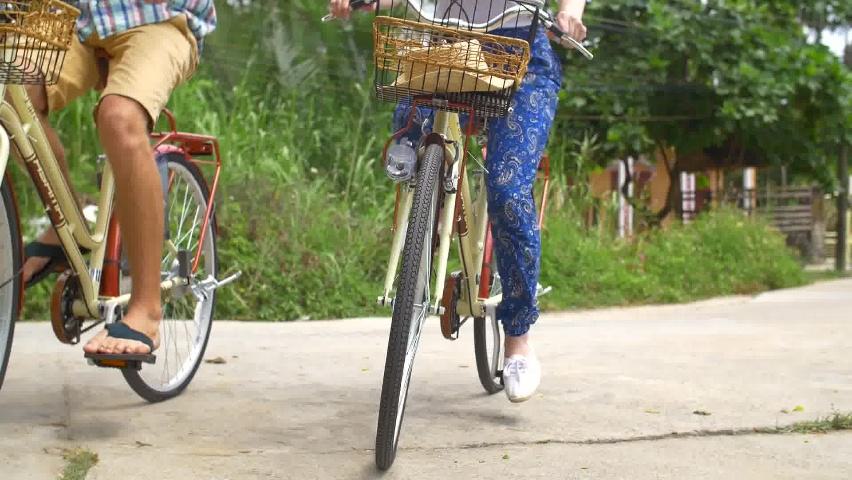}\end{subfigure} & \begin{subfigure}[b]{\myhighreswidth\textwidth}\includegraphics[width=\textwidth]{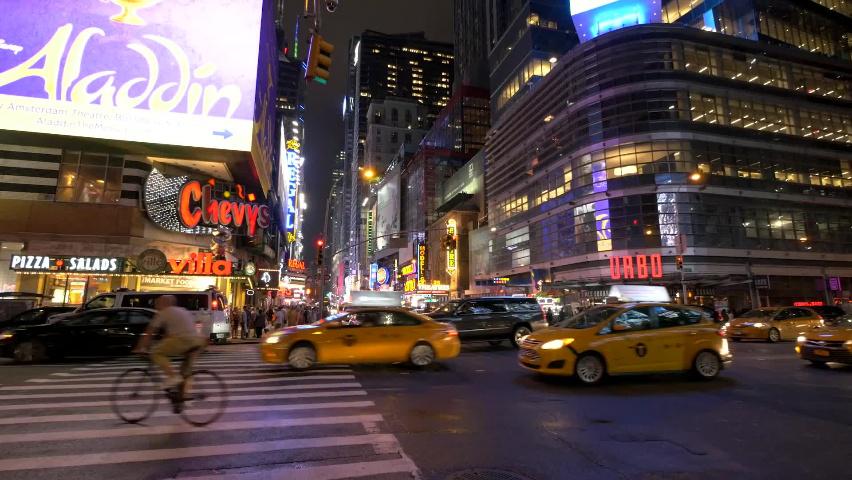}\end{subfigure} & \begin{subfigure}[b]{\myhighreswidth\textwidth}\includegraphics[width=\textwidth]{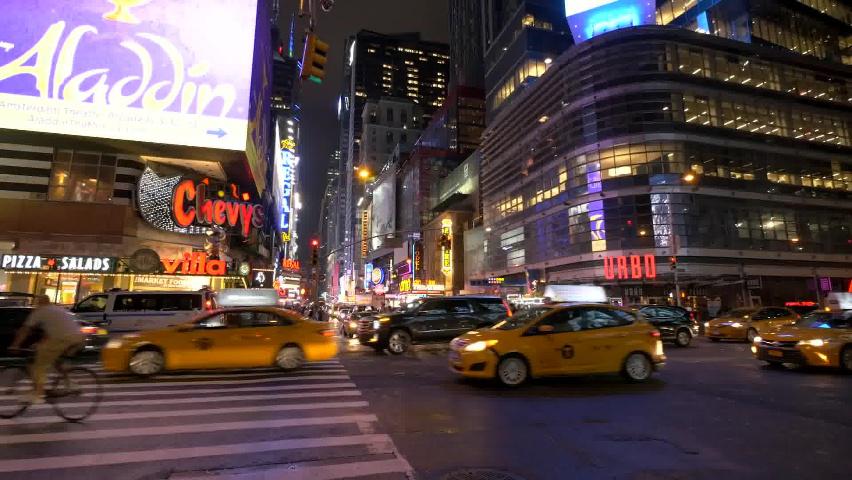}\end{subfigure} & \begin{subfigure}[b]{\myhighreswidth\textwidth}\includegraphics[width=\textwidth]{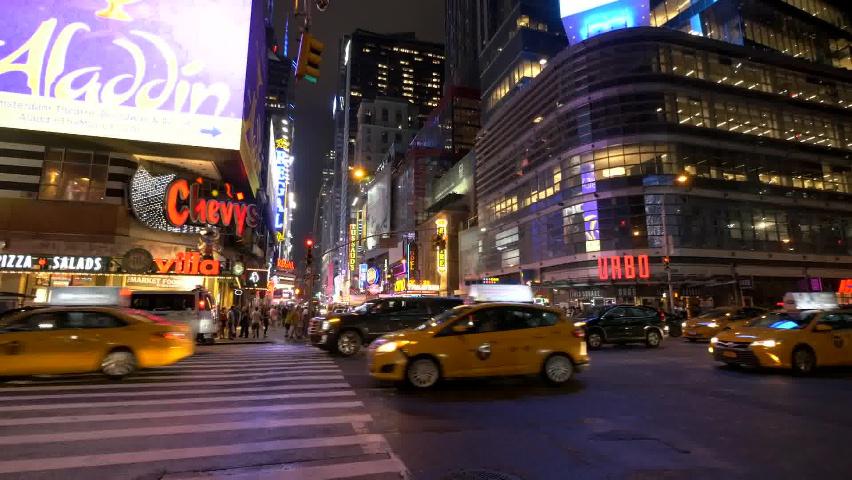}\end{subfigure} & \begin{subfigure}[b]{\myhighreswidth\textwidth}\includegraphics[width=\textwidth]{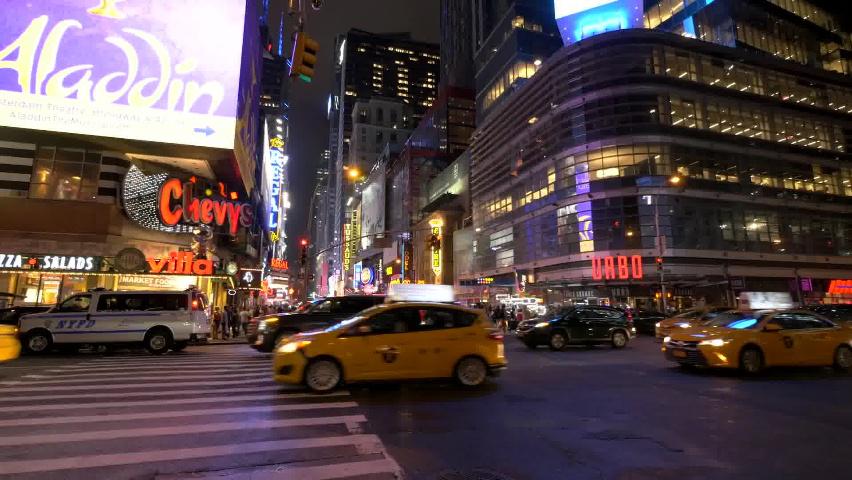}\end{subfigure}\\
\begin{sideways}{\tiny AutoColor}\end{sideways} & \begin{subfigure}[b]{\myhighreswidth\textwidth}\includegraphics[width=\textwidth]{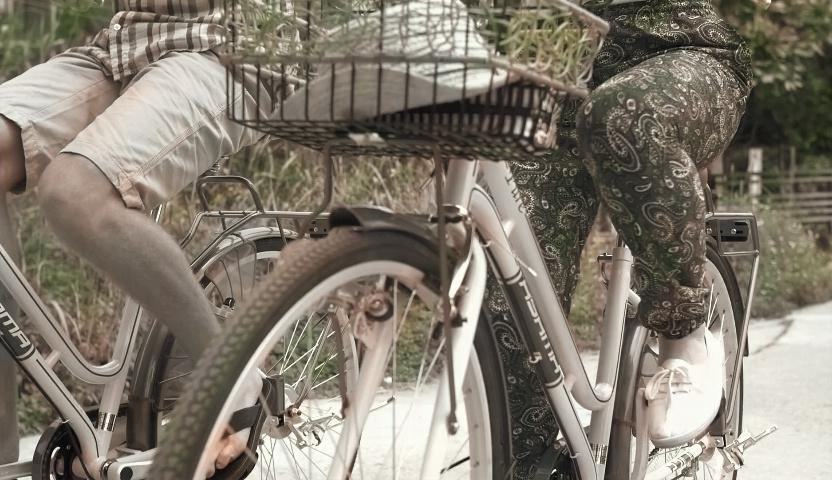}\end{subfigure} & \begin{subfigure}[b]{\myhighreswidth\textwidth}\includegraphics[width=\textwidth]{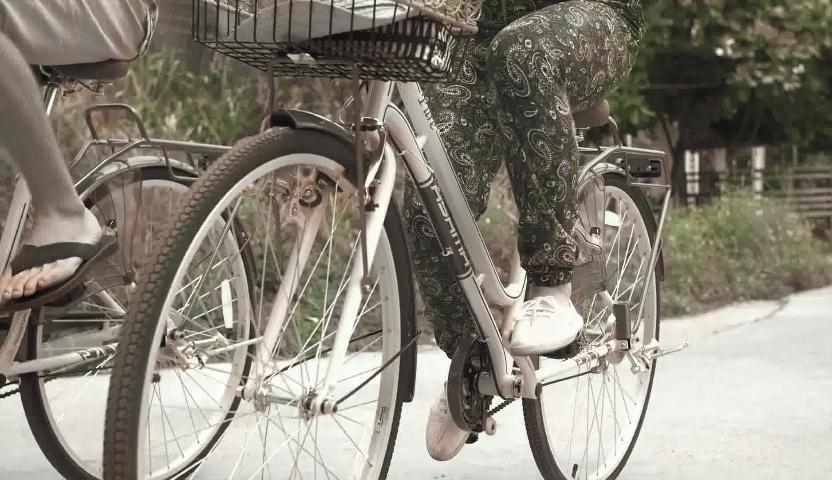}\end{subfigure} & \begin{subfigure}[b]{\myhighreswidth\textwidth}\includegraphics[width=\textwidth]{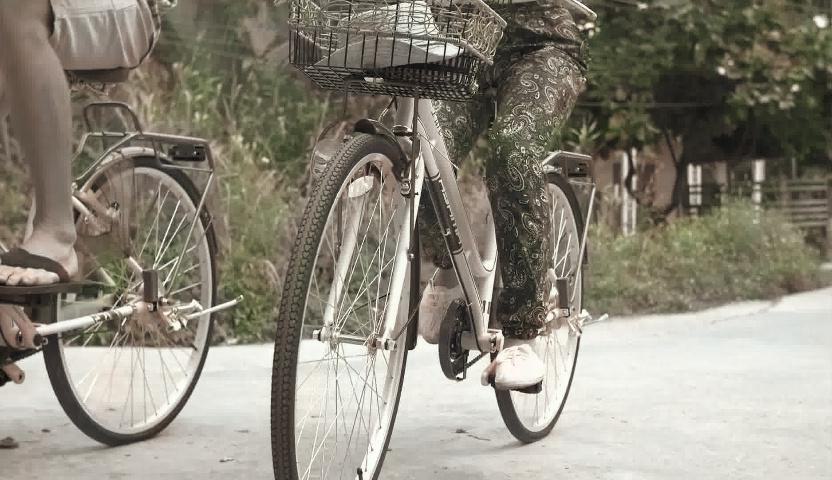}\end{subfigure} & \begin{subfigure}[b]{\myhighreswidth\textwidth}\includegraphics[width=\textwidth]{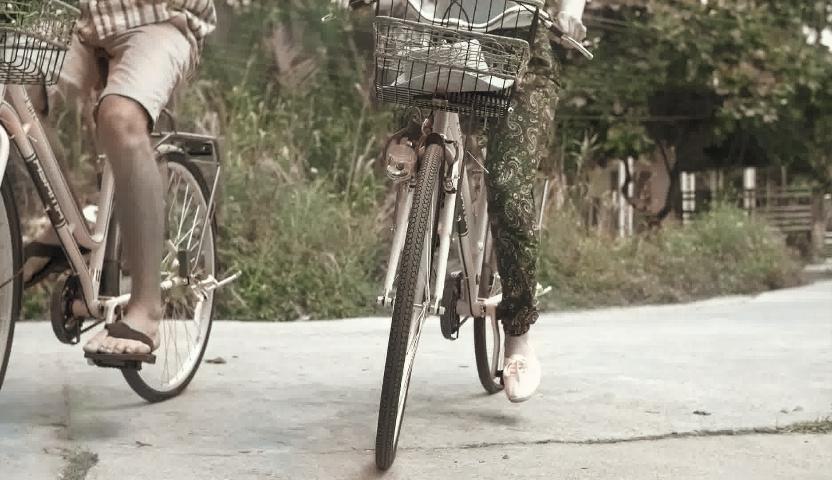}\end{subfigure} & \begin{subfigure}[b]{\myhighreswidth\textwidth}\includegraphics[width=\textwidth]{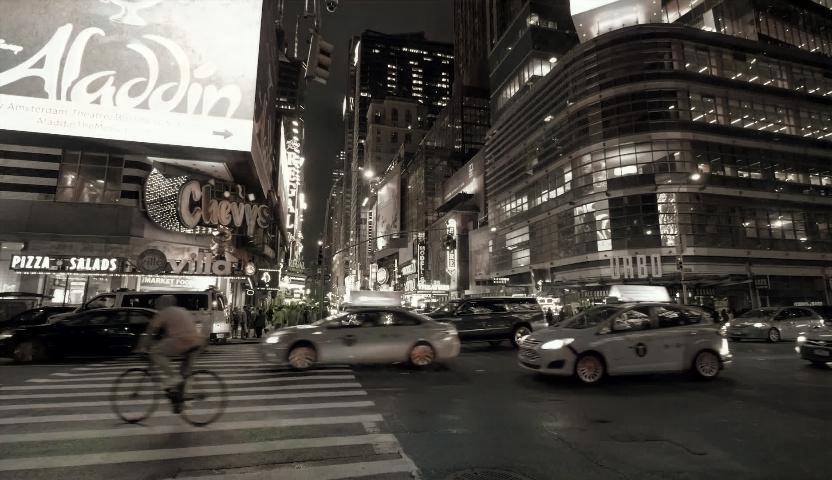}\end{subfigure} & \begin{subfigure}[b]{\myhighreswidth\textwidth}\includegraphics[width=\textwidth]{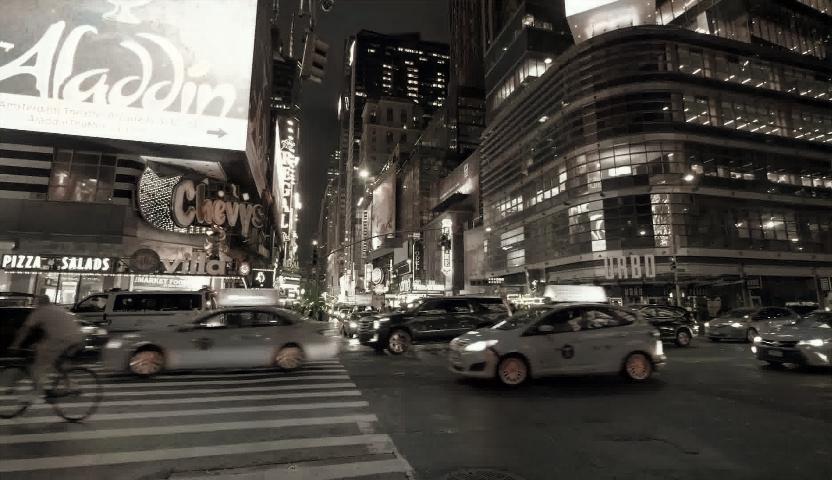}\end{subfigure} & \begin{subfigure}[b]{\myhighreswidth\textwidth}\includegraphics[width=\textwidth]{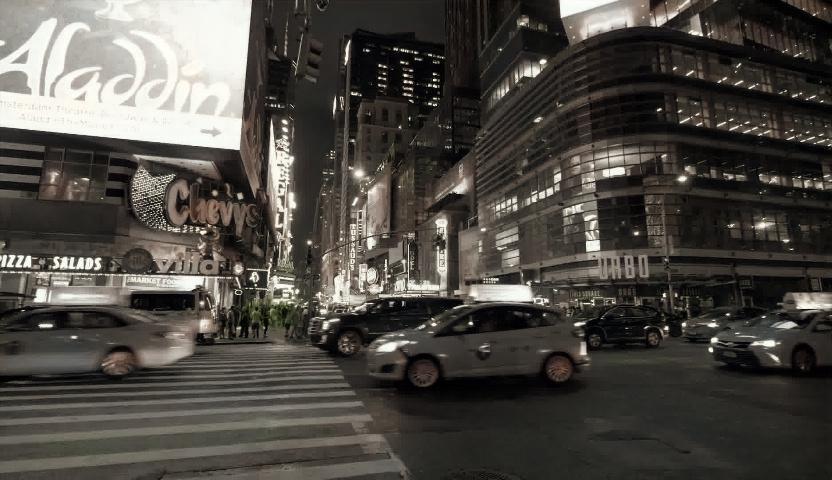}\end{subfigure} & \begin{subfigure}[b]{\myhighreswidth\textwidth}\includegraphics[width=\textwidth]{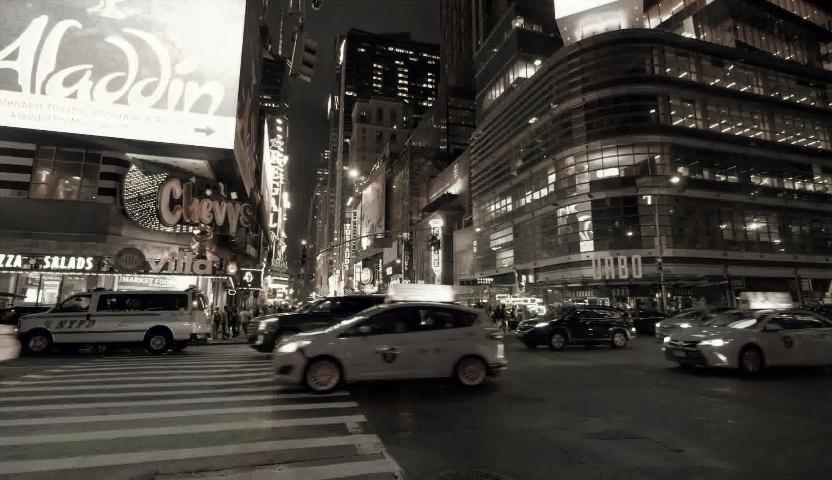}\end{subfigure}\\
\begin{sideways}{\tiny Deoldify}\end{sideways} & \begin{subfigure}[b]{\myhighreswidth\textwidth}\includegraphics[width=\textwidth]{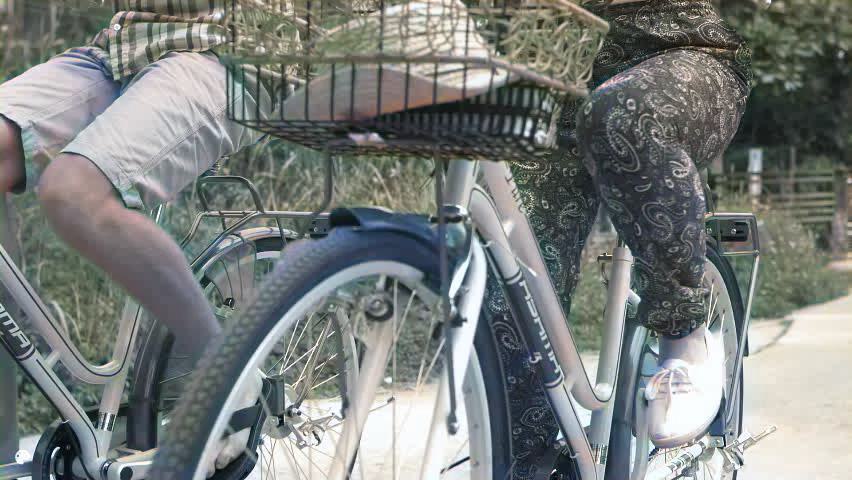}\end{subfigure} & \begin{subfigure}[b]{\myhighreswidth\textwidth}\includegraphics[width=\textwidth]{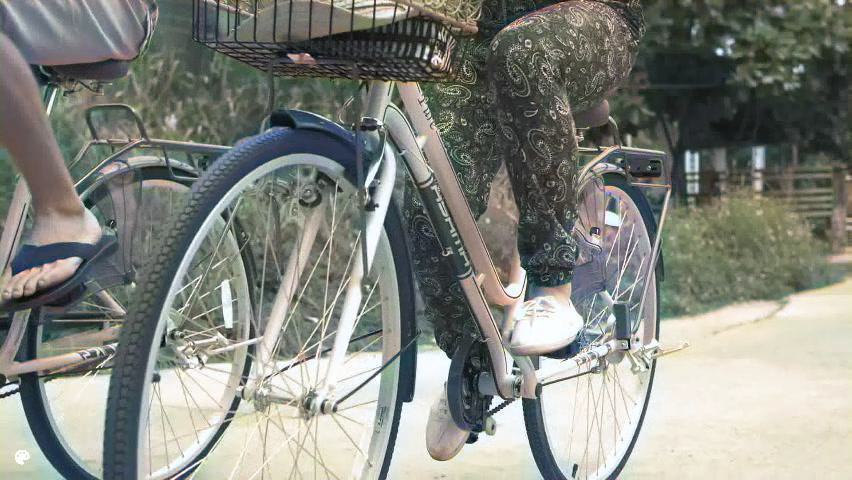}\end{subfigure} & \begin{subfigure}[b]{\myhighreswidth\textwidth}\includegraphics[width=\textwidth]{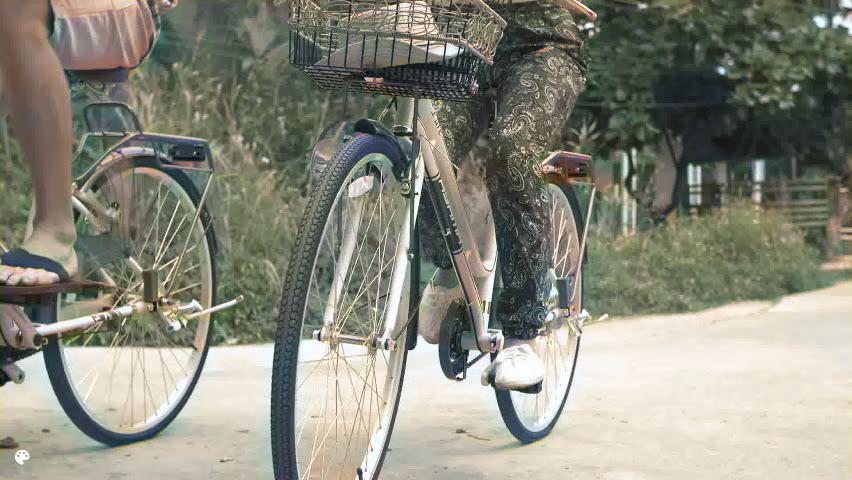}\end{subfigure} & \begin{subfigure}[b]{\myhighreswidth\textwidth}\includegraphics[width=\textwidth]{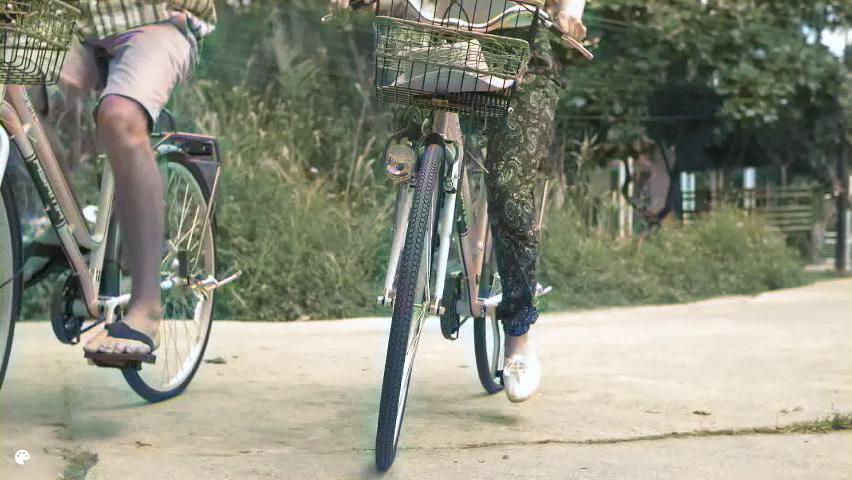}\end{subfigure} & \begin{subfigure}[b]{\myhighreswidth\textwidth}\includegraphics[width=\textwidth]{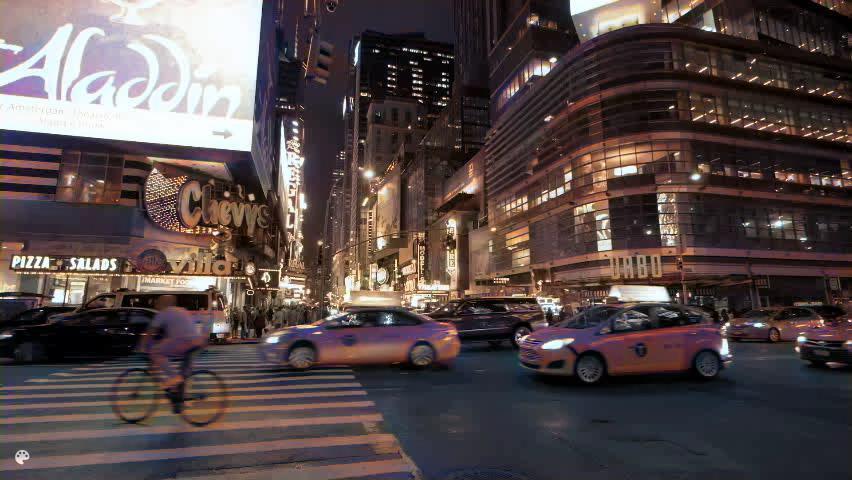}\end{subfigure} & \begin{subfigure}[b]{\myhighreswidth\textwidth}\includegraphics[width=\textwidth]{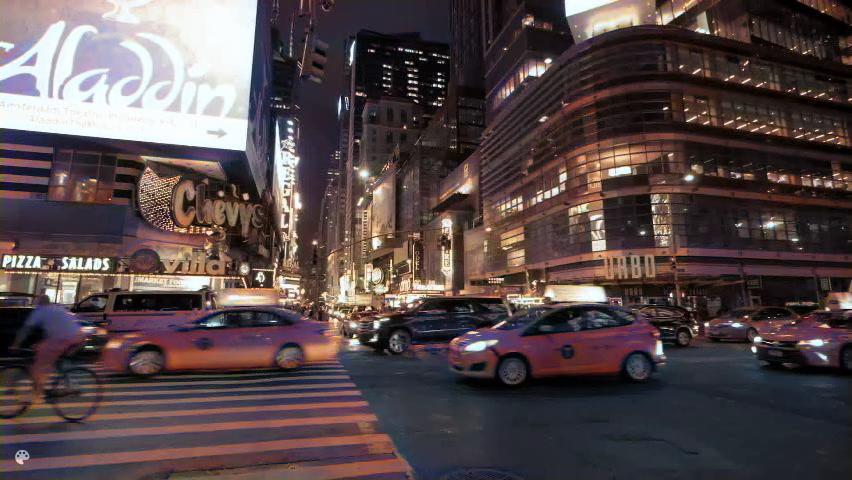}\end{subfigure} & \begin{subfigure}[b]{\myhighreswidth\textwidth}\includegraphics[width=\textwidth]{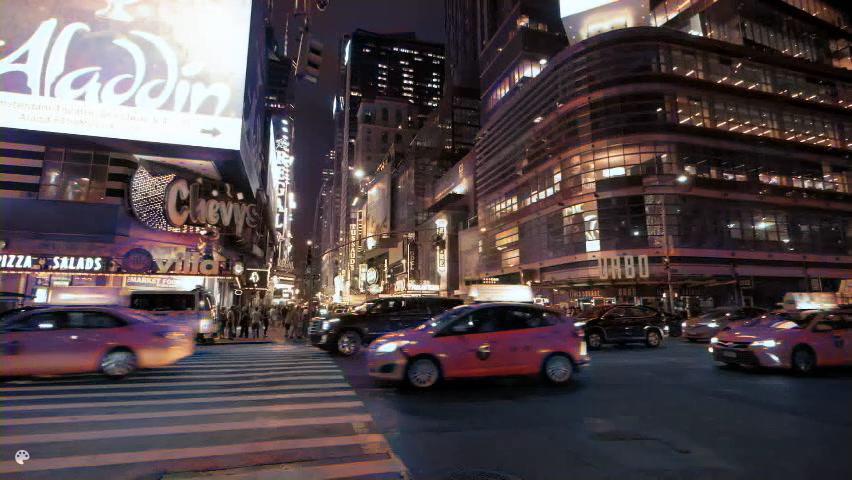}\end{subfigure} & \begin{subfigure}[b]{\myhighreswidth\textwidth}\includegraphics[width=\textwidth]{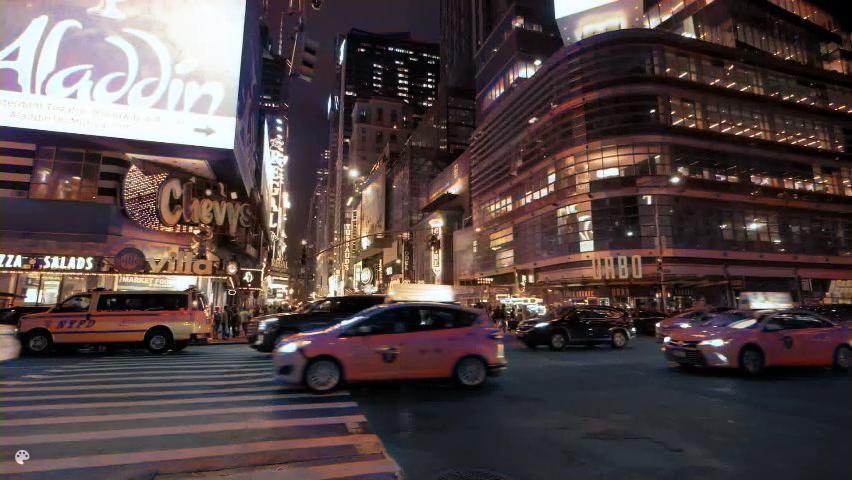}\end{subfigure}\\
\begin{sideways}{\tiny DeepExemplar}\end{sideways} & \begin{subfigure}[b]{\myhighreswidth\textwidth}\includegraphics[width=\textwidth]{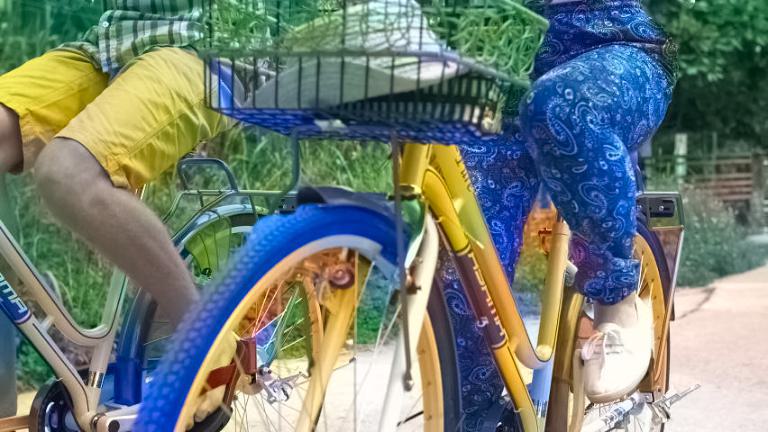}\end{subfigure} & \begin{subfigure}[b]{\myhighreswidth\textwidth}\includegraphics[width=\textwidth]{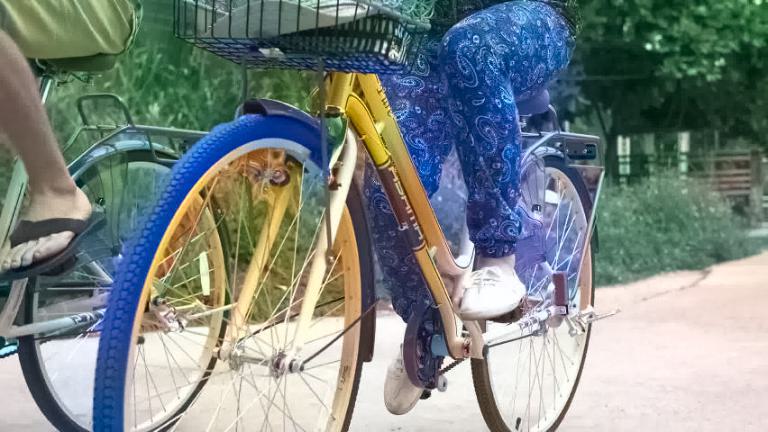}\end{subfigure} & \begin{subfigure}[b]{\myhighreswidth\textwidth}\includegraphics[width=\textwidth]{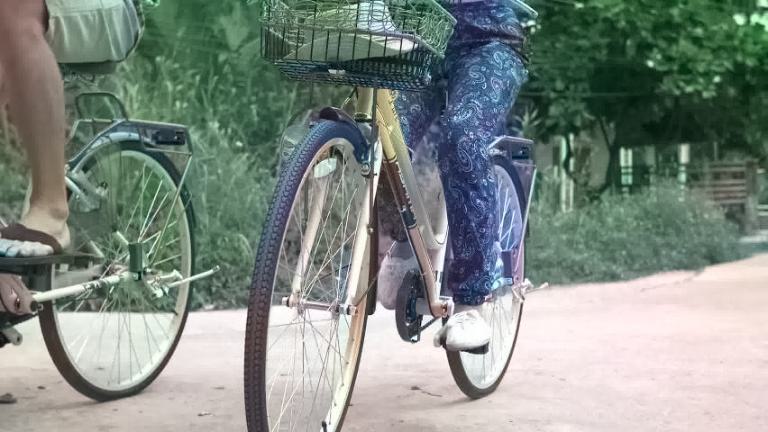}\end{subfigure} & \begin{subfigure}[b]{\myhighreswidth\textwidth}\includegraphics[width=\textwidth]{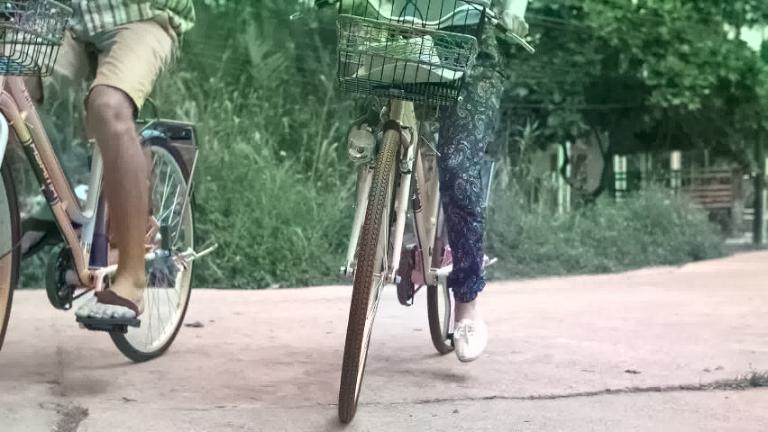}\end{subfigure} & \begin{subfigure}[b]{\myhighreswidth\textwidth}\includegraphics[width=\textwidth]{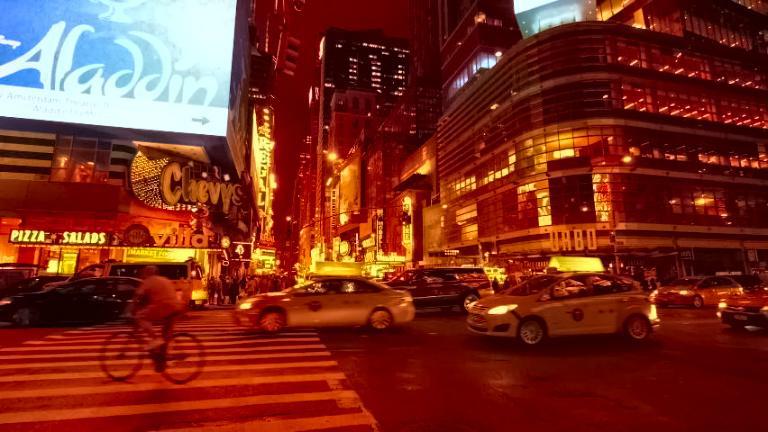}\end{subfigure} & \begin{subfigure}[b]{\myhighreswidth\textwidth}\includegraphics[width=\textwidth]{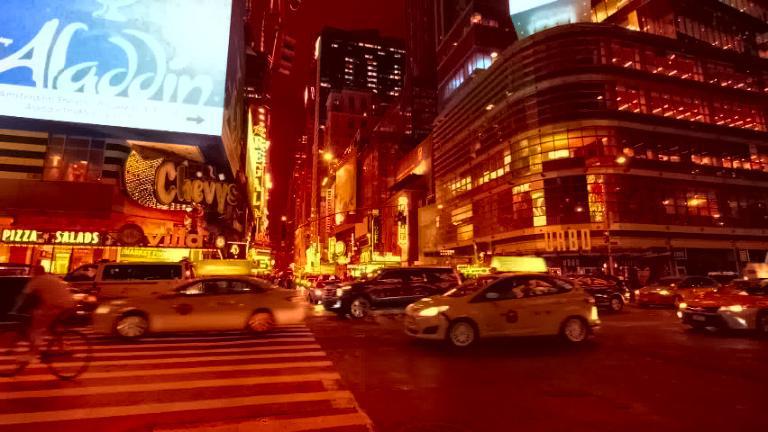}\end{subfigure} & \begin{subfigure}[b]{\myhighreswidth\textwidth}\includegraphics[width=\textwidth]{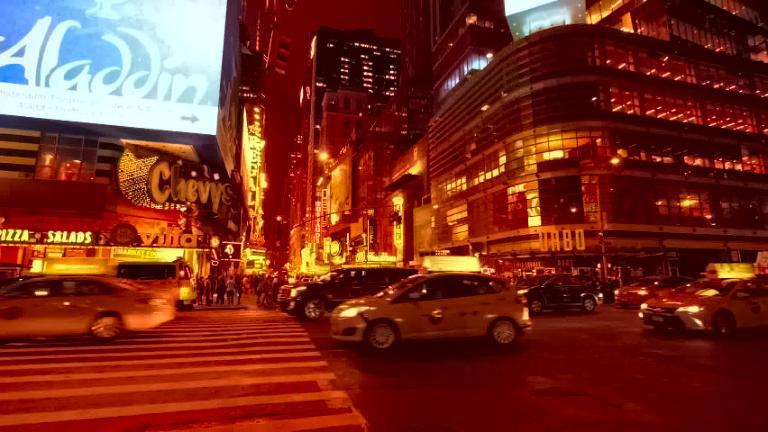}\end{subfigure} & \begin{subfigure}[b]{\myhighreswidth\textwidth}\includegraphics[width=\textwidth]{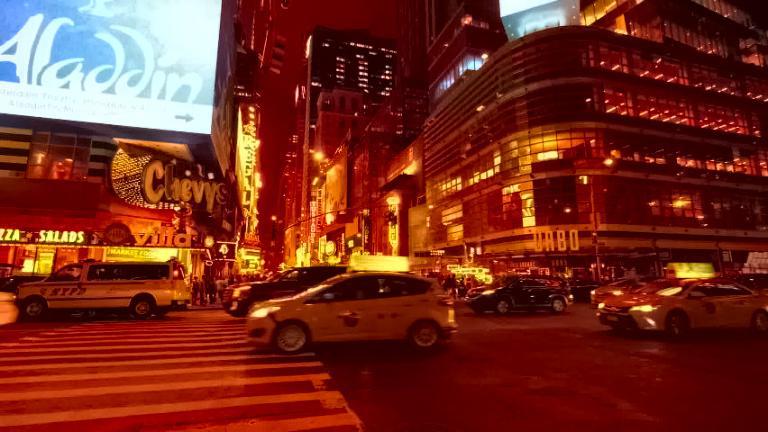}\end{subfigure}\\
\begin{sideways}{\tiny DeepRemaster}\end{sideways} & \begin{subfigure}[b]{\myhighreswidth\textwidth}\includegraphics[width=\textwidth]{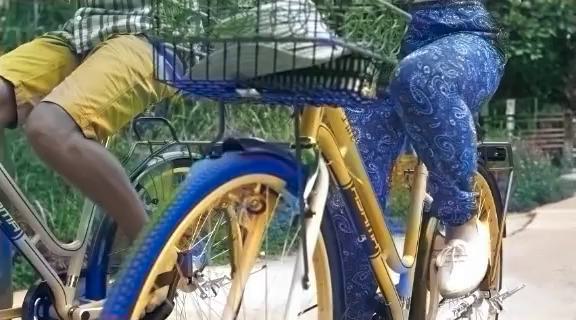}\end{subfigure} & \begin{subfigure}[b]{\myhighreswidth\textwidth}\includegraphics[width=\textwidth]{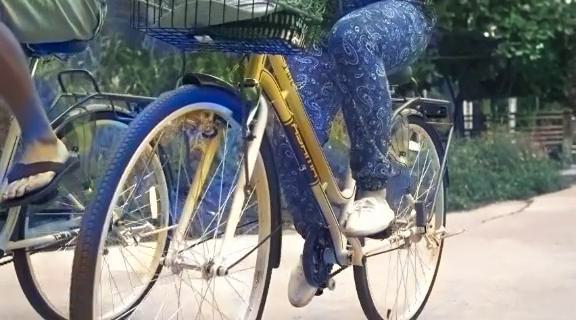}\end{subfigure} & \begin{subfigure}[b]{\myhighreswidth\textwidth}\includegraphics[width=\textwidth]{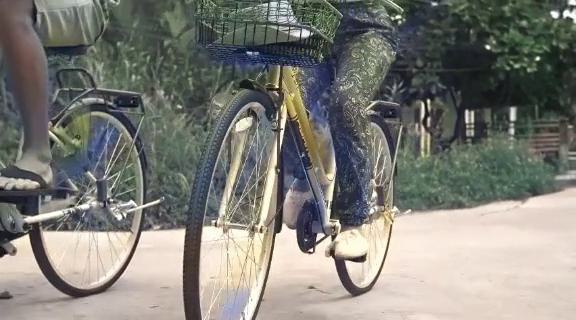}\end{subfigure} & \begin{subfigure}[b]{\myhighreswidth\textwidth}\includegraphics[width=\textwidth]{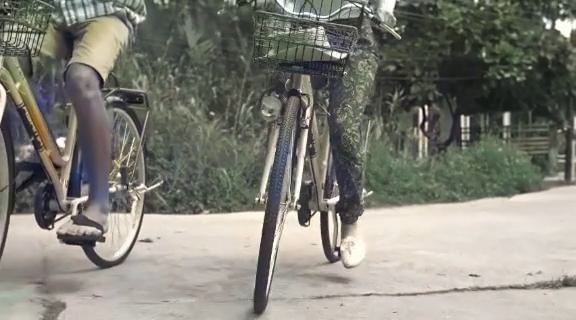}\end{subfigure} & \begin{subfigure}[b]{\myhighreswidth\textwidth}\includegraphics[width=\textwidth]{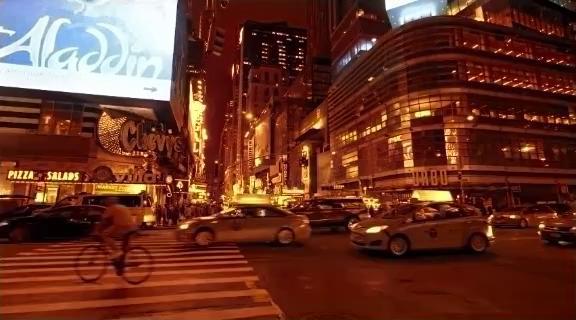}\end{subfigure} & \begin{subfigure}[b]{\myhighreswidth\textwidth}\includegraphics[width=\textwidth]{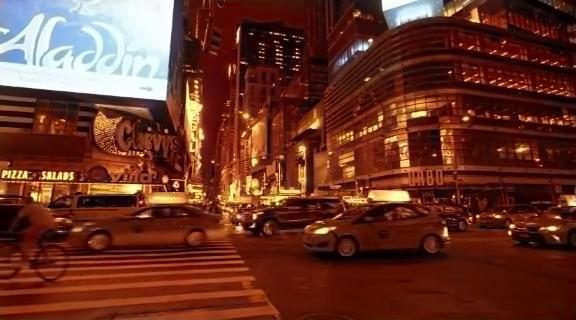}\end{subfigure} & \begin{subfigure}[b]{\myhighreswidth\textwidth}\includegraphics[width=\textwidth]{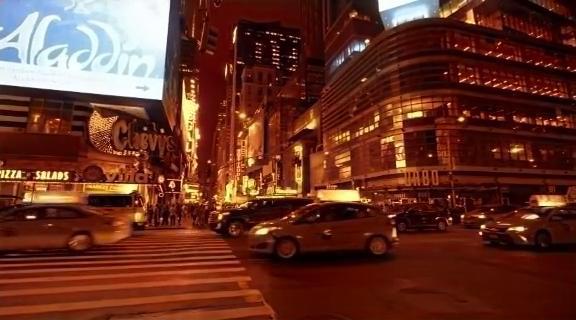}\end{subfigure} & \begin{subfigure}[b]{\myhighreswidth\textwidth}\includegraphics[width=\textwidth]{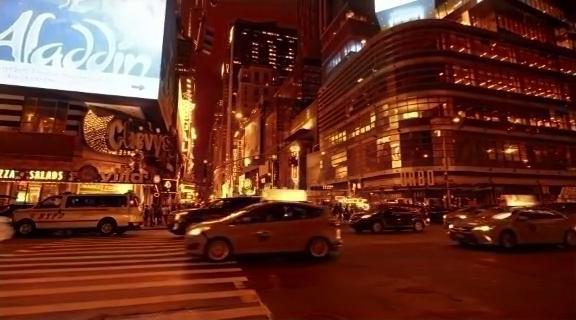}\end{subfigure}\\
\begin{sideways}{\tiny TCVC}\end{sideways} & \begin{subfigure}[b]{\myhighreswidth\textwidth}\includegraphics[width=\textwidth]{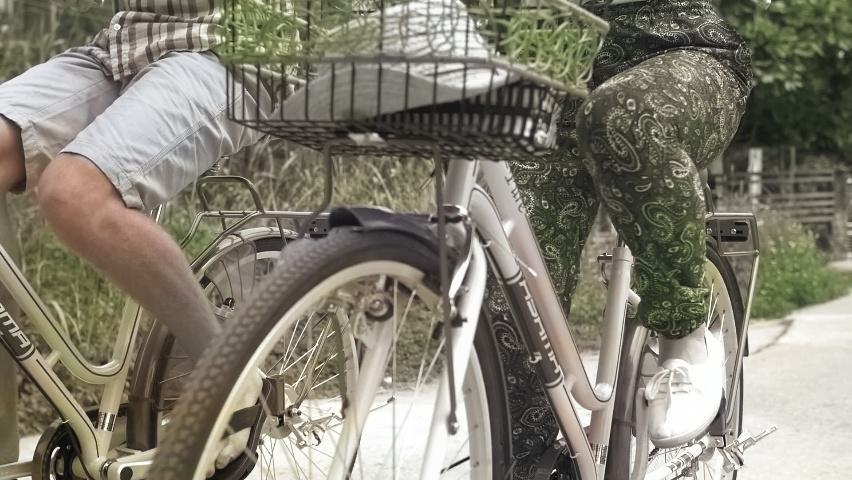}\end{subfigure} & \begin{subfigure}[b]{\myhighreswidth\textwidth}\includegraphics[width=\textwidth]{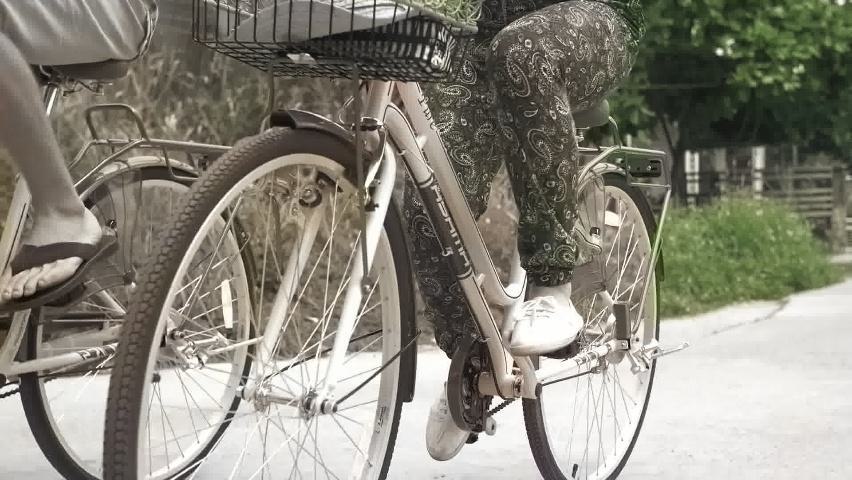}\end{subfigure} & \begin{subfigure}[b]{\myhighreswidth\textwidth}\includegraphics[width=\textwidth]{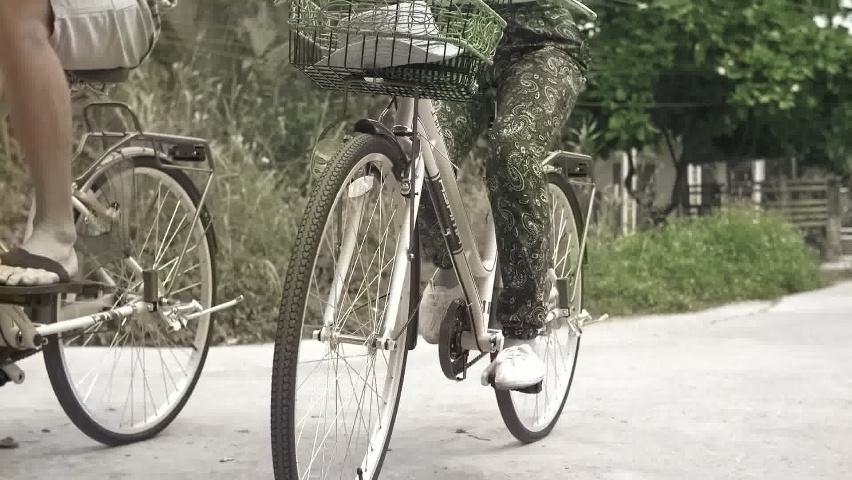}\end{subfigure} & \begin{subfigure}[b]{\myhighreswidth\textwidth}\includegraphics[width=\textwidth]{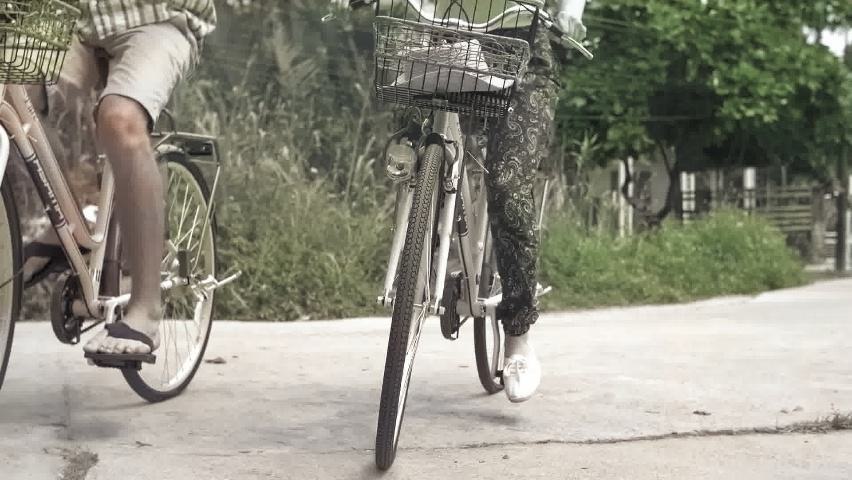}\end{subfigure} & \begin{subfigure}[b]{\myhighreswidth\textwidth}\includegraphics[width=\textwidth]{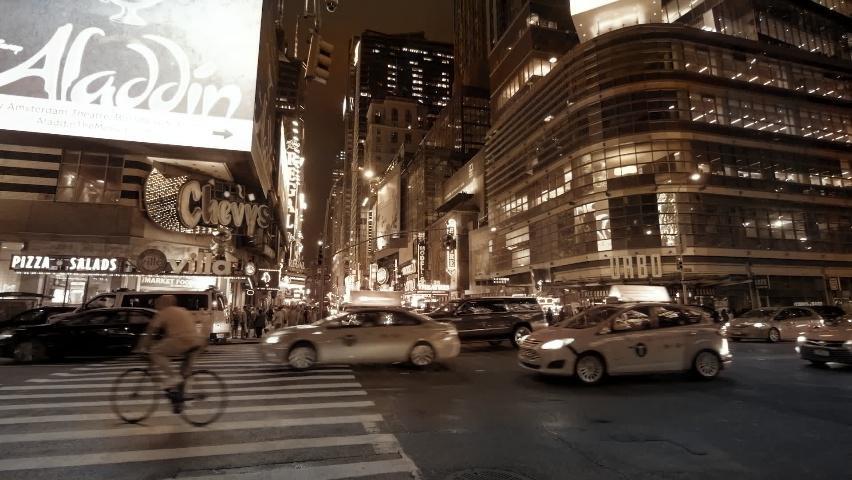}\end{subfigure} & \begin{subfigure}[b]{\myhighreswidth\textwidth}\includegraphics[width=\textwidth]{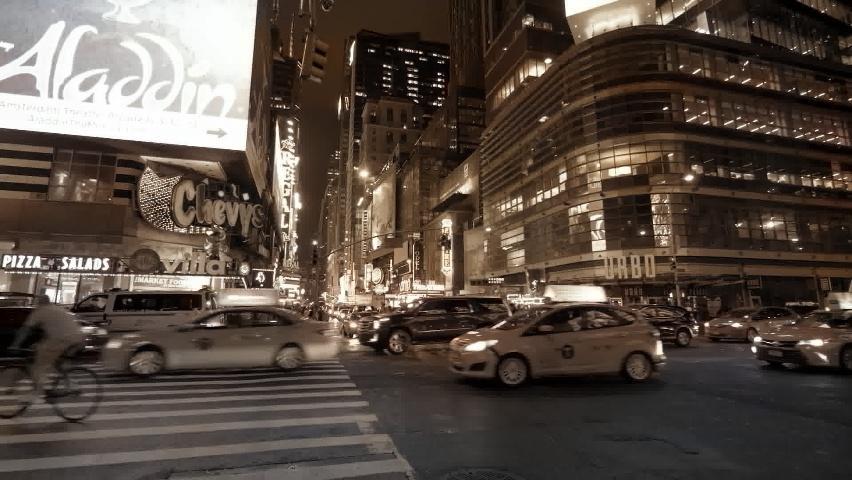}\end{subfigure} & \begin{subfigure}[b]{\myhighreswidth\textwidth}\includegraphics[width=\textwidth]{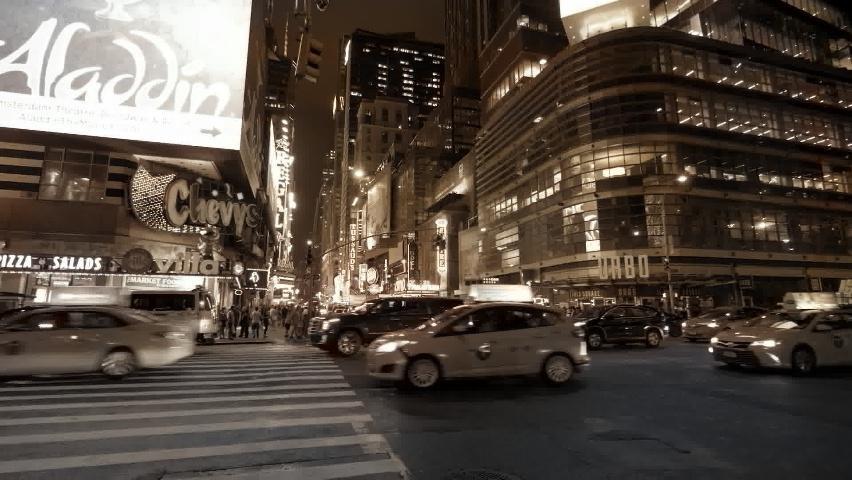}\end{subfigure} & \begin{subfigure}[b]{\myhighreswidth\textwidth}\includegraphics[width=\textwidth]{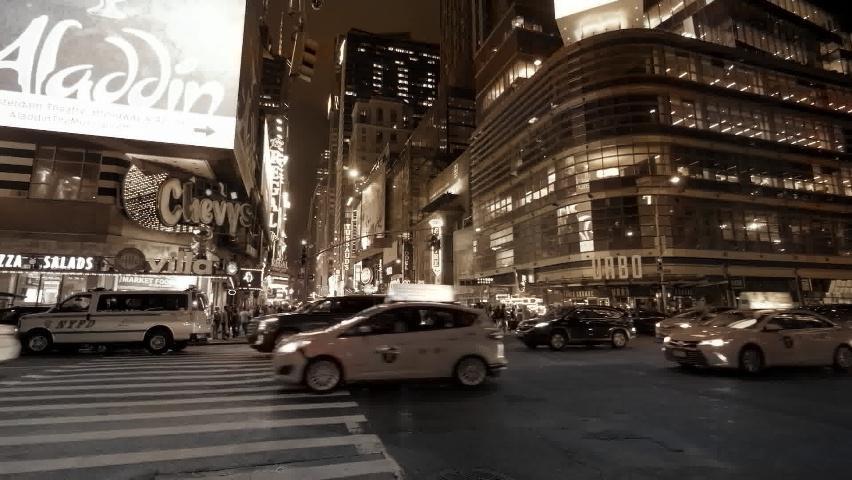}\end{subfigure}\\
\begin{sideways}{\tiny VCGAN}\end{sideways} & \begin{subfigure}[b]{\myhighreswidth\textwidth}\includegraphics[width=\textwidth]{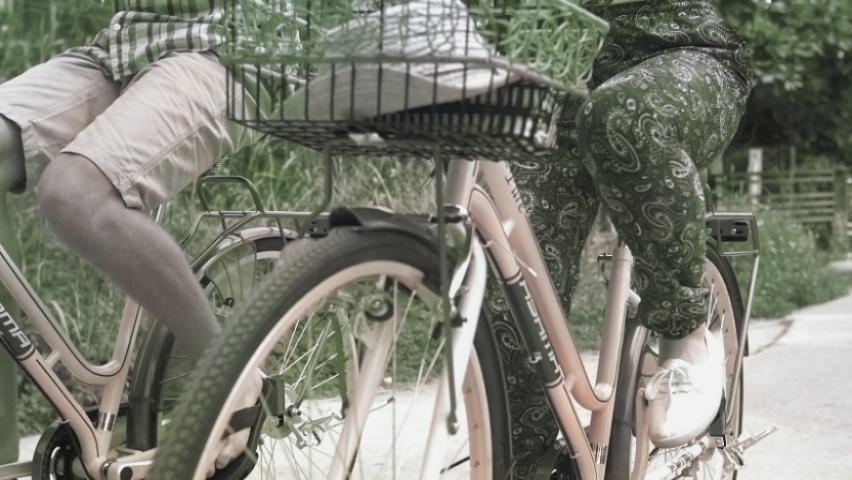}\end{subfigure} & \begin{subfigure}[b]{\myhighreswidth\textwidth}\includegraphics[width=\textwidth]{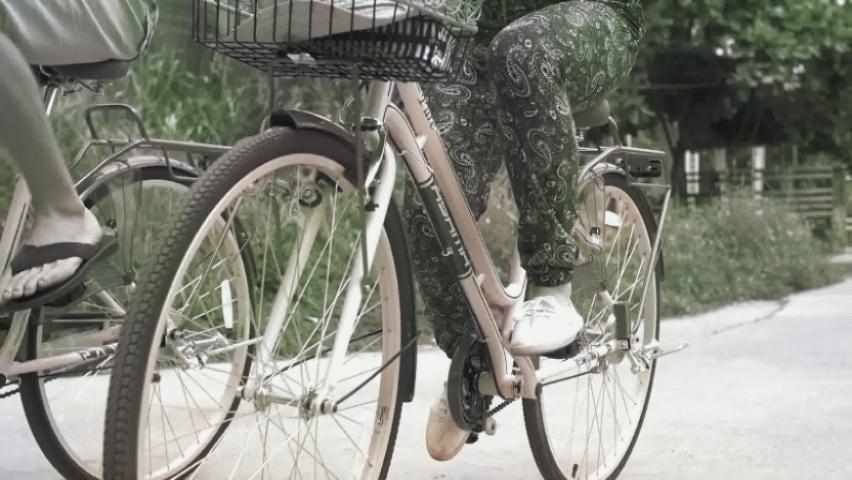}\end{subfigure} & \begin{subfigure}[b]{\myhighreswidth\textwidth}\includegraphics[width=\textwidth]{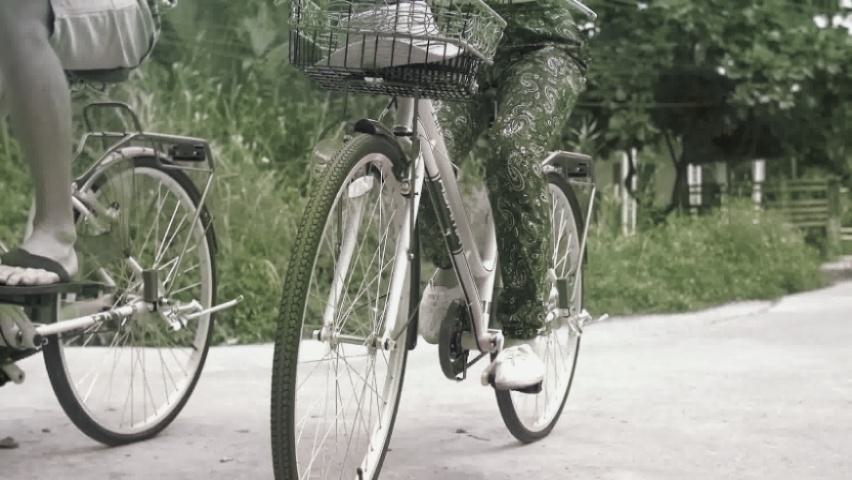}\end{subfigure} & \begin{subfigure}[b]{\myhighreswidth\textwidth}\includegraphics[width=\textwidth]{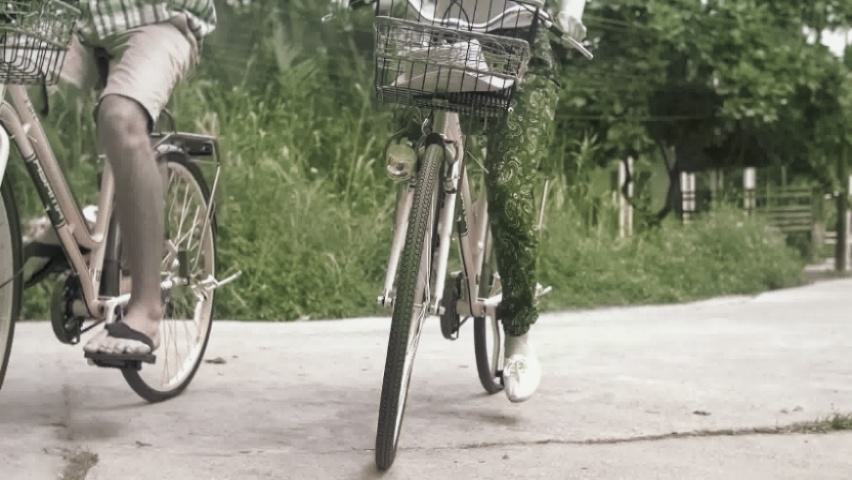}\end{subfigure} & \begin{subfigure}[b]{\myhighreswidth\textwidth}\includegraphics[width=\textwidth]{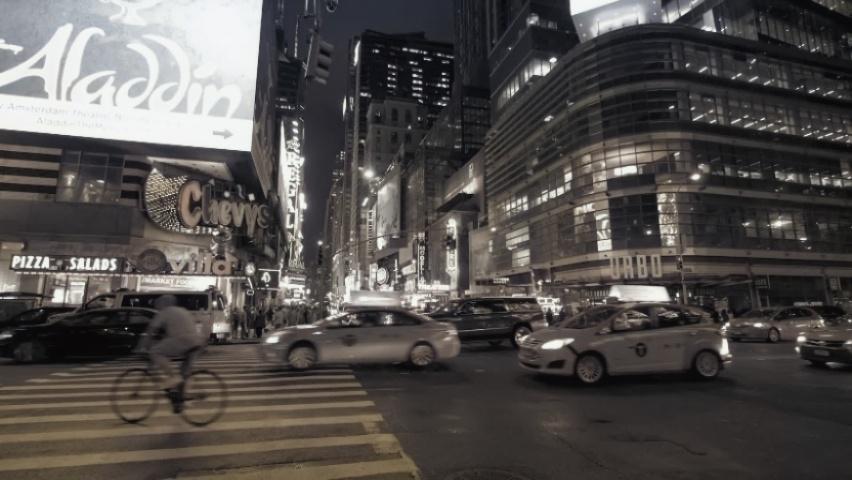}\end{subfigure} & \begin{subfigure}[b]{\myhighreswidth\textwidth}\includegraphics[width=\textwidth]{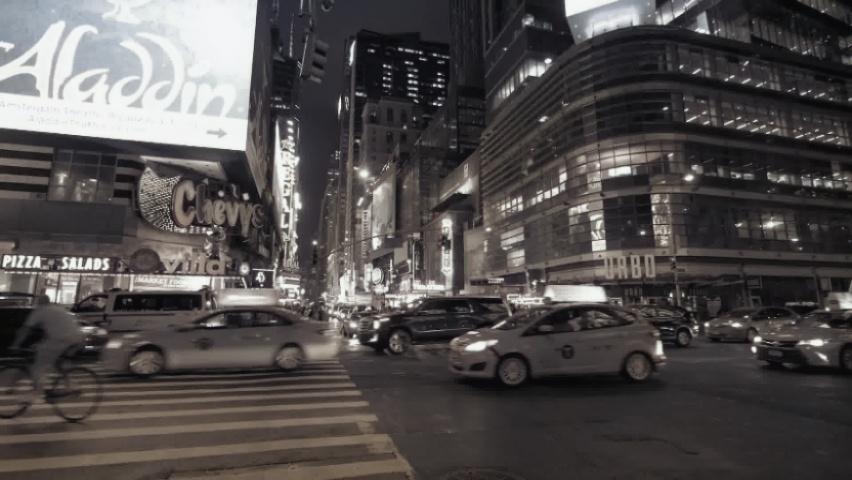}\end{subfigure} & \begin{subfigure}[b]{\myhighreswidth\textwidth}\includegraphics[width=\textwidth]{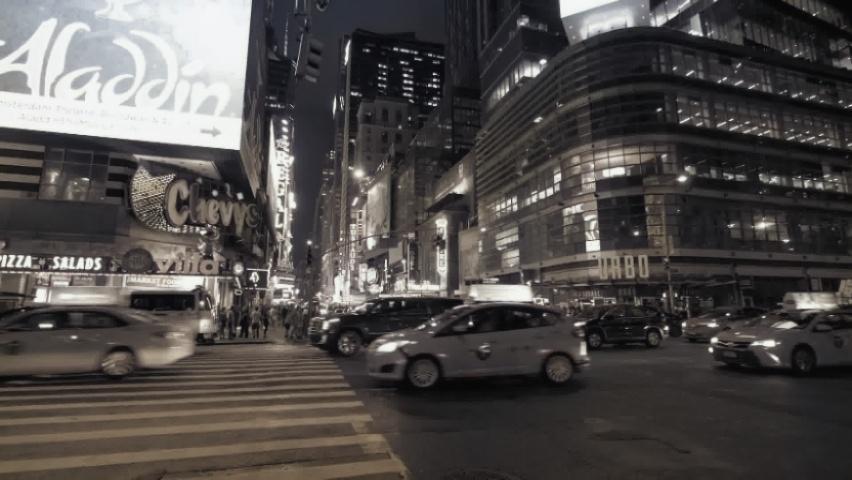}\end{subfigure} & \begin{subfigure}[b]{\myhighreswidth\textwidth}\includegraphics[width=\textwidth]{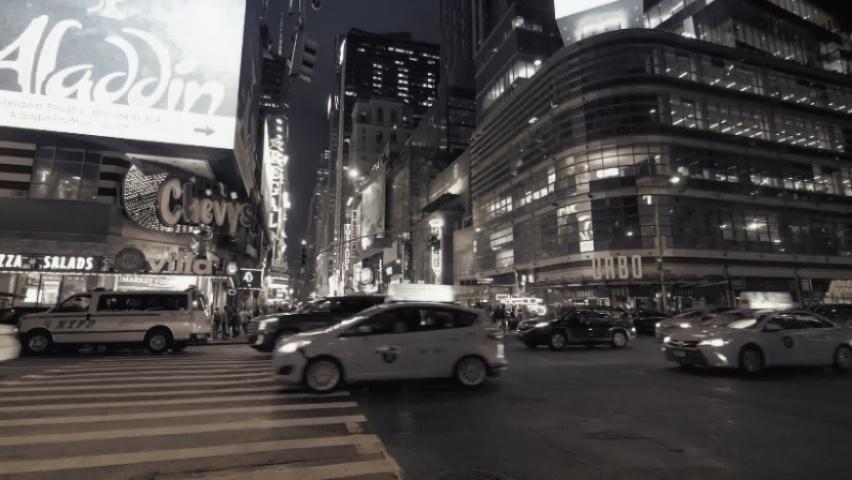}\end{subfigure}\\
\begin{sideways}{\tiny Ours}\end{sideways} & \begin{subfigure}[b]{\myhighreswidth\textwidth}\includegraphics[width=\textwidth]{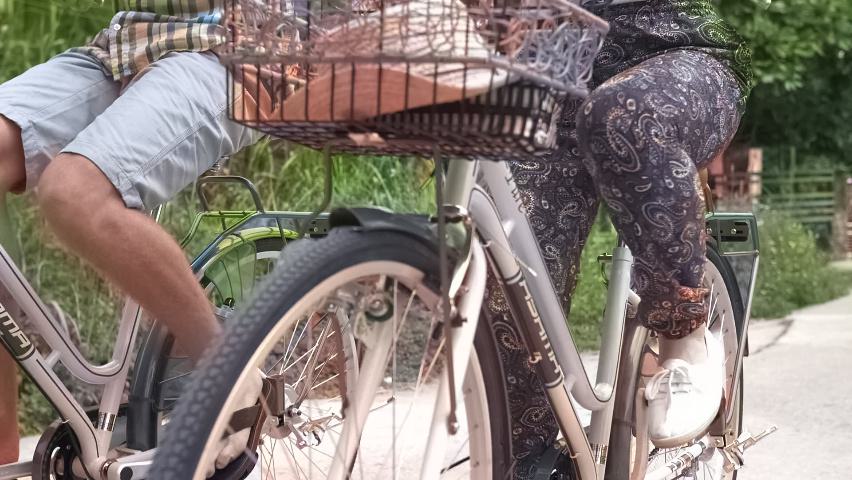}\end{subfigure} & \begin{subfigure}[b]{\myhighreswidth\textwidth}\includegraphics[width=\textwidth]{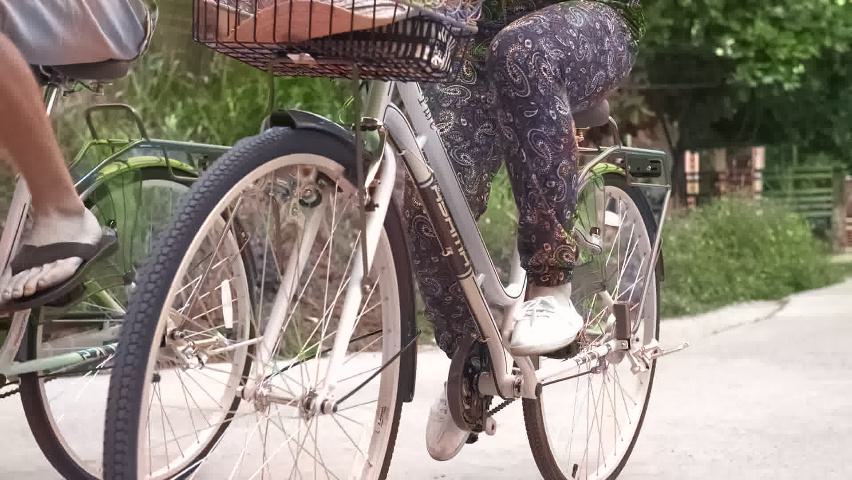}\end{subfigure} & \begin{subfigure}[b]{\myhighreswidth\textwidth}\includegraphics[width=\textwidth]{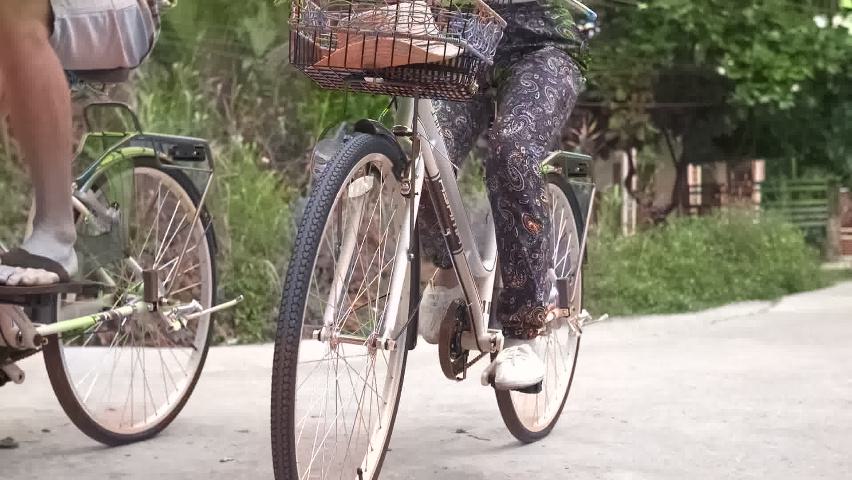}\end{subfigure} & \begin{subfigure}[b]{\myhighreswidth\textwidth}\includegraphics[width=\textwidth]{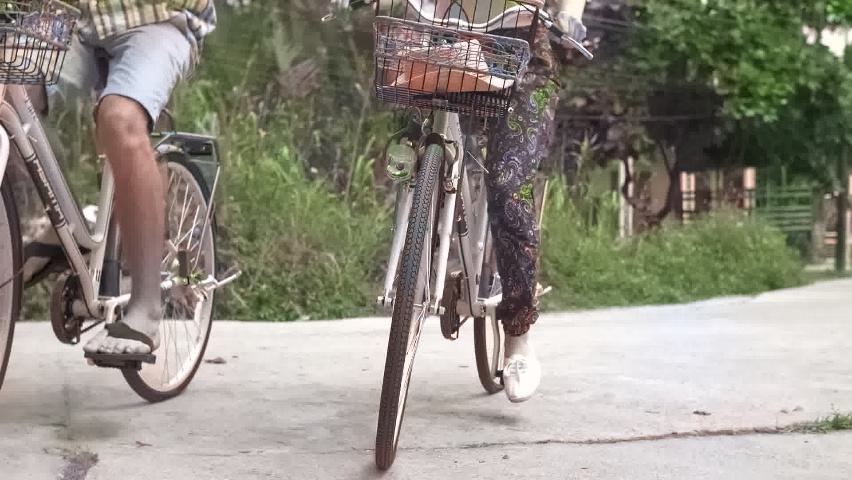}\end{subfigure} & \begin{subfigure}[b]{\myhighreswidth\textwidth}\includegraphics[width=\textwidth]{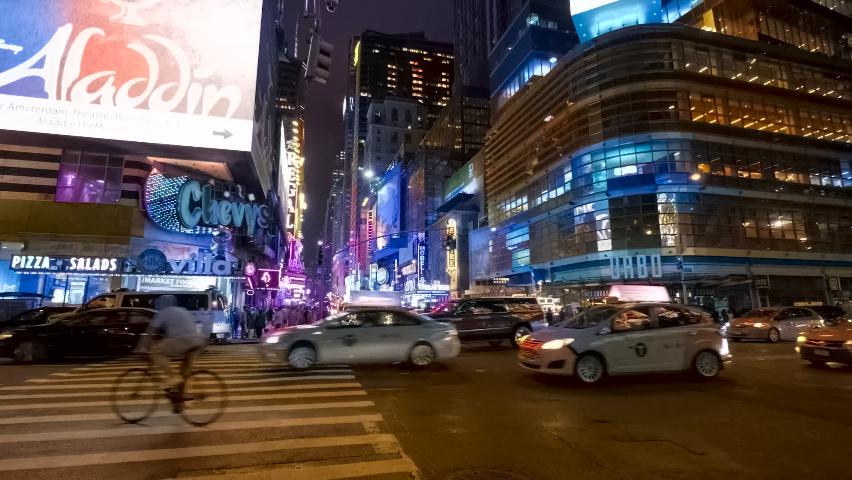}\end{subfigure} & \begin{subfigure}[b]{\myhighreswidth\textwidth}\includegraphics[width=\textwidth]{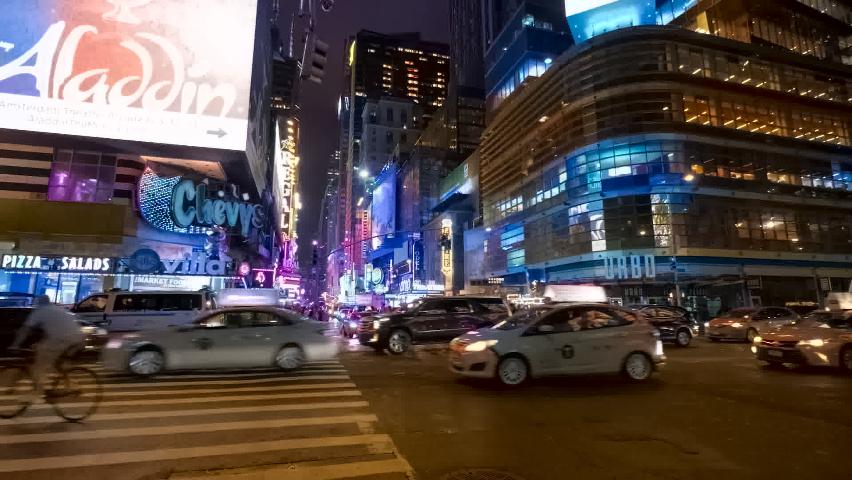}\end{subfigure} & \begin{subfigure}[b]{\myhighreswidth\textwidth}\includegraphics[width=\textwidth]{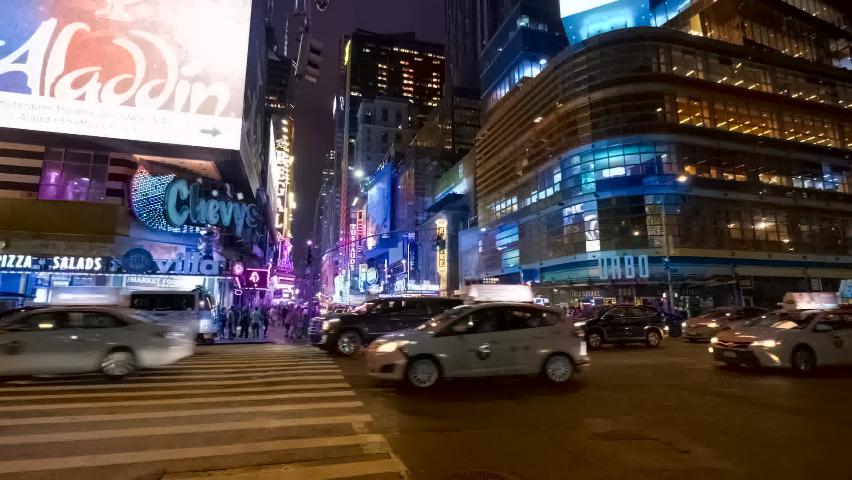}\end{subfigure} & \begin{subfigure}[b]{\myhighreswidth\textwidth}\includegraphics[width=\textwidth]{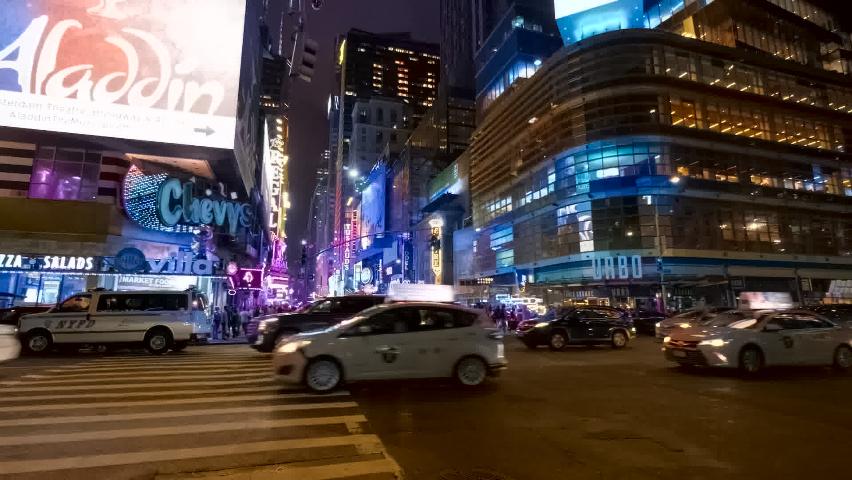}\end{subfigure}\\

& T=0 & T=15 & T=30 & T=45 & T=0 & T=15 & T=30 & T=45\\
\begin{sideways}{\tiny Grayscale Input}\end{sideways} & \begin{subfigure}[b]{\myhighreswidth\textwidth}\includegraphics[width=\textwidth]{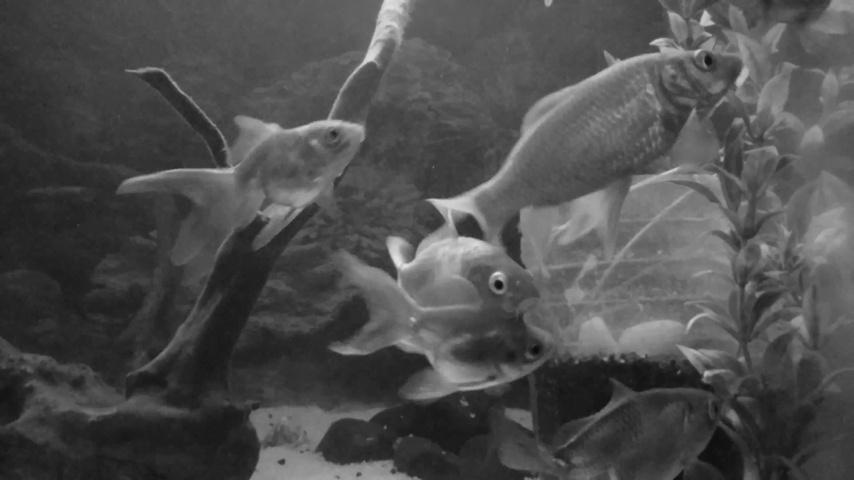}\end{subfigure} & \begin{subfigure}[b]{\myhighreswidth\textwidth}\includegraphics[width=\textwidth]{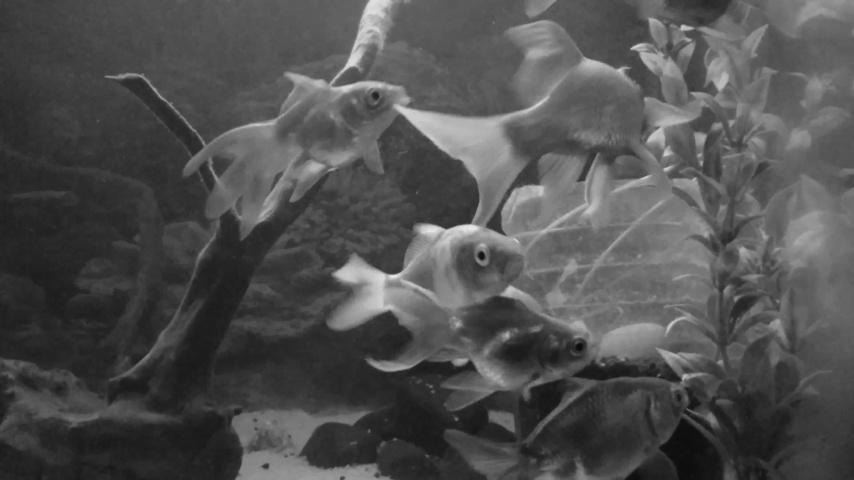}\end{subfigure} & \begin{subfigure}[b]{\myhighreswidth\textwidth}\includegraphics[width=\textwidth]{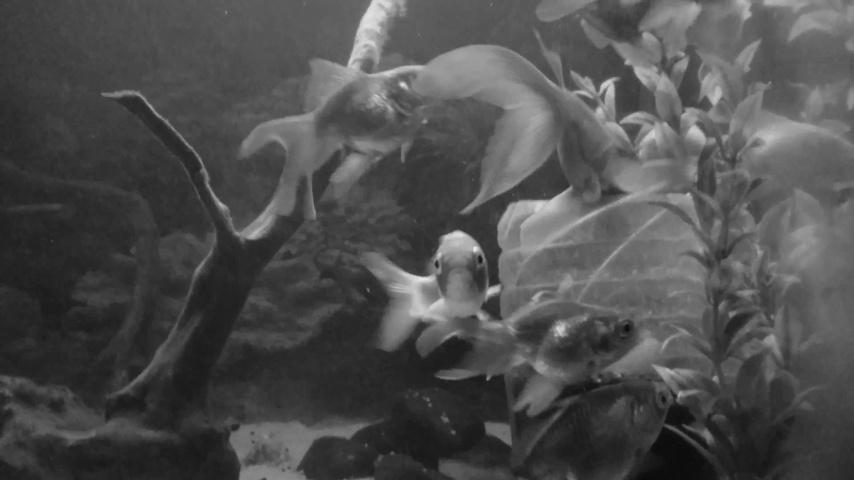}\end{subfigure} & \begin{subfigure}[b]{\myhighreswidth\textwidth}\includegraphics[width=\textwidth]{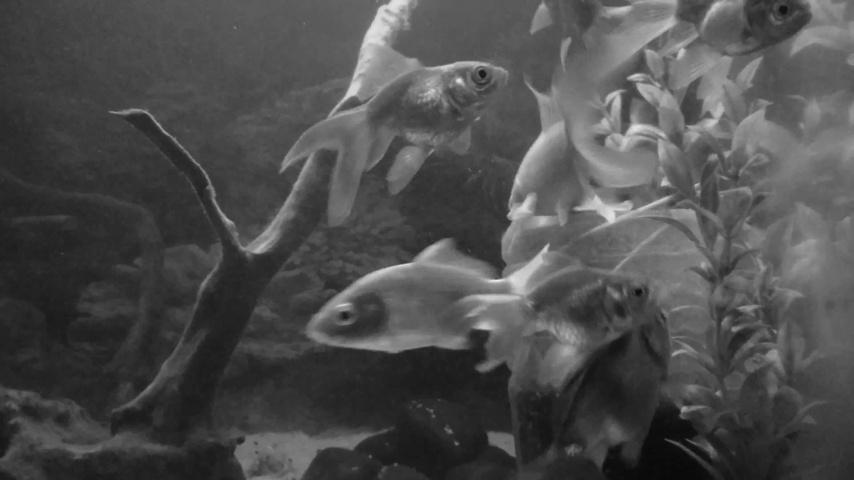}\end{subfigure} & \begin{subfigure}[b]{\myhighreswidth\textwidth}\includegraphics[width=\textwidth]{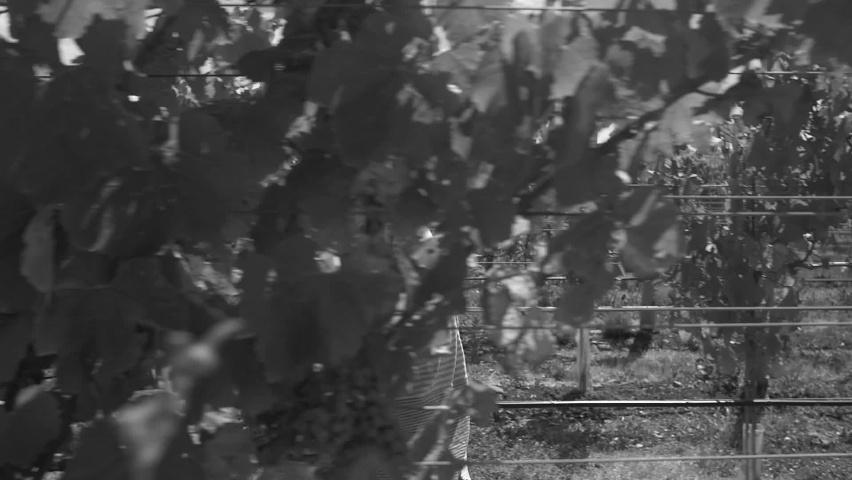}\end{subfigure} & \begin{subfigure}[b]{\myhighreswidth\textwidth}\includegraphics[width=\textwidth]{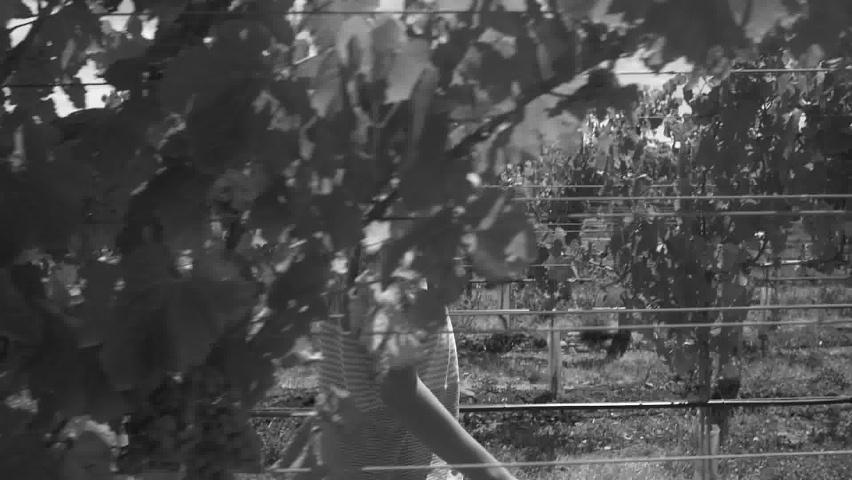}\end{subfigure} & \begin{subfigure}[b]{\myhighreswidth\textwidth}\includegraphics[width=\textwidth]{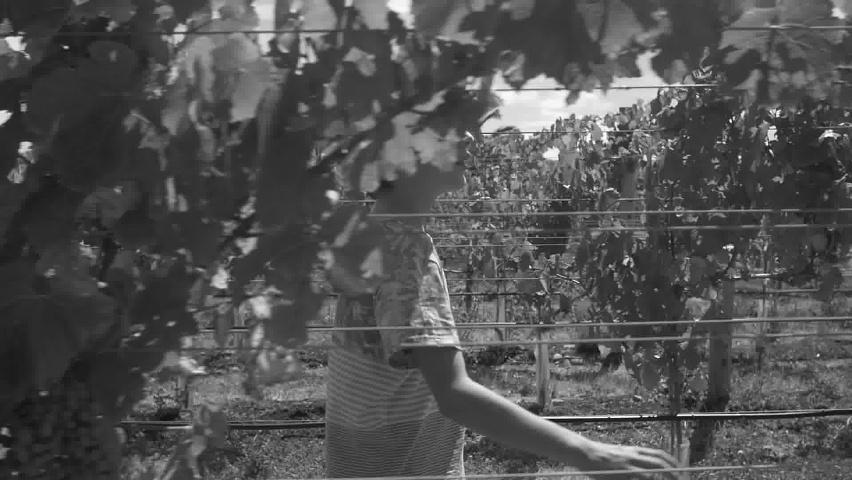}\end{subfigure} & \begin{subfigure}[b]{\myhighreswidth\textwidth}\includegraphics[width=\textwidth]{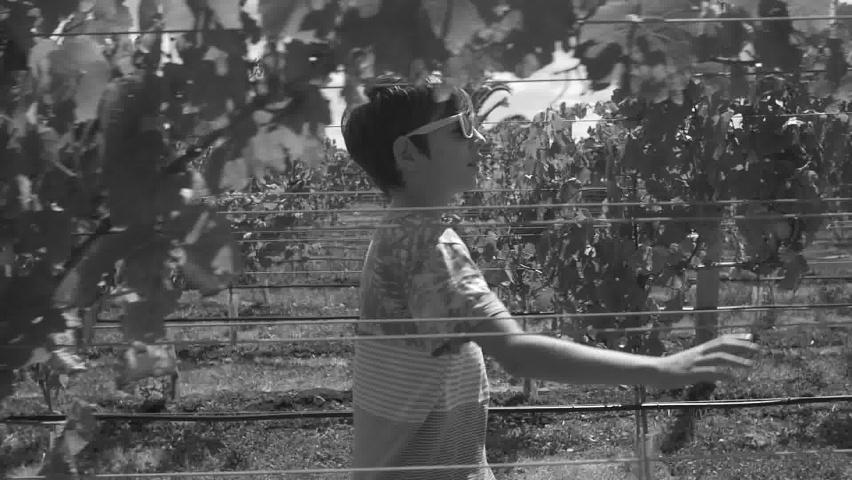}\end{subfigure}\\
\begin{sideways}{\tiny Ground Truth}\end{sideways} & \begin{subfigure}[b]{\myhighreswidth\textwidth}\includegraphics[width=\textwidth]{graphs/cmp00/gt_00000.jpg}\end{subfigure} & \begin{subfigure}[b]{\myhighreswidth\textwidth}\includegraphics[width=\textwidth]{graphs/cmp00/gt_00015.jpg}\end{subfigure} & \begin{subfigure}[b]{\myhighreswidth\textwidth}\includegraphics[width=\textwidth]{graphs/cmp00/gt_00030.jpg}\end{subfigure} & \begin{subfigure}[b]{\myhighreswidth\textwidth}\includegraphics[width=\textwidth]{graphs/cmp00/gt_00045.jpg}\end{subfigure} & \begin{subfigure}[b]{\myhighreswidth\textwidth}\includegraphics[width=\textwidth]{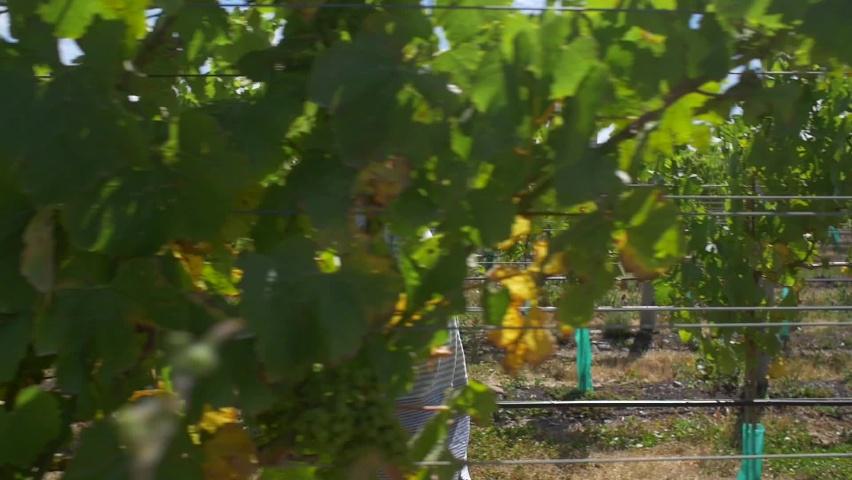}\end{subfigure} & \begin{subfigure}[b]{\myhighreswidth\textwidth}\includegraphics[width=\textwidth]{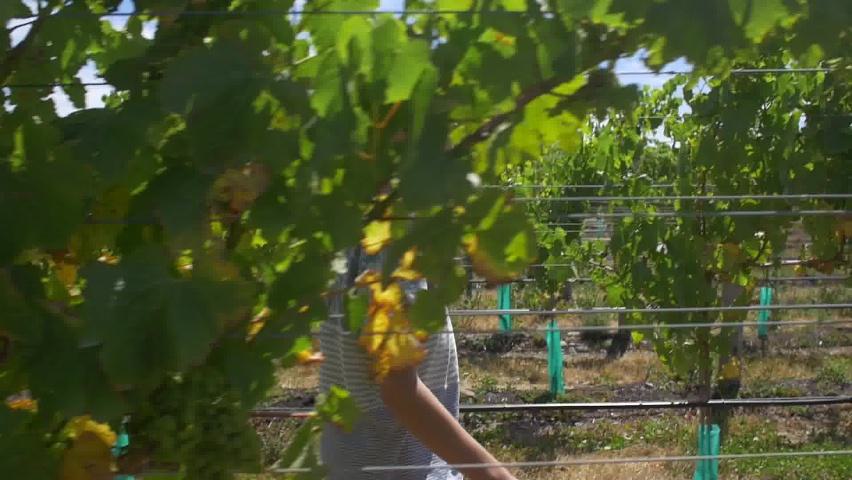}\end{subfigure} & \begin{subfigure}[b]{\myhighreswidth\textwidth}\includegraphics[width=\textwidth]{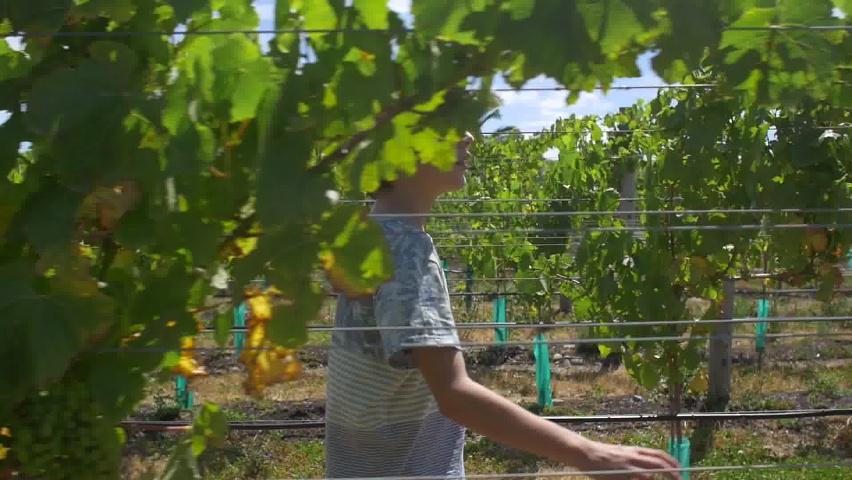}\end{subfigure} & \begin{subfigure}[b]{\myhighreswidth\textwidth}\includegraphics[width=\textwidth]{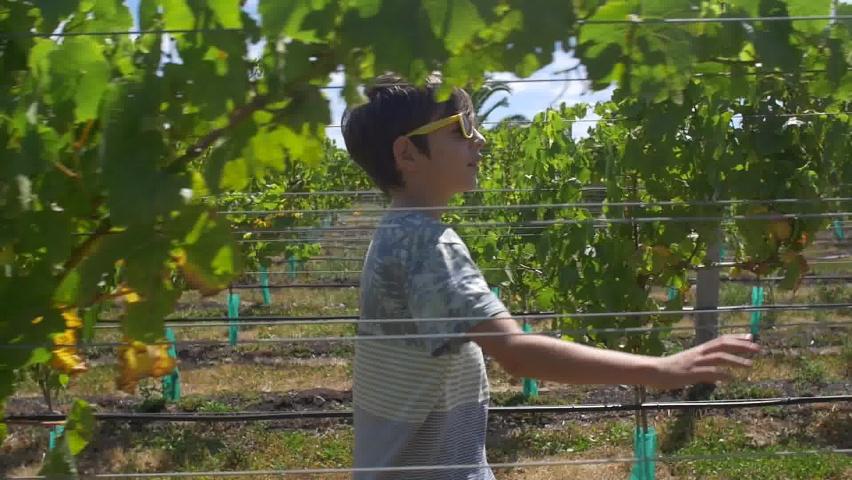}\end{subfigure}\\
\begin{sideways}{\tiny AutoColor}\end{sideways} & \begin{subfigure}[b]{\myhighreswidth\textwidth}\includegraphics[width=\textwidth]{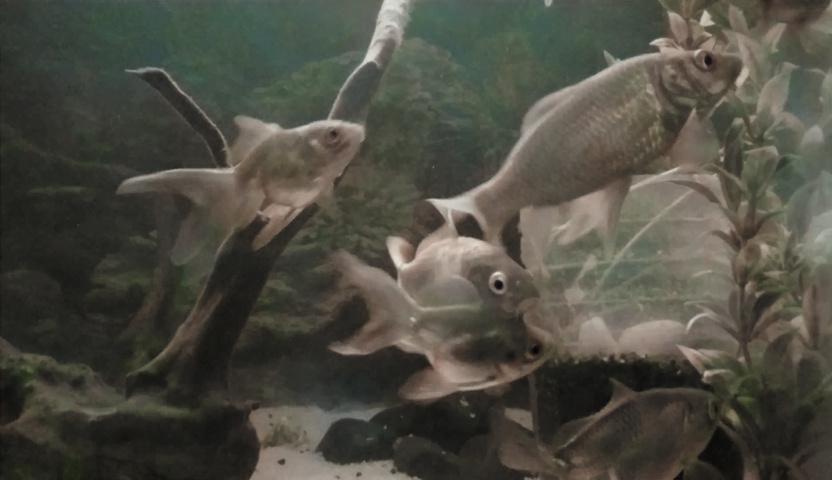}\end{subfigure} & \begin{subfigure}[b]{\myhighreswidth\textwidth}\includegraphics[width=\textwidth]{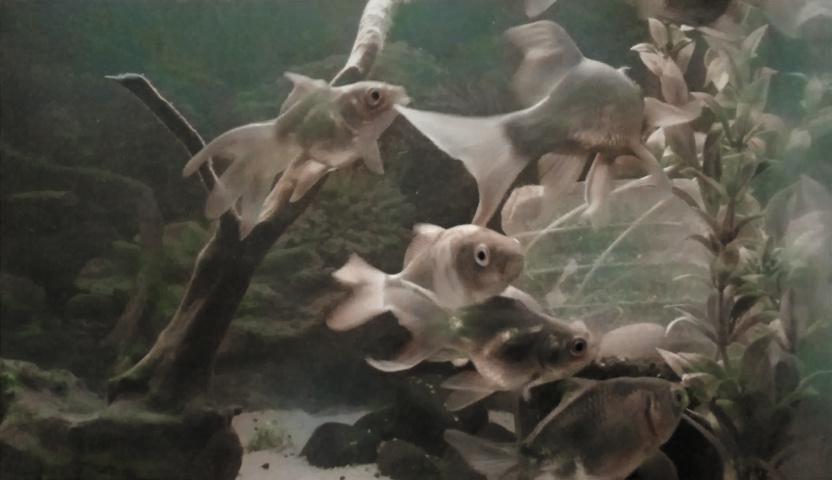}\end{subfigure} & \begin{subfigure}[b]{\myhighreswidth\textwidth}\includegraphics[width=\textwidth]{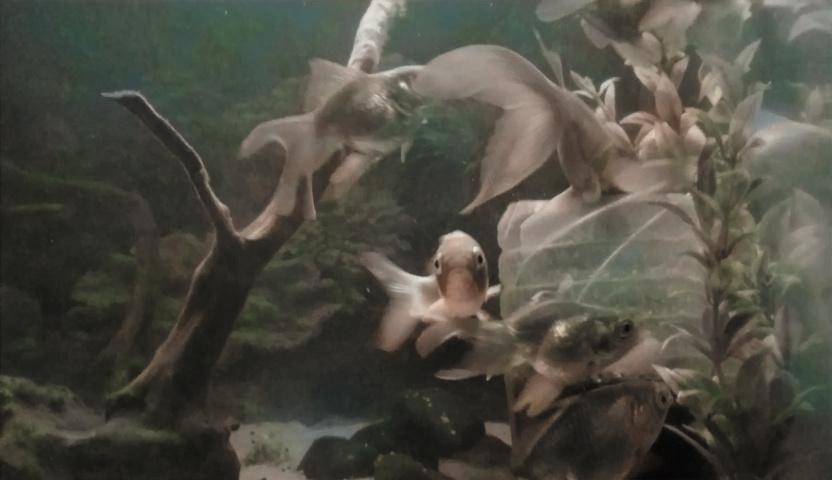}\end{subfigure} & \begin{subfigure}[b]{\myhighreswidth\textwidth}\includegraphics[width=\textwidth]{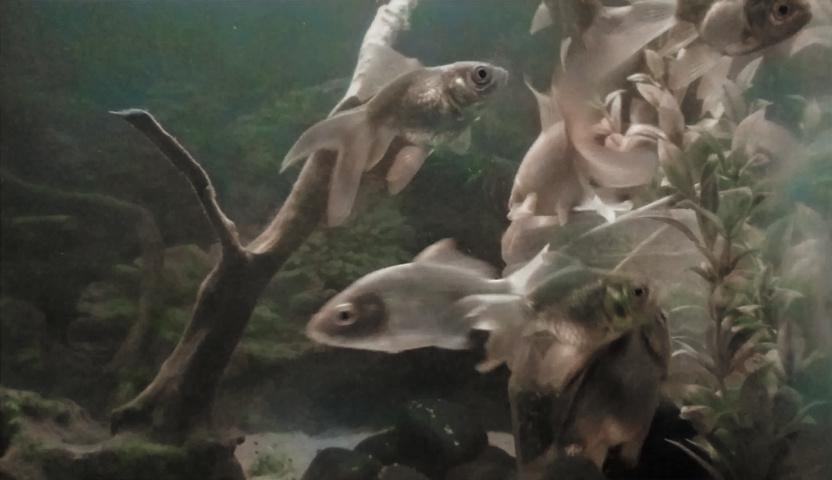}\end{subfigure} & \begin{subfigure}[b]{\myhighreswidth\textwidth}\includegraphics[width=\textwidth]{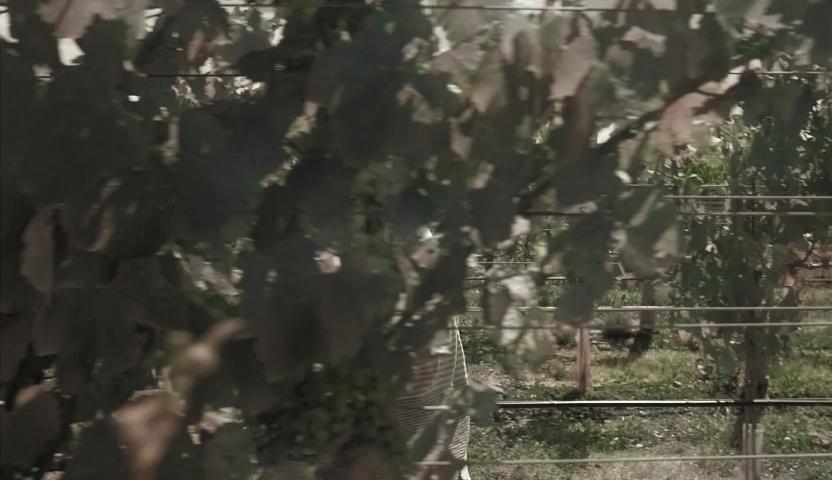}\end{subfigure} & \begin{subfigure}[b]{\myhighreswidth\textwidth}\includegraphics[width=\textwidth]{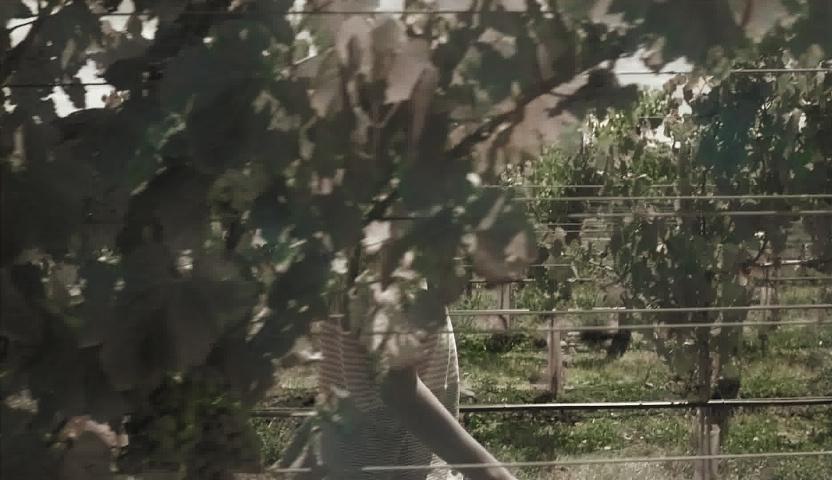}\end{subfigure} & \begin{subfigure}[b]{\myhighreswidth\textwidth}\includegraphics[width=\textwidth]{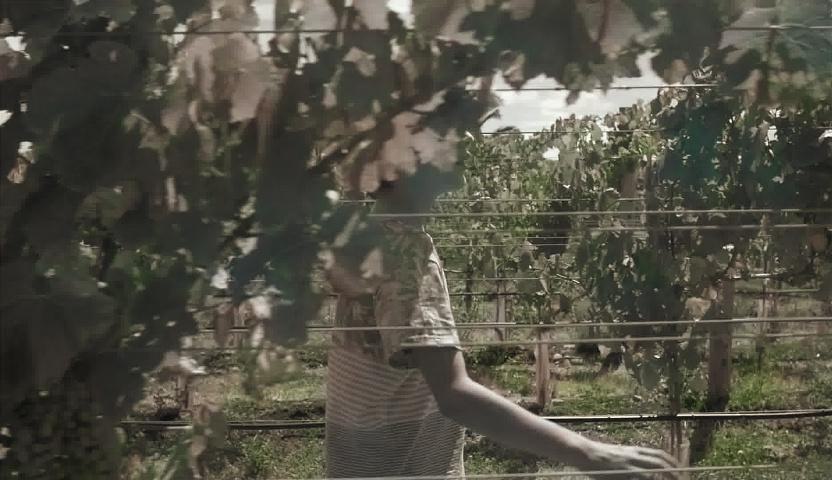}\end{subfigure} & \begin{subfigure}[b]{\myhighreswidth\textwidth}\includegraphics[width=\textwidth]{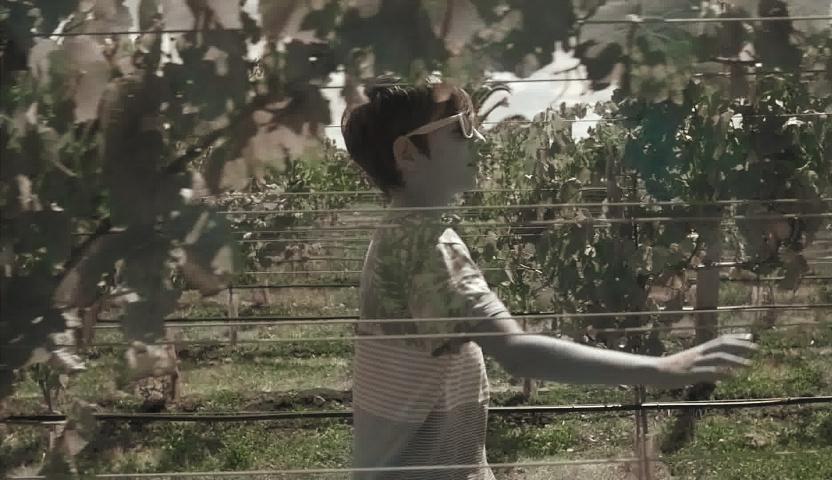}\end{subfigure}\\
\begin{sideways}{\tiny Deoldify}\end{sideways} & \begin{subfigure}[b]{\myhighreswidth\textwidth}\includegraphics[width=\textwidth]{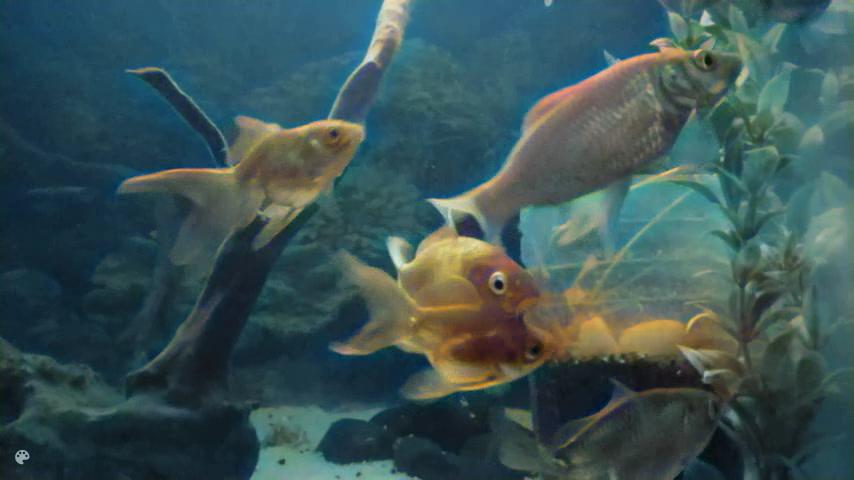}\end{subfigure} & \begin{subfigure}[b]{\myhighreswidth\textwidth}\includegraphics[width=\textwidth]{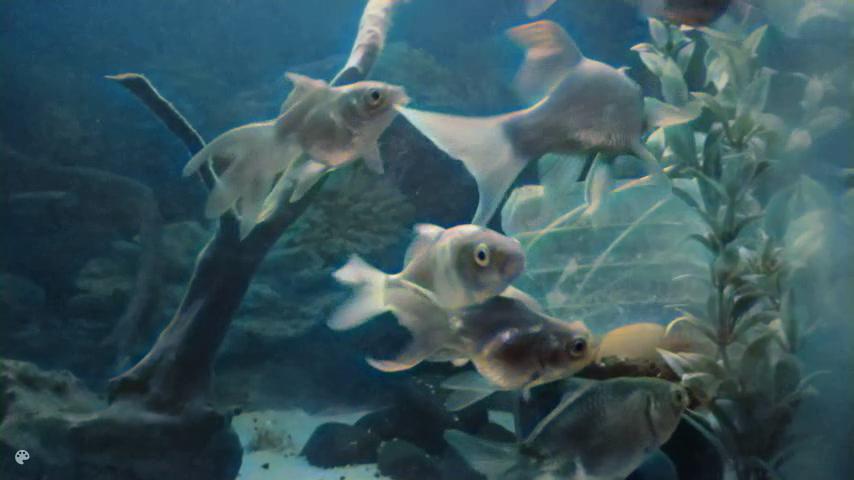}\end{subfigure} & \begin{subfigure}[b]{\myhighreswidth\textwidth}\includegraphics[width=\textwidth]{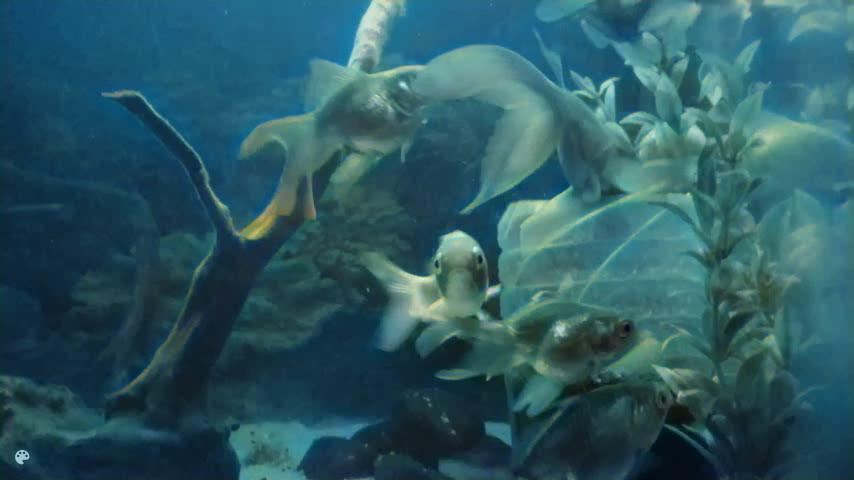}\end{subfigure} & \begin{subfigure}[b]{\myhighreswidth\textwidth}\includegraphics[width=\textwidth]{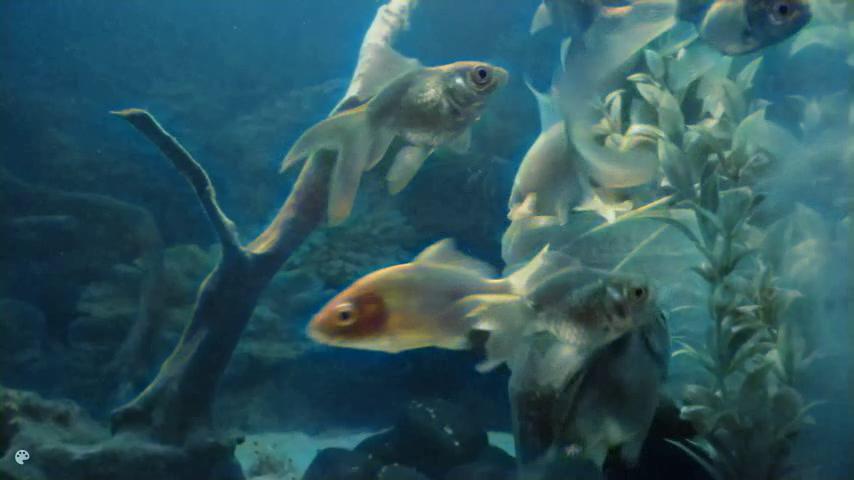}\end{subfigure} & \begin{subfigure}[b]{\myhighreswidth\textwidth}\includegraphics[width=\textwidth]{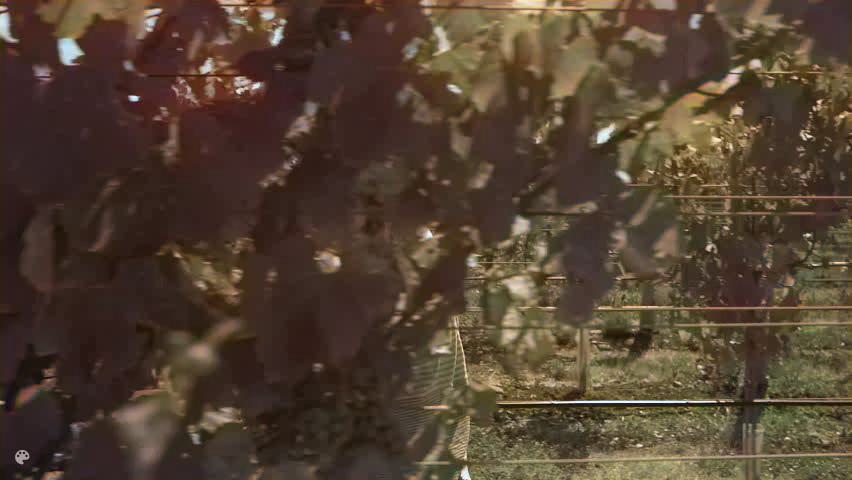}\end{subfigure} & \begin{subfigure}[b]{\myhighreswidth\textwidth}\includegraphics[width=\textwidth]{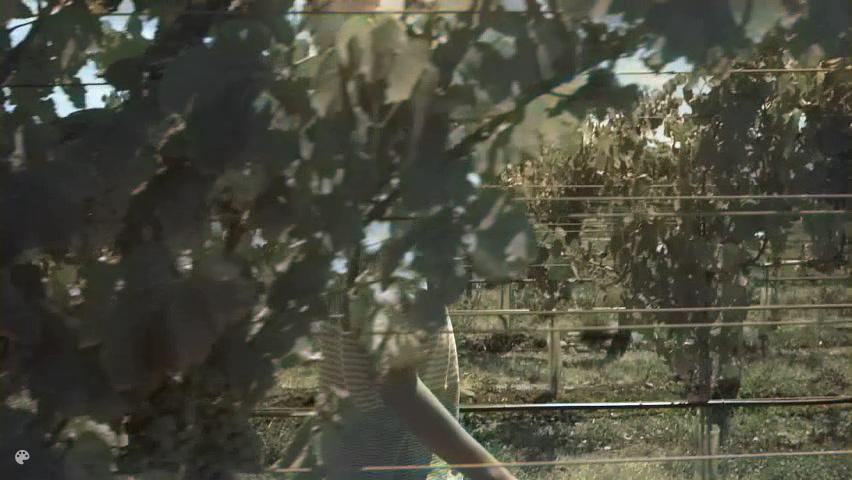}\end{subfigure} & \begin{subfigure}[b]{\myhighreswidth\textwidth}\includegraphics[width=\textwidth]{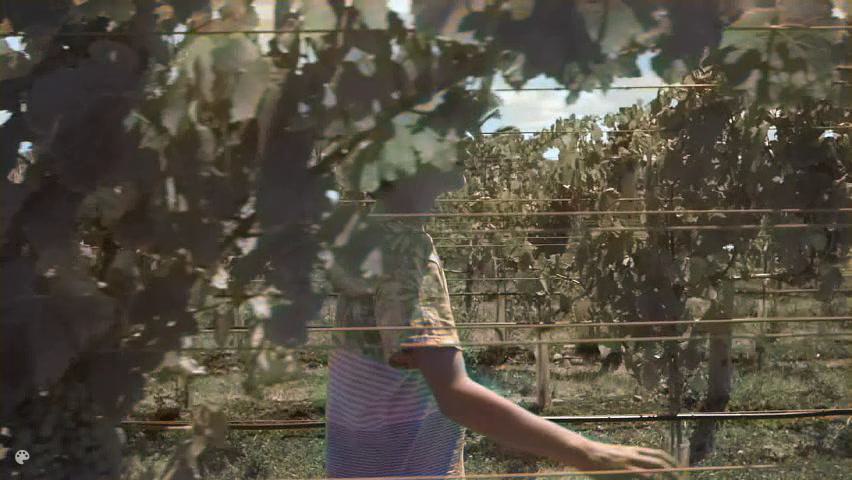}\end{subfigure} & \begin{subfigure}[b]{\myhighreswidth\textwidth}\includegraphics[width=\textwidth]{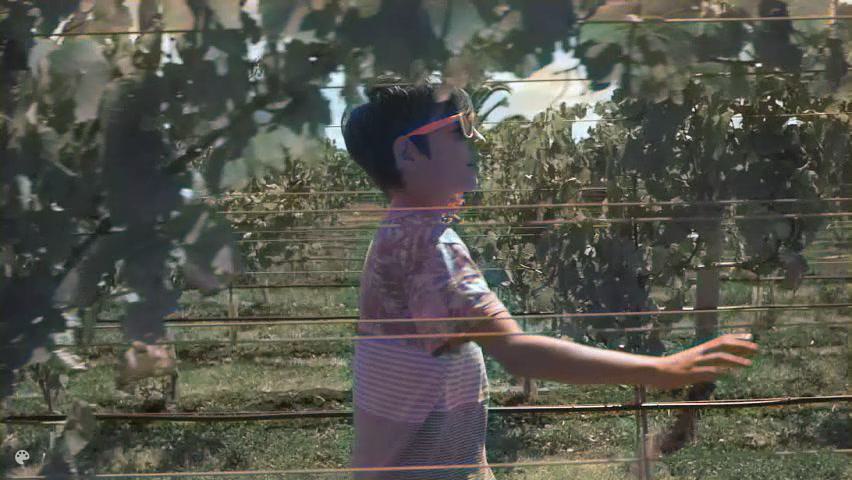}\end{subfigure}\\
\begin{sideways}{\tiny DeepExemplar}\end{sideways} & \begin{subfigure}[b]{\myhighreswidth\textwidth}\includegraphics[width=\textwidth]{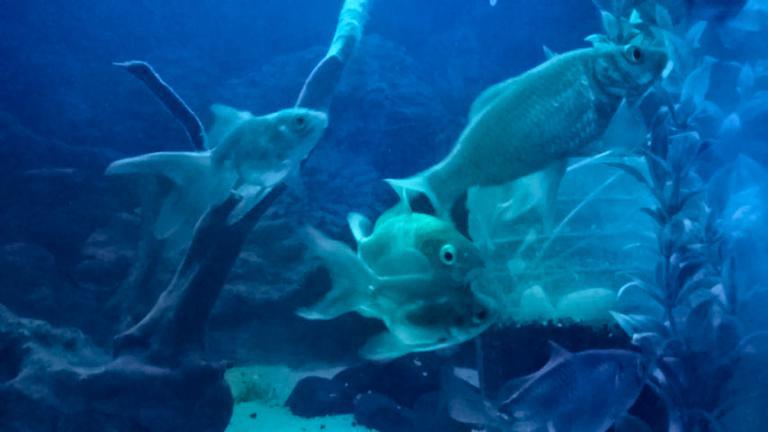}\end{subfigure} & \begin{subfigure}[b]{\myhighreswidth\textwidth}\includegraphics[width=\textwidth]{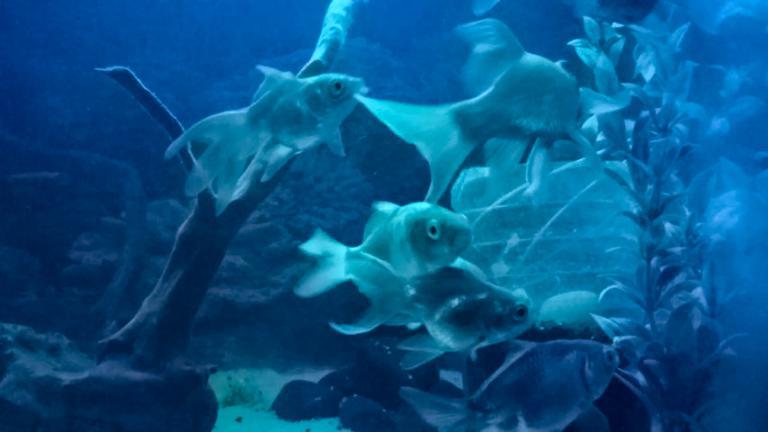}\end{subfigure} & \begin{subfigure}[b]{\myhighreswidth\textwidth}\includegraphics[width=\textwidth]{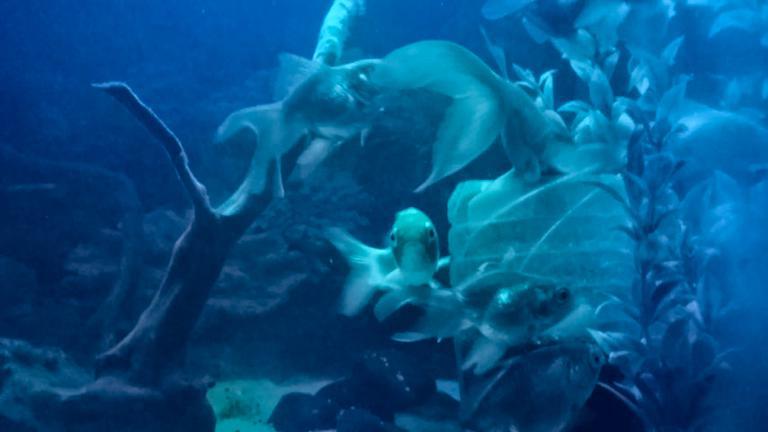}\end{subfigure} & \begin{subfigure}[b]{\myhighreswidth\textwidth}\includegraphics[width=\textwidth]{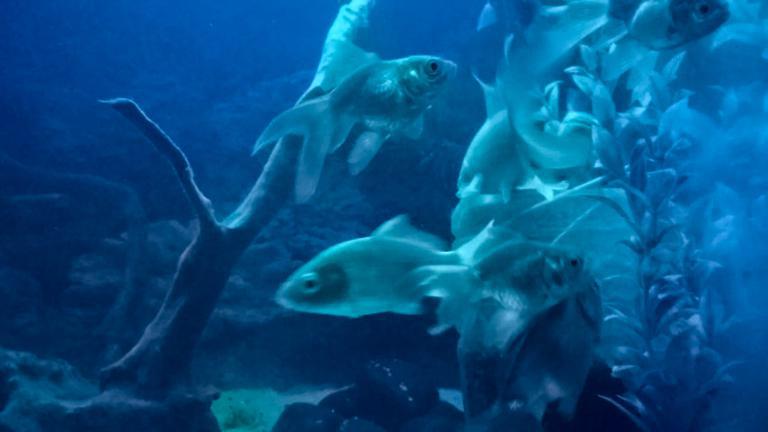}\end{subfigure} & \begin{subfigure}[b]{\myhighreswidth\textwidth}\includegraphics[width=\textwidth]{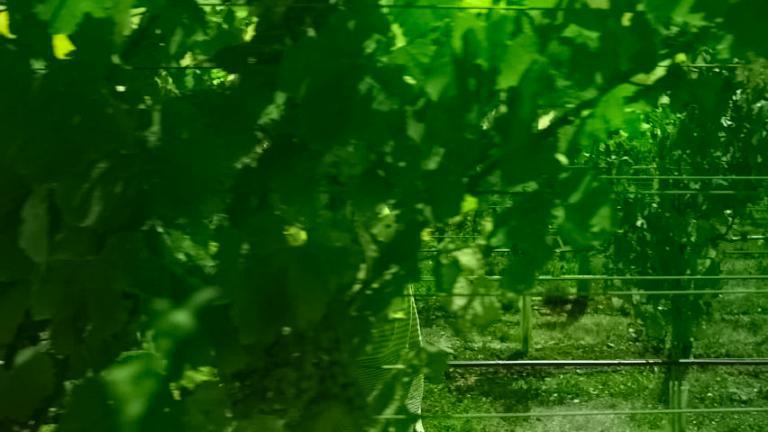}\end{subfigure} & \begin{subfigure}[b]{\myhighreswidth\textwidth}\includegraphics[width=\textwidth]{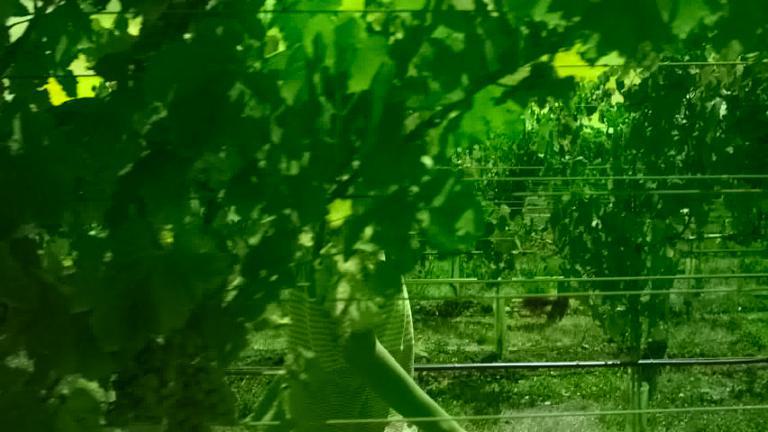}\end{subfigure} & \begin{subfigure}[b]{\myhighreswidth\textwidth}\includegraphics[width=\textwidth]{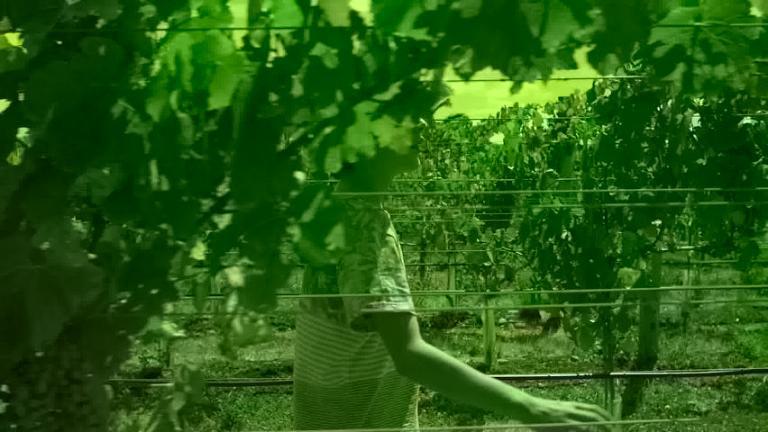}\end{subfigure} & \begin{subfigure}[b]{\myhighreswidth\textwidth}\includegraphics[width=\textwidth]{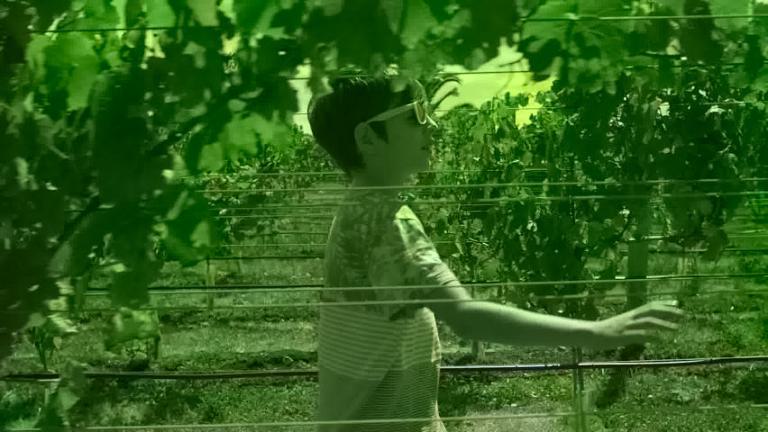}\end{subfigure}\\
\begin{sideways}{\tiny DeepRemaster}\end{sideways} & \begin{subfigure}[b]{\myhighreswidth\textwidth}\includegraphics[width=\textwidth]{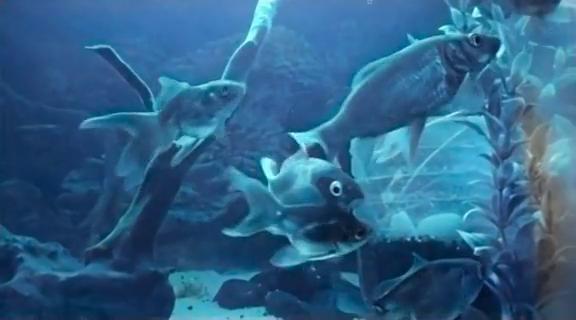}\end{subfigure} & \begin{subfigure}[b]{\myhighreswidth\textwidth}\includegraphics[width=\textwidth]{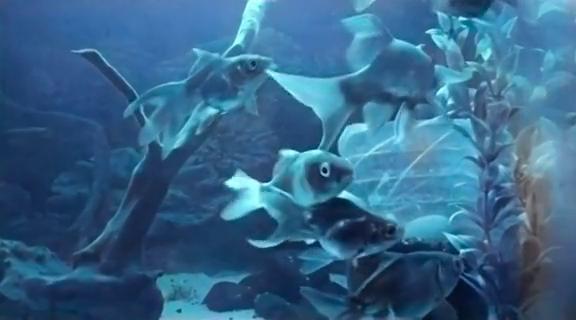}\end{subfigure} & \begin{subfigure}[b]{\myhighreswidth\textwidth}\includegraphics[width=\textwidth]{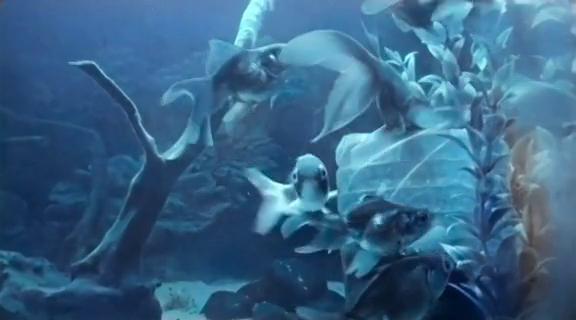}\end{subfigure} & \begin{subfigure}[b]{\myhighreswidth\textwidth}\includegraphics[width=\textwidth]{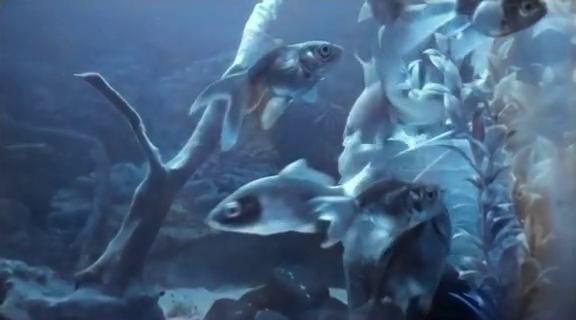}\end{subfigure} & \begin{subfigure}[b]{\myhighreswidth\textwidth}\includegraphics[width=\textwidth]{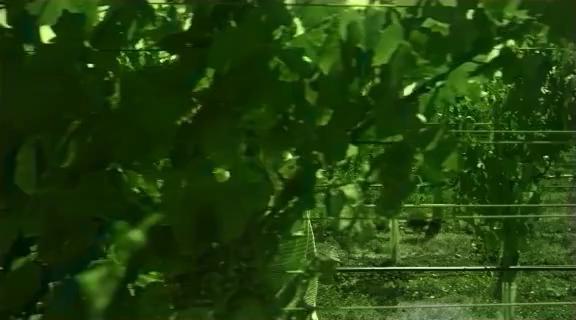}\end{subfigure} & \begin{subfigure}[b]{\myhighreswidth\textwidth}\includegraphics[width=\textwidth]{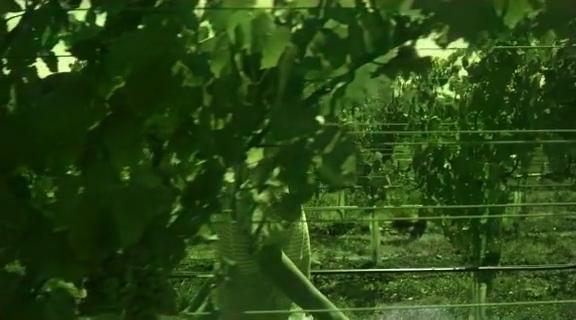}\end{subfigure} & \begin{subfigure}[b]{\myhighreswidth\textwidth}\includegraphics[width=\textwidth]{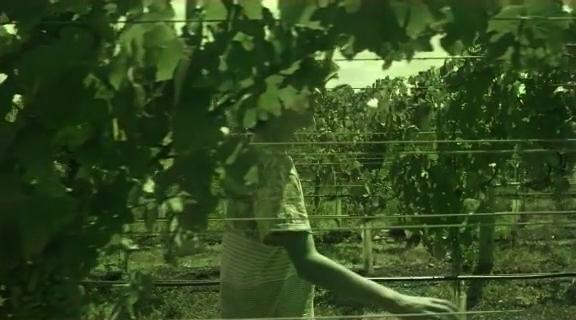}\end{subfigure} & \begin{subfigure}[b]{\myhighreswidth\textwidth}\includegraphics[width=\textwidth]{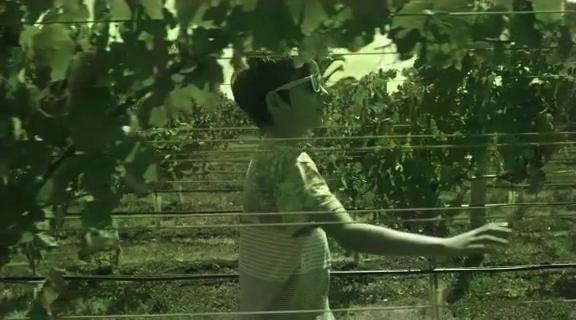}\end{subfigure}\\
\begin{sideways}{\tiny TCVC}\end{sideways} & \begin{subfigure}[b]{\myhighreswidth\textwidth}\includegraphics[width=\textwidth]{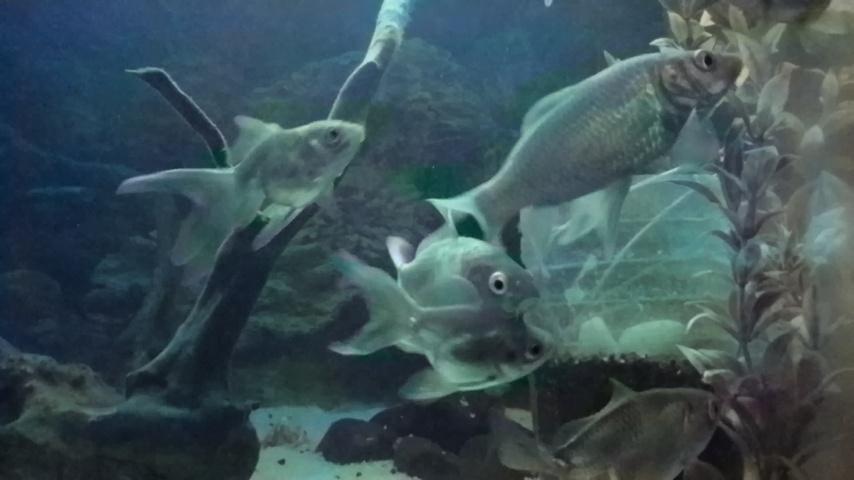}\end{subfigure} & \begin{subfigure}[b]{\myhighreswidth\textwidth}\includegraphics[width=\textwidth]{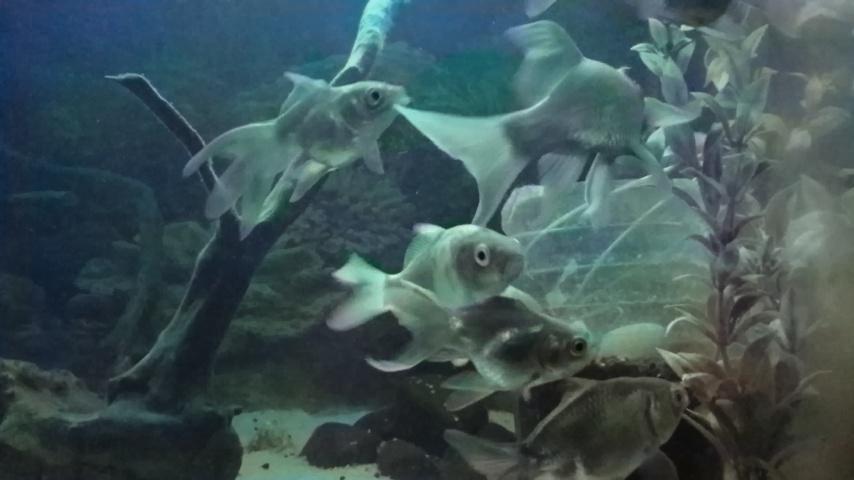}\end{subfigure} & \begin{subfigure}[b]{\myhighreswidth\textwidth}\includegraphics[width=\textwidth]{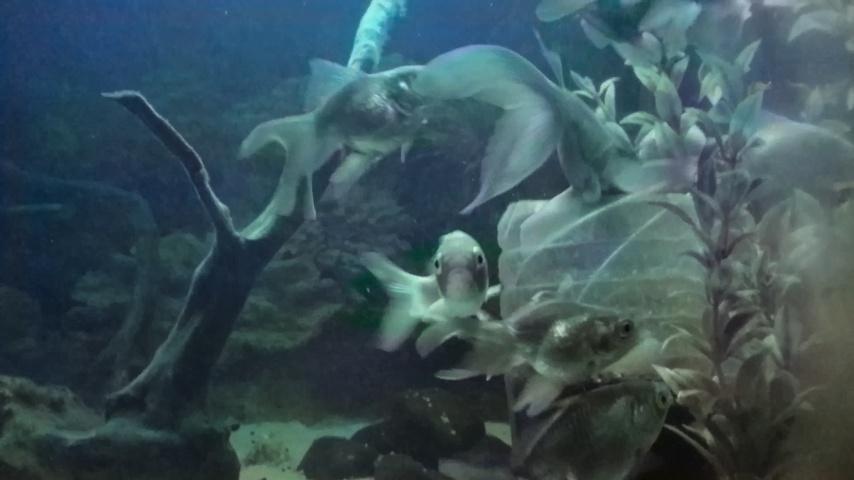}\end{subfigure} & \begin{subfigure}[b]{\myhighreswidth\textwidth}\includegraphics[width=\textwidth]{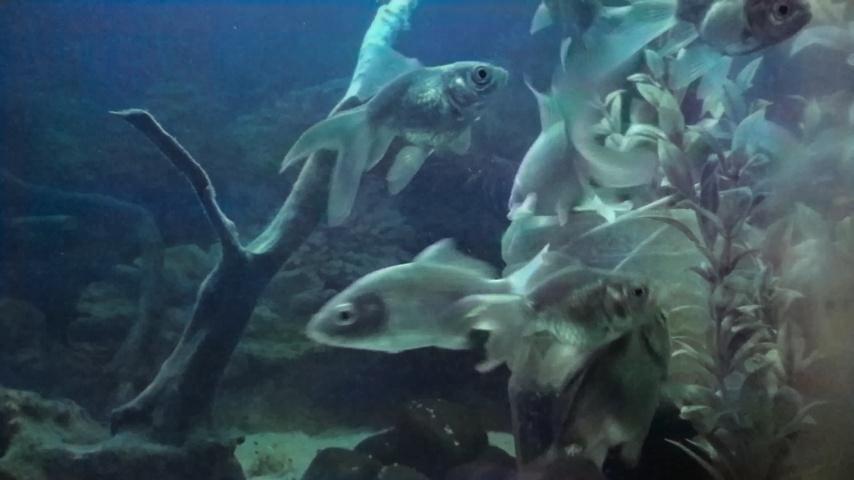}\end{subfigure} & \begin{subfigure}[b]{\myhighreswidth\textwidth}\includegraphics[width=\textwidth]{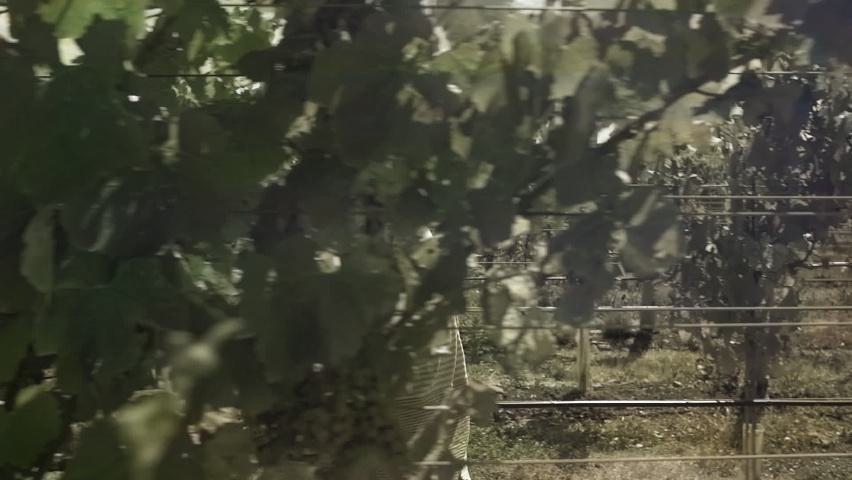}\end{subfigure} & \begin{subfigure}[b]{\myhighreswidth\textwidth}\includegraphics[width=\textwidth]{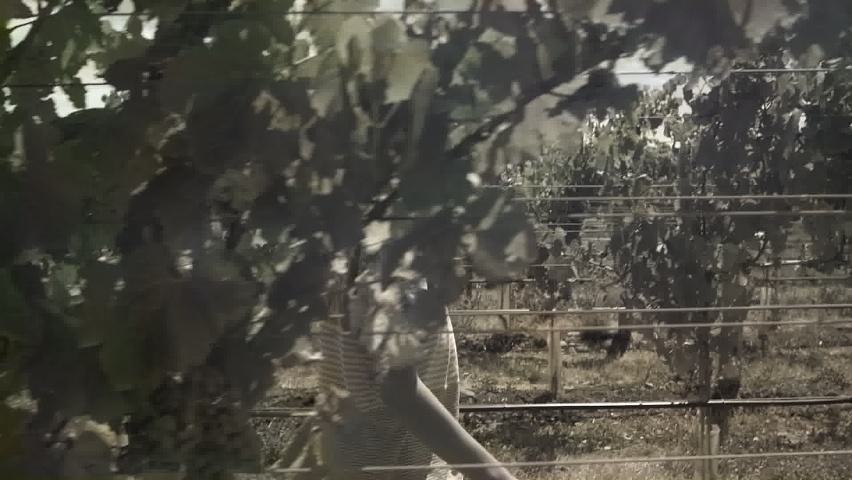}\end{subfigure} & \begin{subfigure}[b]{\myhighreswidth\textwidth}\includegraphics[width=\textwidth]{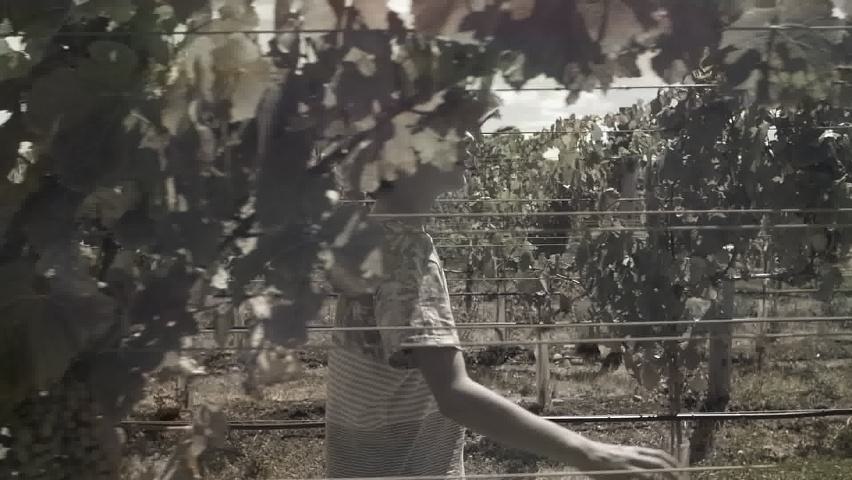}\end{subfigure} & \begin{subfigure}[b]{\myhighreswidth\textwidth}\includegraphics[width=\textwidth]{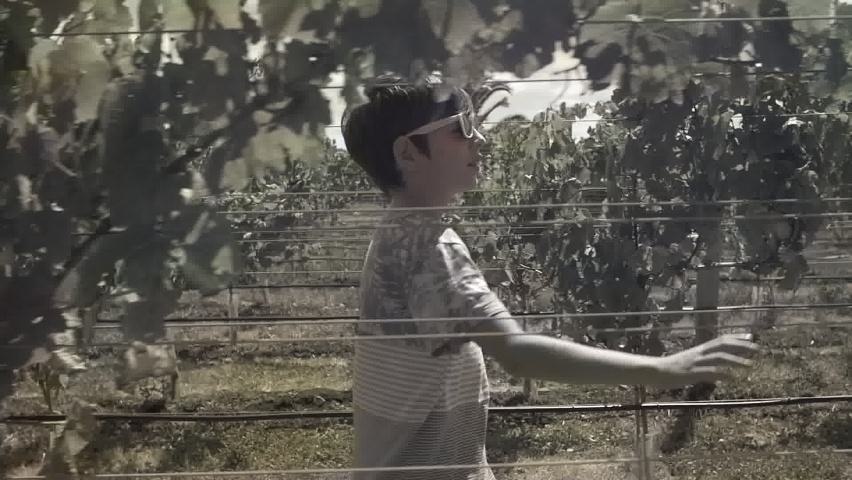}\end{subfigure}\\
\begin{sideways}{\tiny VCGAN}\end{sideways} & \begin{subfigure}[b]{\myhighreswidth\textwidth}\includegraphics[width=\textwidth]{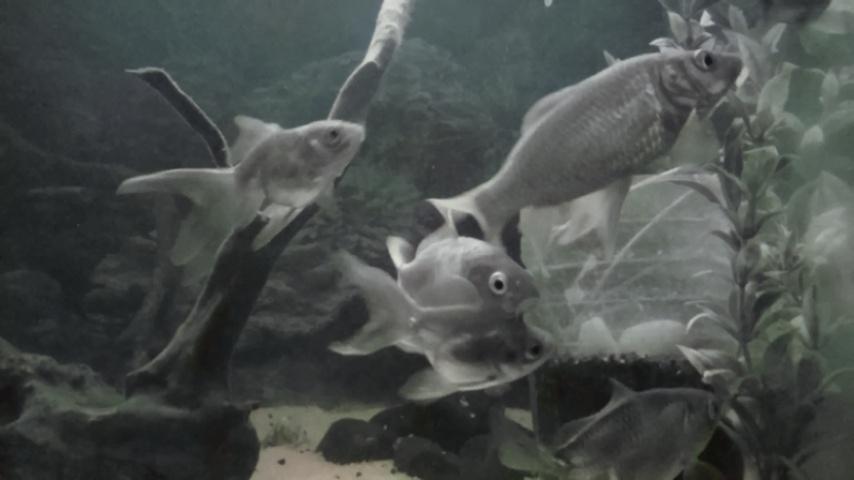}\end{subfigure} & \begin{subfigure}[b]{\myhighreswidth\textwidth}\includegraphics[width=\textwidth]{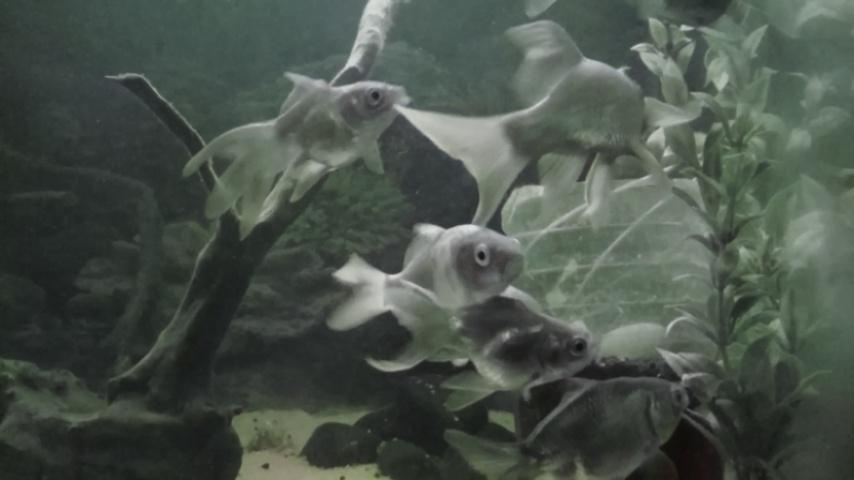}\end{subfigure} & \begin{subfigure}[b]{\myhighreswidth\textwidth}\includegraphics[width=\textwidth]{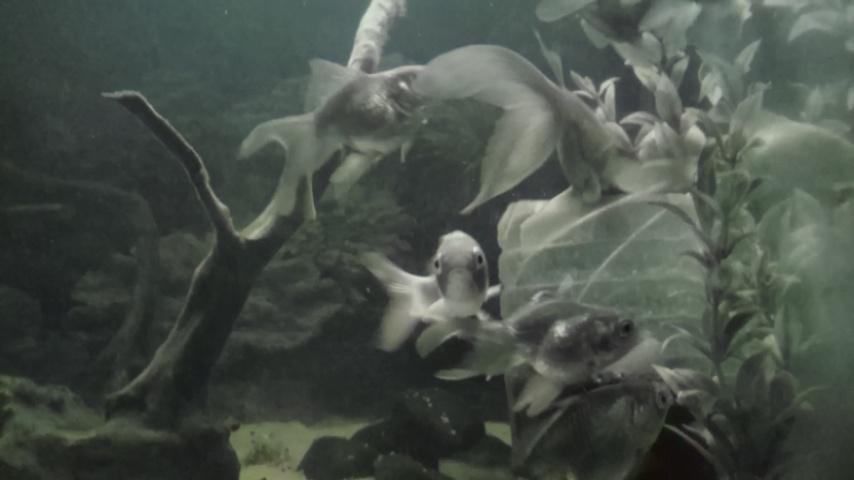}\end{subfigure} & \begin{subfigure}[b]{\myhighreswidth\textwidth}\includegraphics[width=\textwidth]{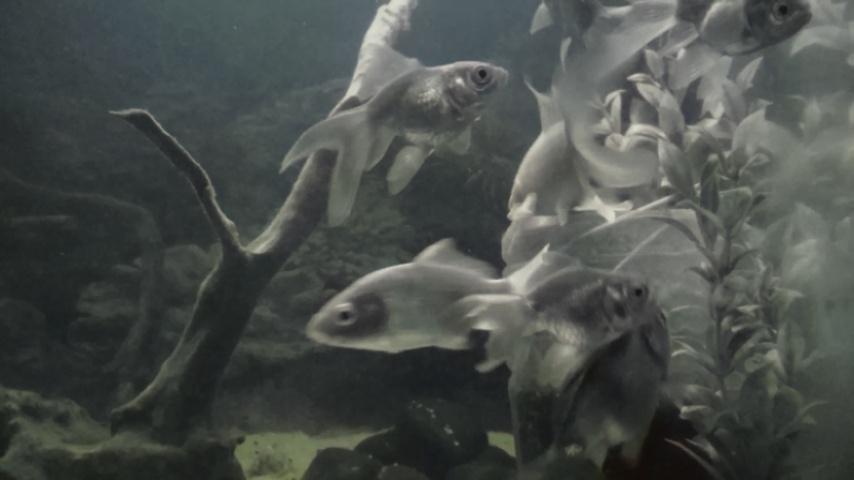}\end{subfigure} & \begin{subfigure}[b]{\myhighreswidth\textwidth}\includegraphics[width=\textwidth]{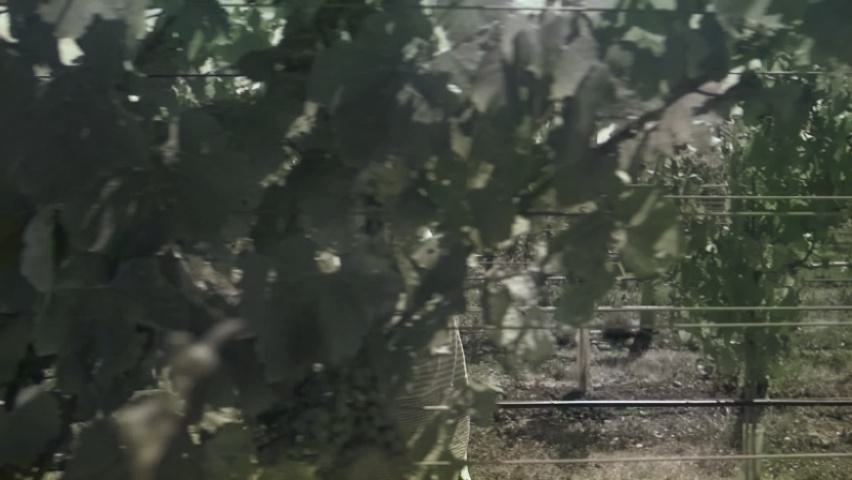}\end{subfigure} & \begin{subfigure}[b]{\myhighreswidth\textwidth}\includegraphics[width=\textwidth]{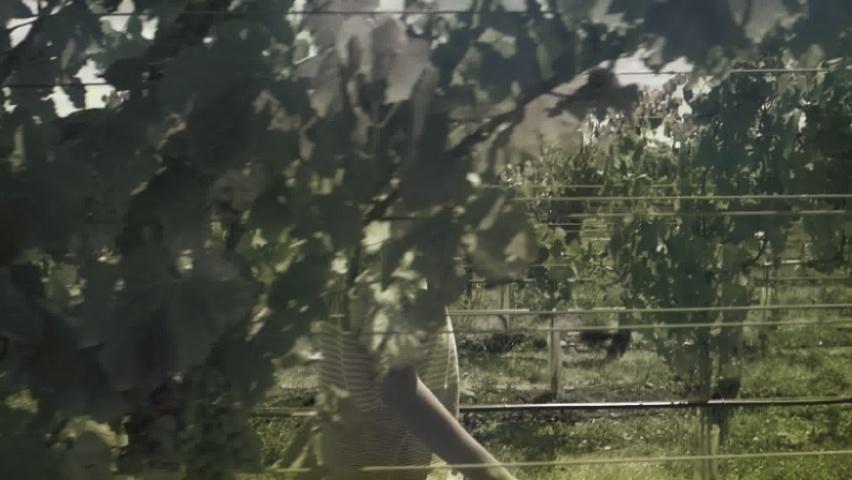}\end{subfigure} & \begin{subfigure}[b]{\myhighreswidth\textwidth}\includegraphics[width=\textwidth]{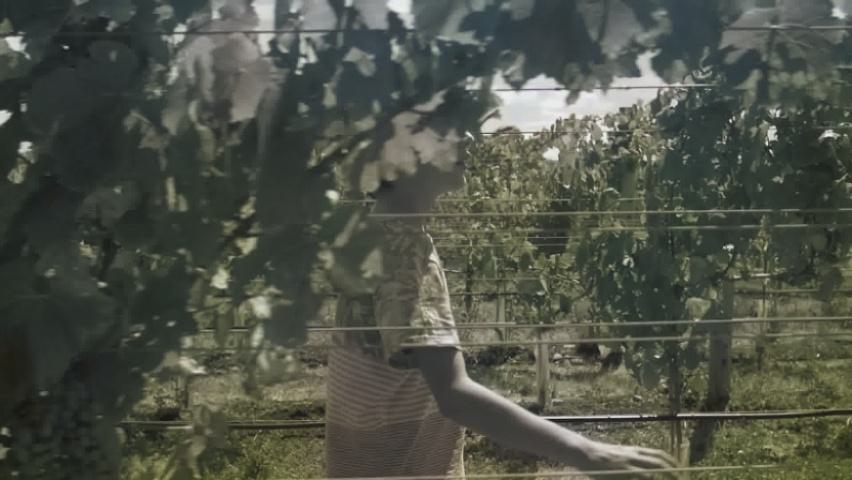}\end{subfigure} & \begin{subfigure}[b]{\myhighreswidth\textwidth}\includegraphics[width=\textwidth]{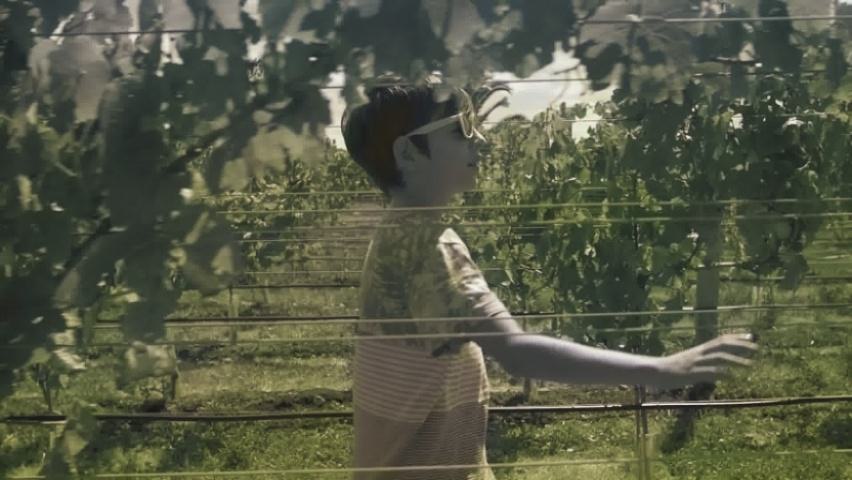}\end{subfigure}\\
\begin{sideways}{\tiny Ours}\end{sideways} & \begin{subfigure}[b]{\myhighreswidth\textwidth}\includegraphics[width=\textwidth]{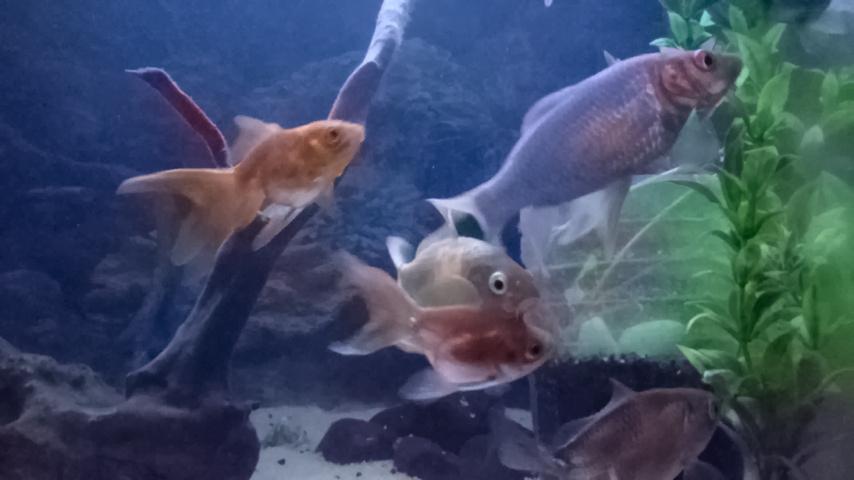}\end{subfigure} & \begin{subfigure}[b]{\myhighreswidth\textwidth}\includegraphics[width=\textwidth]{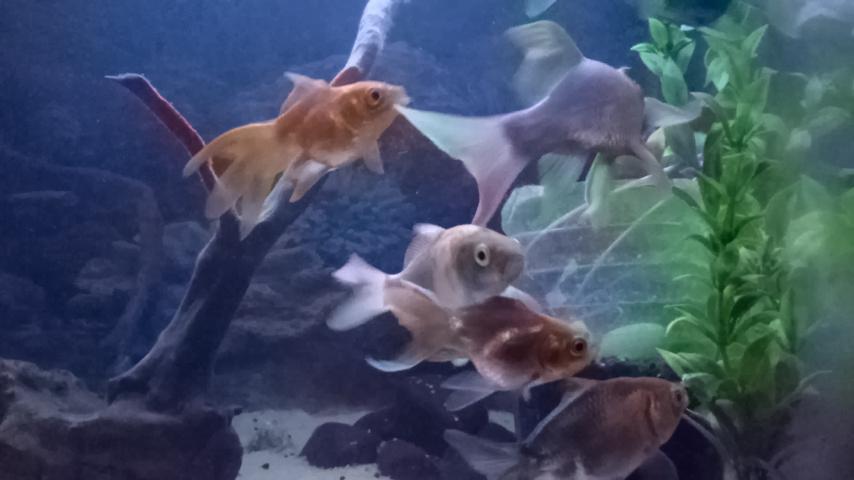}\end{subfigure} & \begin{subfigure}[b]{\myhighreswidth\textwidth}\includegraphics[width=\textwidth]{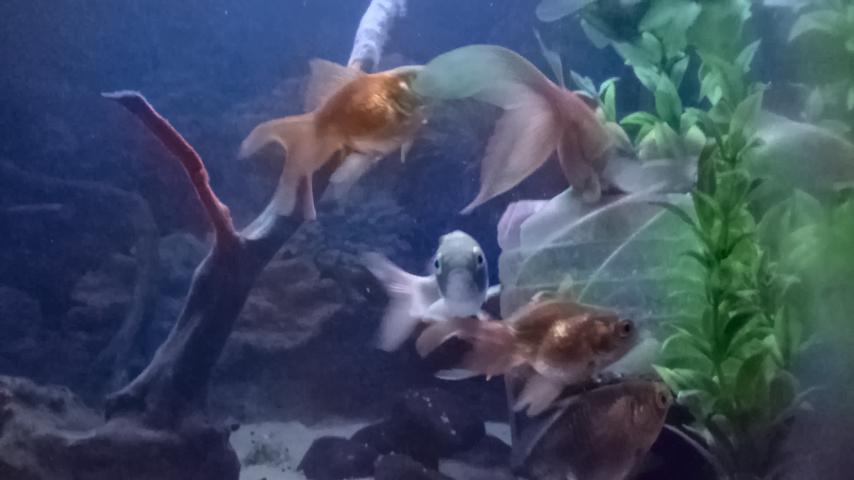}\end{subfigure} & \begin{subfigure}[b]{\myhighreswidth\textwidth}\includegraphics[width=\textwidth]{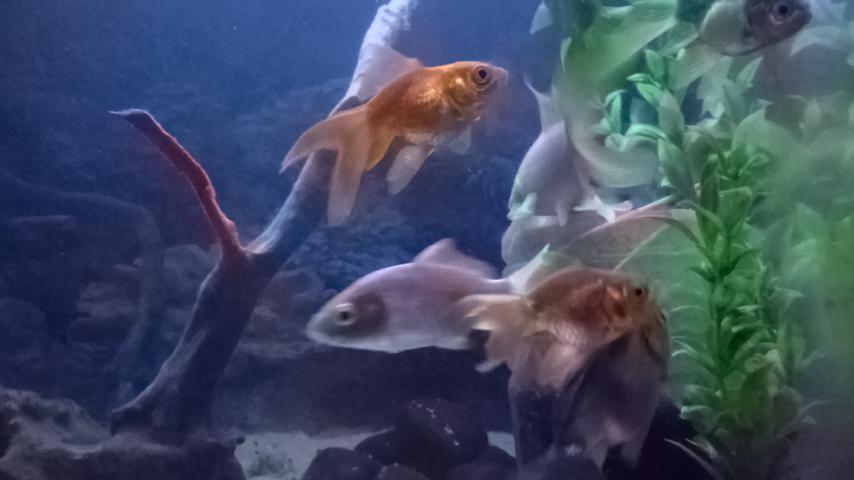}\end{subfigure} & \begin{subfigure}[b]{\myhighreswidth\textwidth}\includegraphics[width=\textwidth]{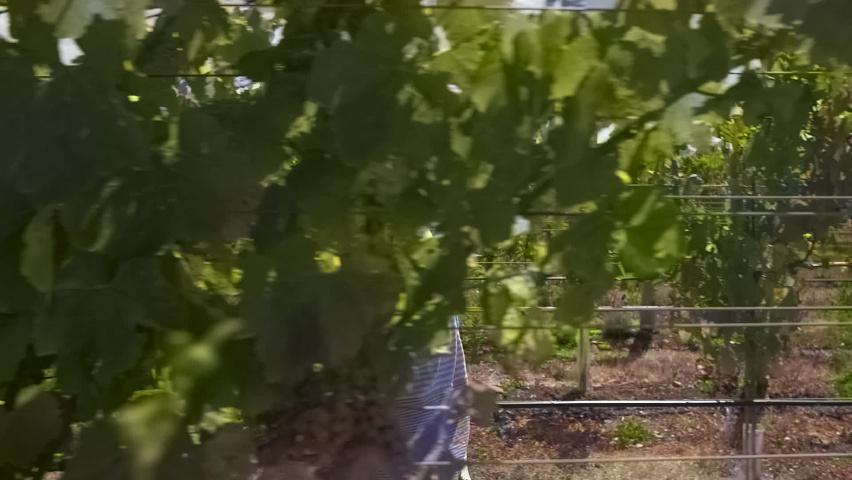}\end{subfigure} & \begin{subfigure}[b]{\myhighreswidth\textwidth}\includegraphics[width=\textwidth]{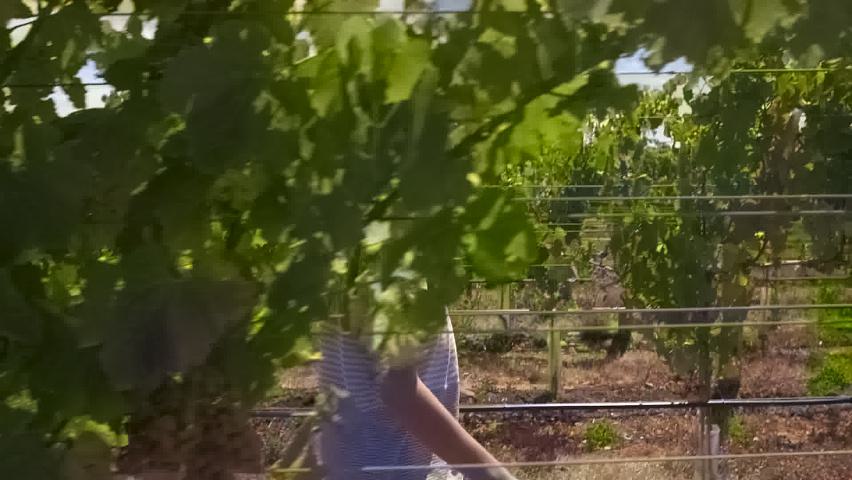}\end{subfigure} & \begin{subfigure}[b]{\myhighreswidth\textwidth}\includegraphics[width=\textwidth]{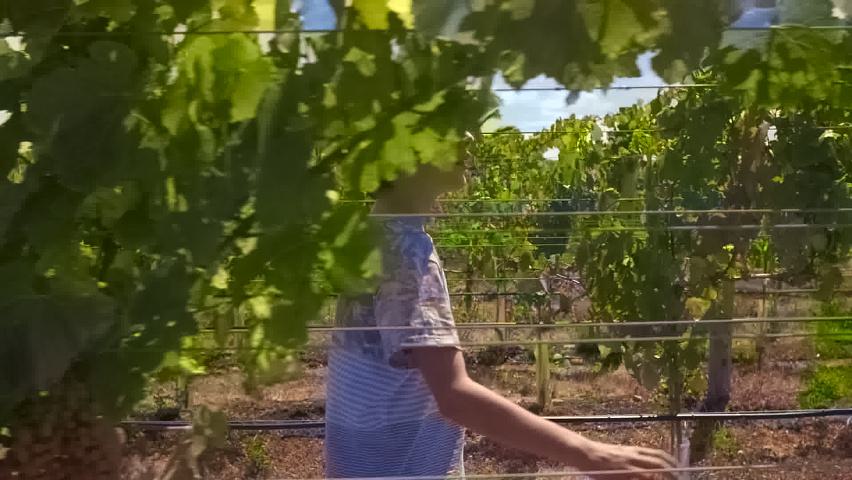}\end{subfigure} & \begin{subfigure}[b]{\myhighreswidth\textwidth}\includegraphics[width=\textwidth]{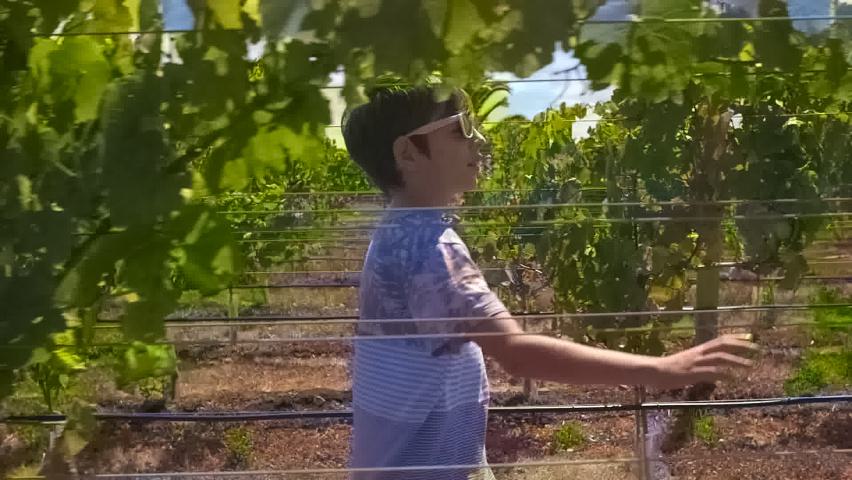}\end{subfigure}\\
& T=0 & T=15 & T=30 & T=45 & T=0 & T=15 & T=30 & T=45\\
\end{tabular}
\renewcommand{\arraystretch}{1.0}

\caption{Qualitative comparison with different video colorization methods.}
\label{fig:qualitative_eval}
\end{figure*}

\begin{figure}[h]
    \centering
    \captionsetup[subfigure]{labelformat=empty}
\def\myhighreswidth{0.12}
\def\myhighresoffset{-0.002}
\centering
\renewcommand{\arraystretch}{0.1}
\setlength{\tabcolsep}{0pt}
\begin{tabular}[c]{ccccc}
\begin{subfigure}[b]{\myhighreswidth\textwidth}\includegraphics[width=\textwidth]{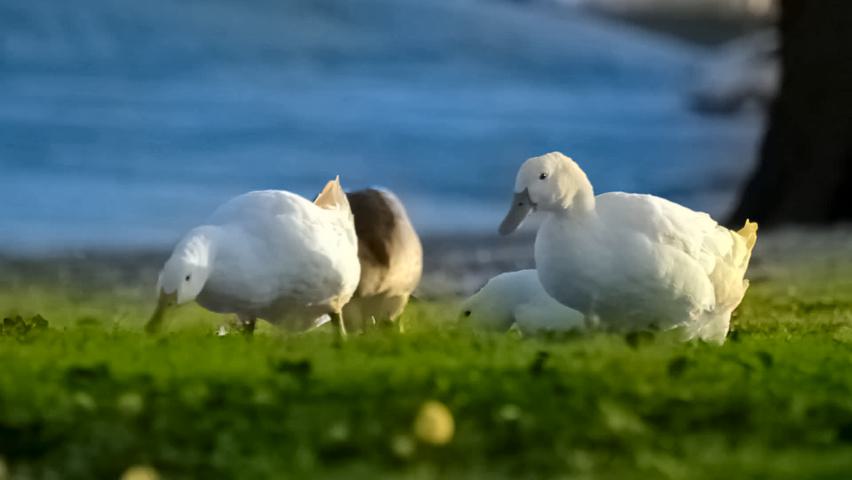}\end{subfigure} & \begin{subfigure}[b]{\myhighreswidth\textwidth}\includegraphics[width=\textwidth]{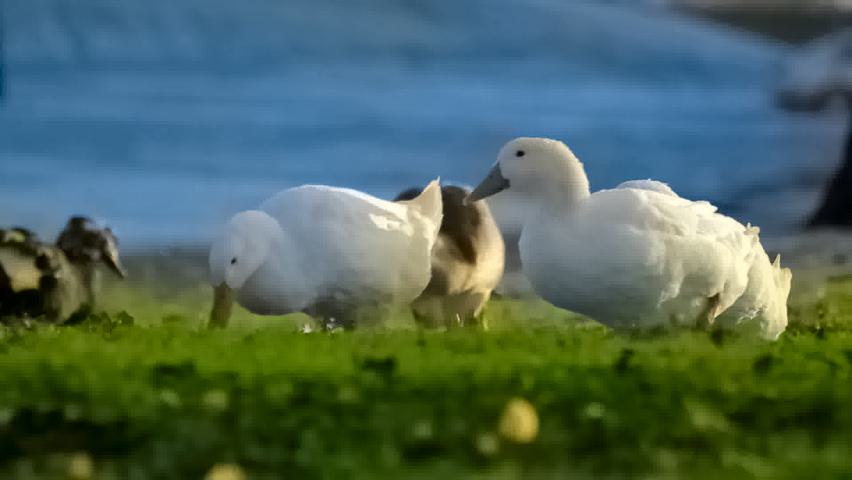}\end{subfigure} & \begin{subfigure}[b]{\myhighreswidth\textwidth}\includegraphics[width=\textwidth]{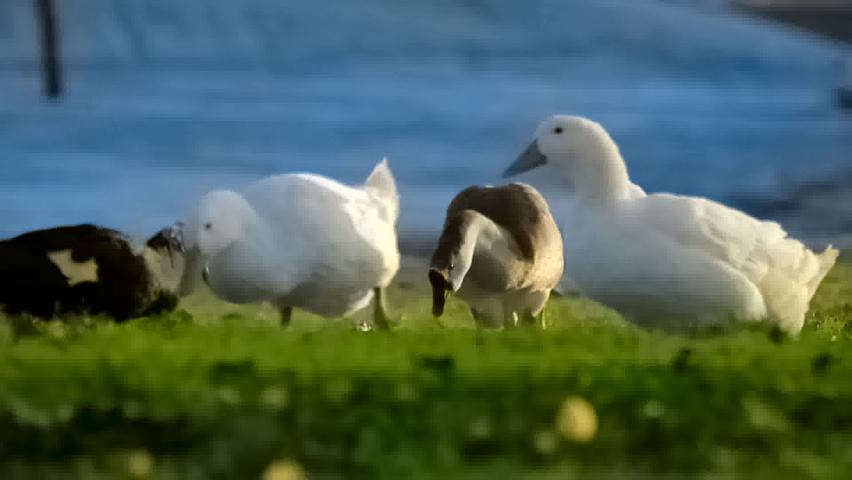}\end{subfigure} & \begin{subfigure}[b]{\myhighreswidth\textwidth}\includegraphics[width=\textwidth]{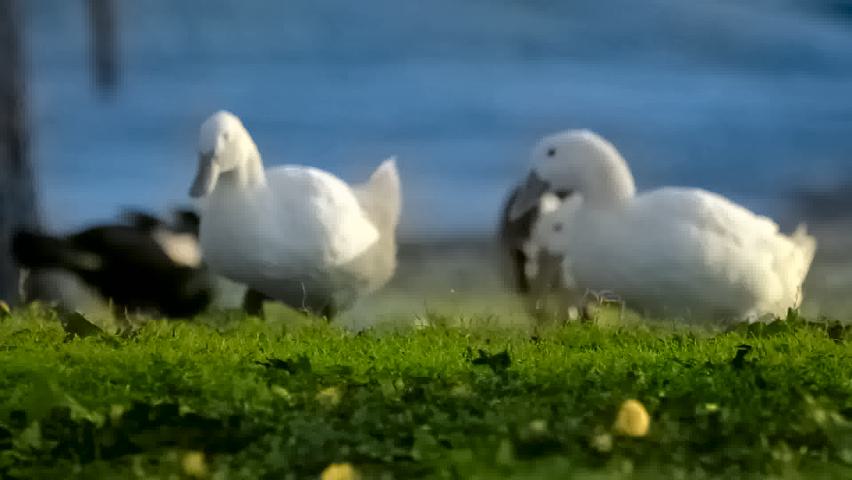}\end{subfigure}\\
\begin{subfigure}[b]{\myhighreswidth\textwidth}\includegraphics[width=\textwidth]{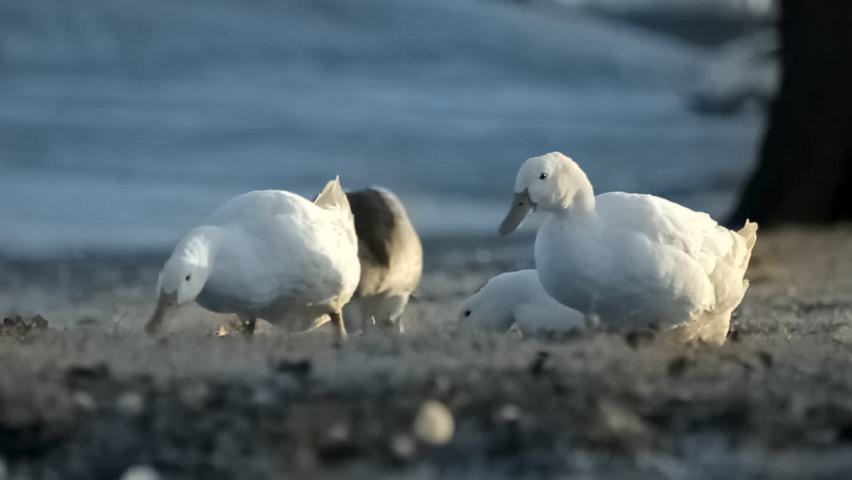}\end{subfigure} & \begin{subfigure}[b]{\myhighreswidth\textwidth}\includegraphics[width=\textwidth]{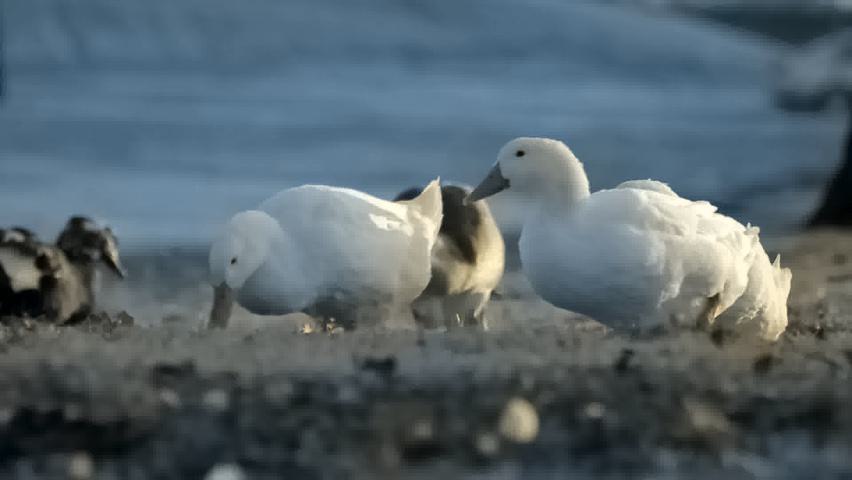}\end{subfigure} & \begin{subfigure}[b]{\myhighreswidth\textwidth}\includegraphics[width=\textwidth]{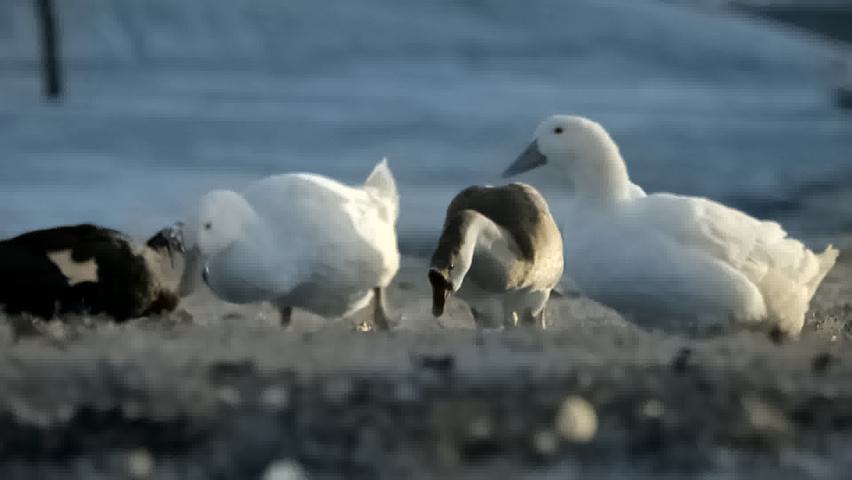}\end{subfigure} & \begin{subfigure}[b]{\myhighreswidth\textwidth}\includegraphics[width=\textwidth]{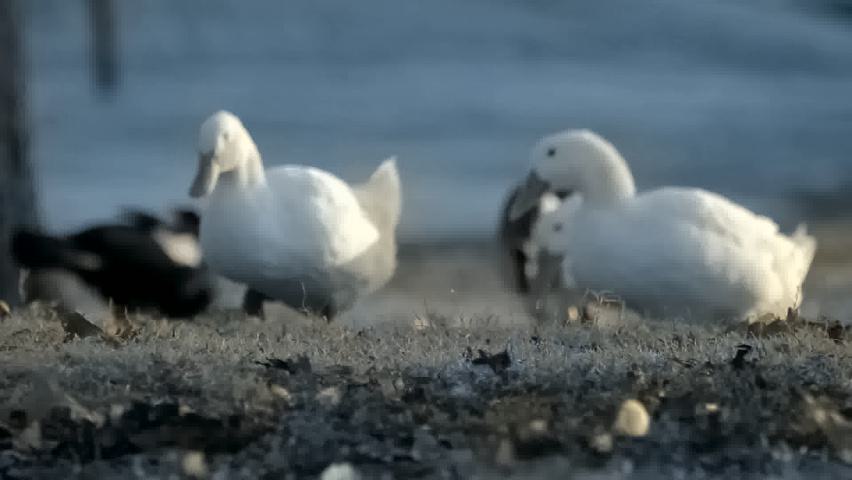}\end{subfigure}\\
\begin{subfigure}[b]{\myhighreswidth\textwidth}\includegraphics[width=\textwidth]{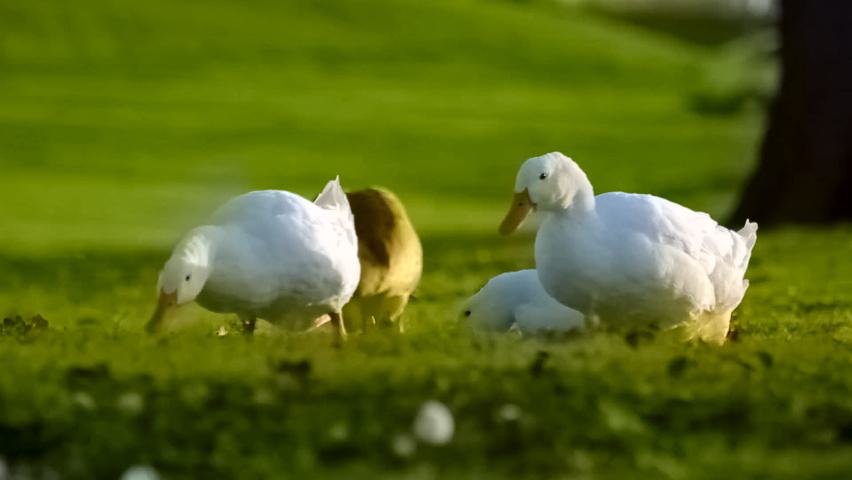}\end{subfigure} & \begin{subfigure}[b]{\myhighreswidth\textwidth}\includegraphics[width=\textwidth]{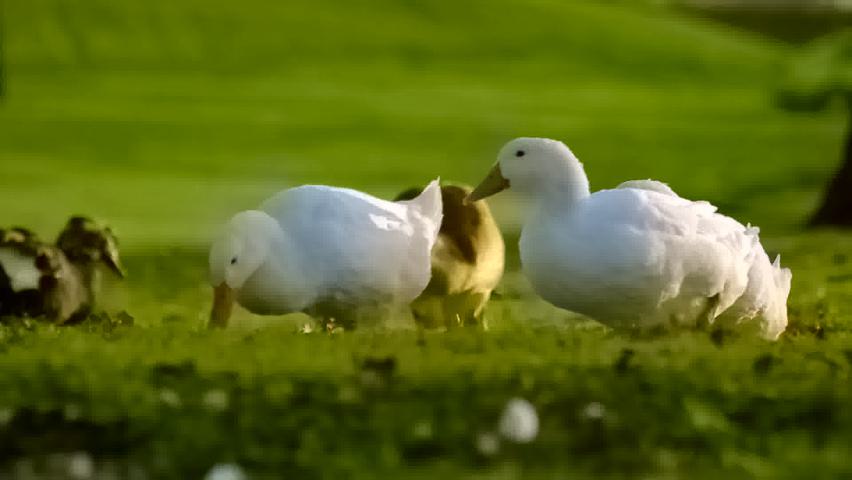}\end{subfigure} & \begin{subfigure}[b]{\myhighreswidth\textwidth}\includegraphics[width=\textwidth]{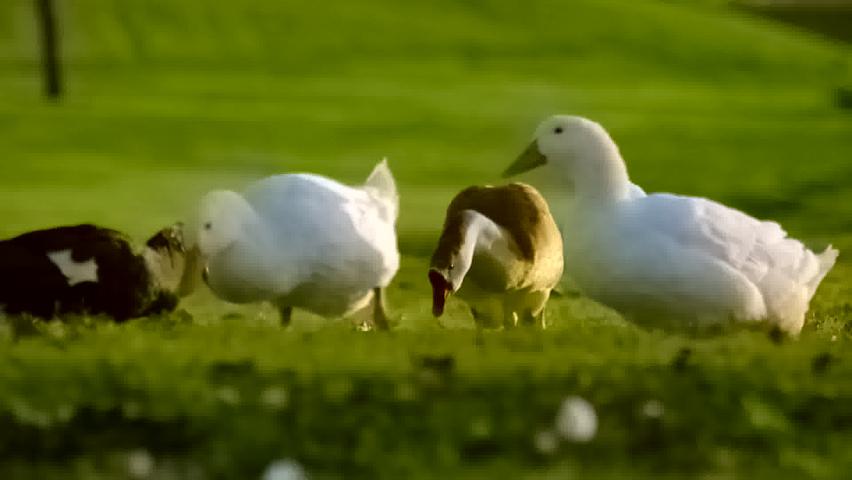}\end{subfigure} & \begin{subfigure}[b]{\myhighreswidth\textwidth}\includegraphics[width=\textwidth]{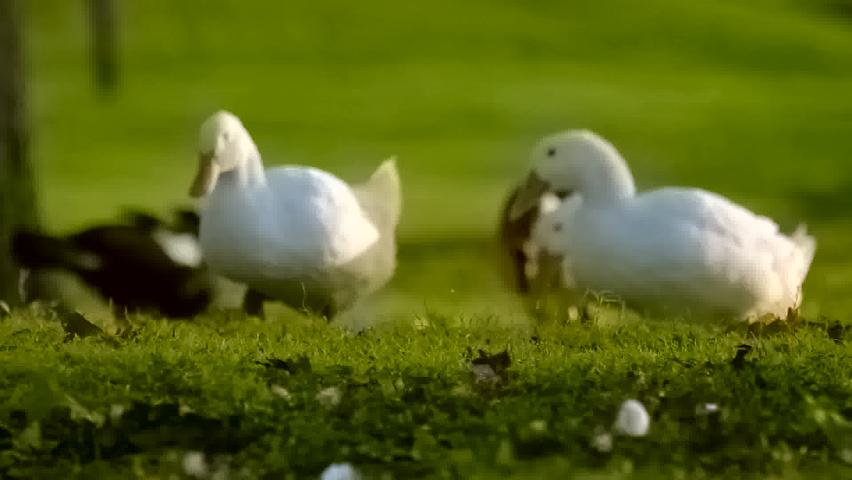}\end{subfigure}\\
&&&\\
\begin{subfigure}[b]{\myhighreswidth\textwidth}\includegraphics[width=\textwidth]{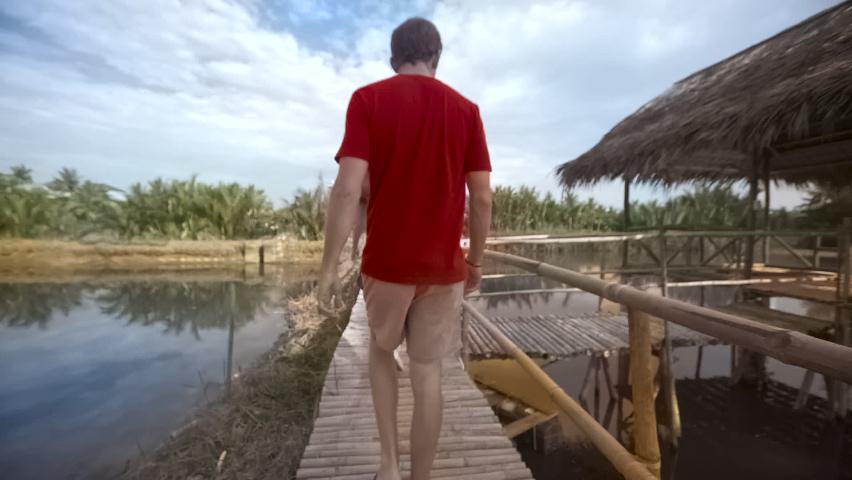}\end{subfigure} & \begin{subfigure}[b]{\myhighreswidth\textwidth}\includegraphics[width=\textwidth]{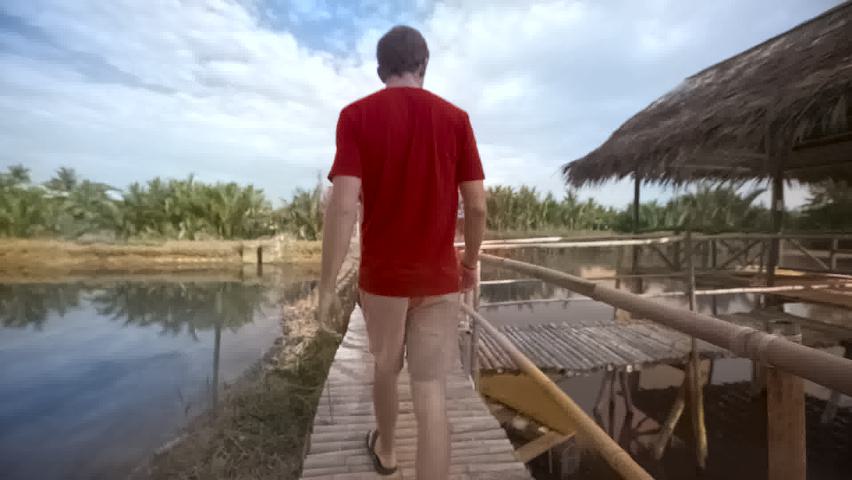}\end{subfigure} & \begin{subfigure}[b]{\myhighreswidth\textwidth}\includegraphics[width=\textwidth]{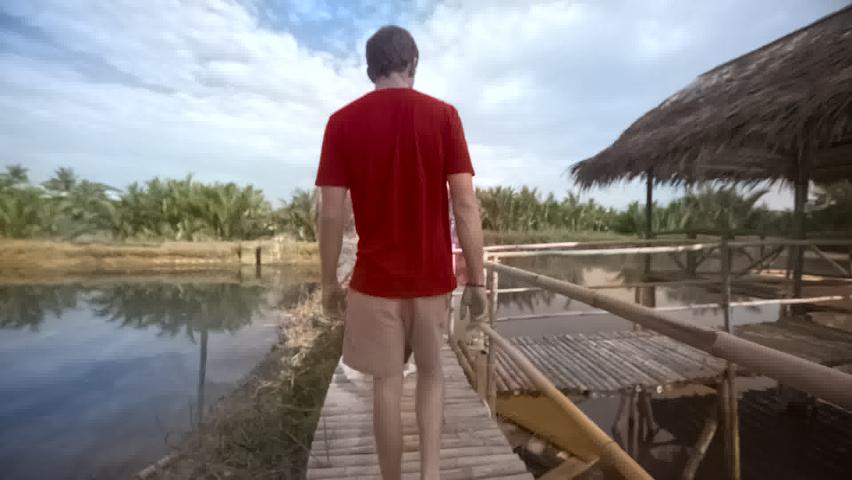}\end{subfigure} & \begin{subfigure}[b]{\myhighreswidth\textwidth}\includegraphics[width=\textwidth]{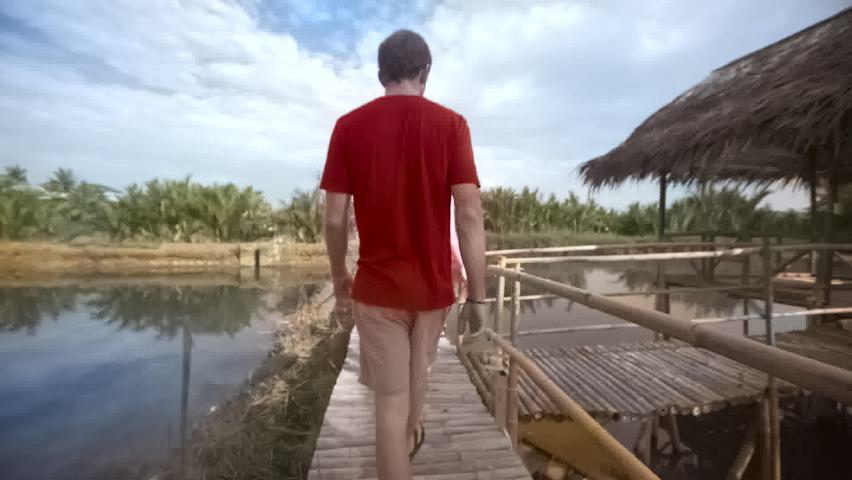}\end{subfigure}\\
\begin{subfigure}[b]{\myhighreswidth\textwidth}\includegraphics[width=\textwidth]{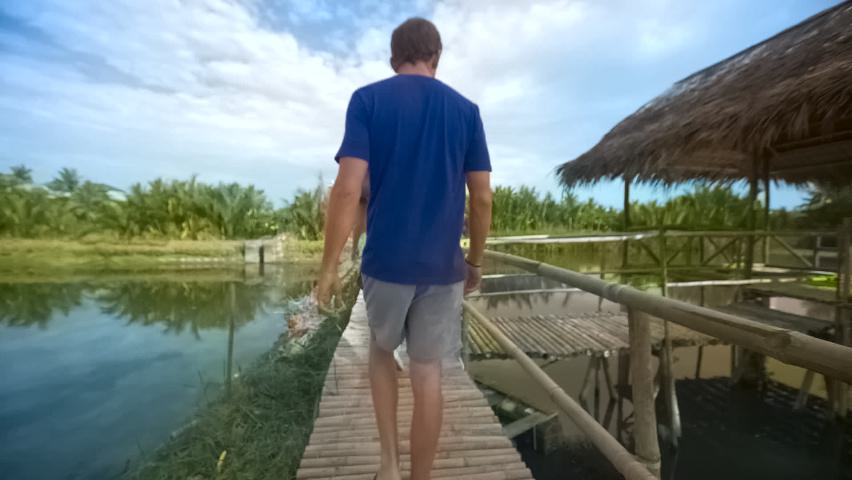}\end{subfigure} & \begin{subfigure}[b]{\myhighreswidth\textwidth}\includegraphics[width=\textwidth]{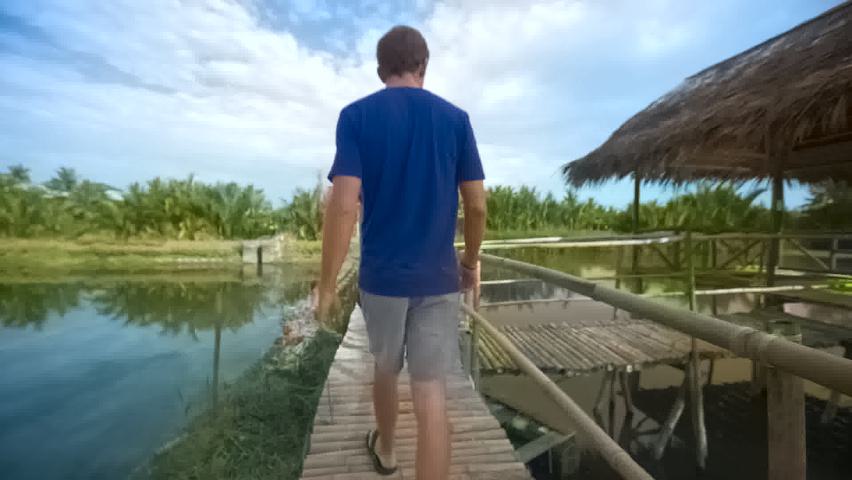}\end{subfigure} & \begin{subfigure}[b]{\myhighreswidth\textwidth}\includegraphics[width=\textwidth]{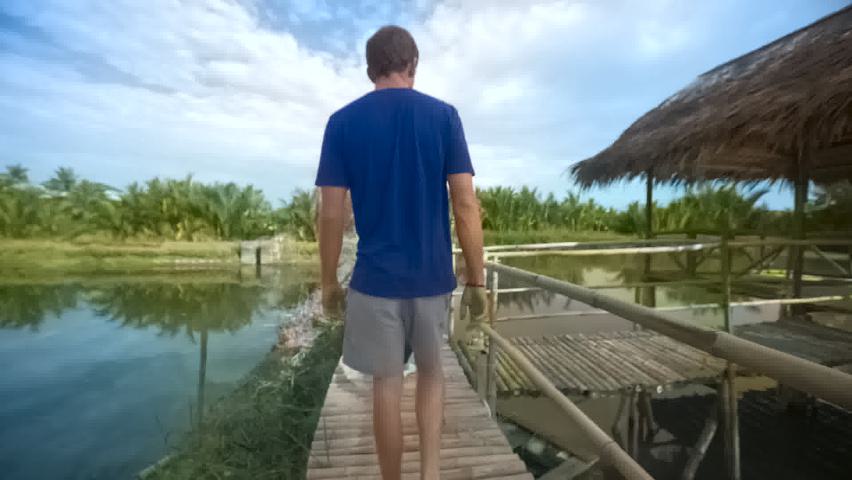}\end{subfigure} & \begin{subfigure}[b]{\myhighreswidth\textwidth}\includegraphics[width=\textwidth]{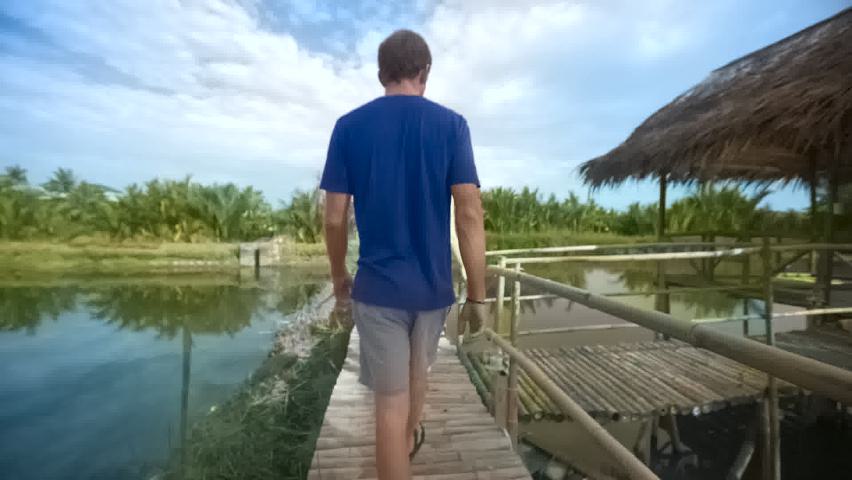}\end{subfigure}\\
\begin{subfigure}[b]{\myhighreswidth\textwidth}\includegraphics[width=\textwidth]{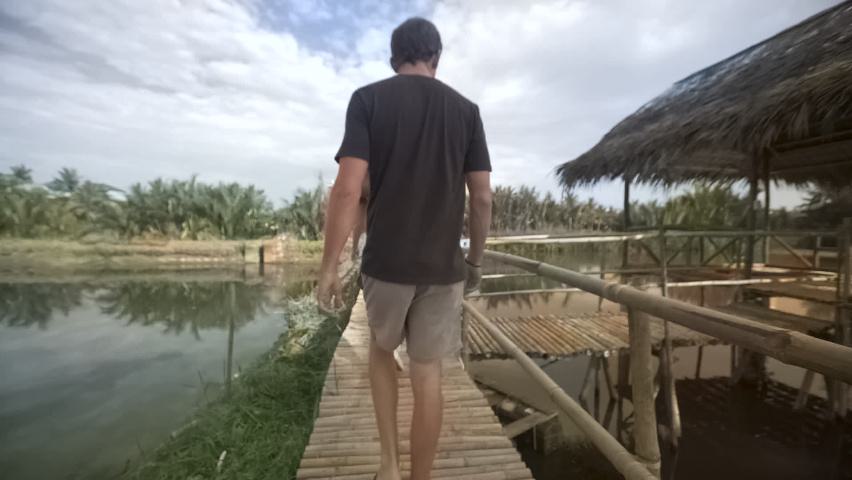}\end{subfigure} & \begin{subfigure}[b]{\myhighreswidth\textwidth}\includegraphics[width=\textwidth]{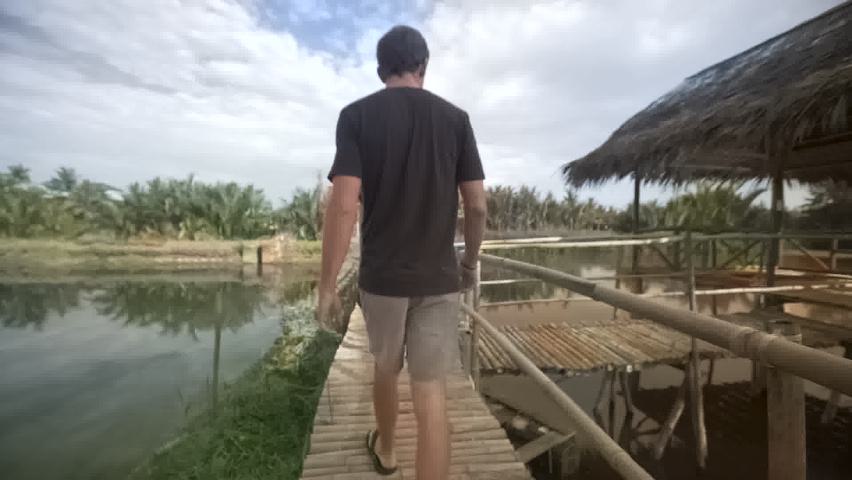}\end{subfigure} & \begin{subfigure}[b]{\myhighreswidth\textwidth}\includegraphics[width=\textwidth]{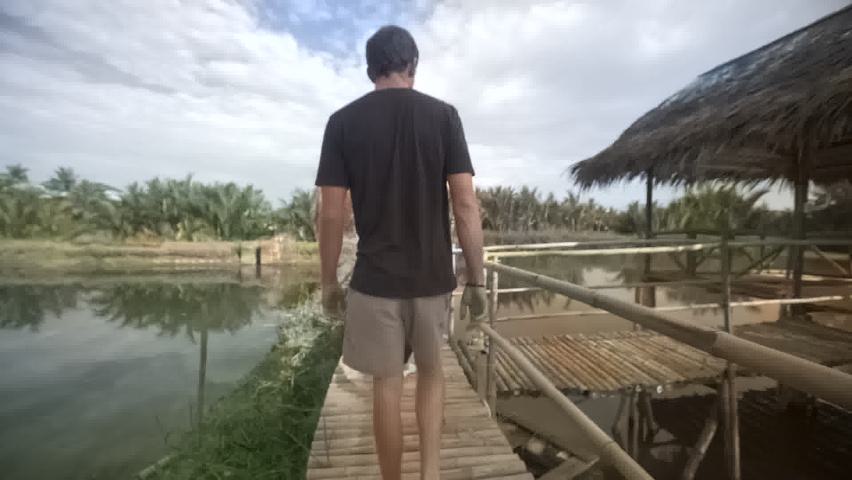}\end{subfigure} & \begin{subfigure}[b]{\myhighreswidth\textwidth}\includegraphics[width=\textwidth]{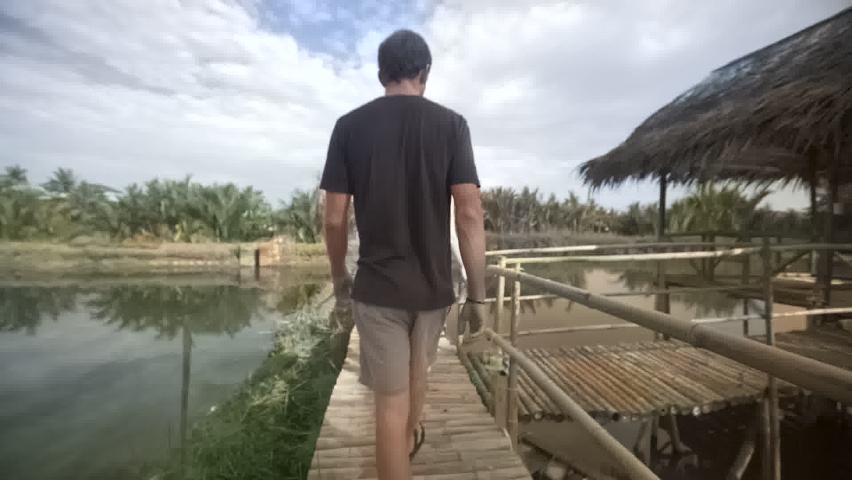}\end{subfigure}\\
&&&\\
\begin{subfigure}[b]{\myhighreswidth\textwidth}\includegraphics[width=\textwidth]{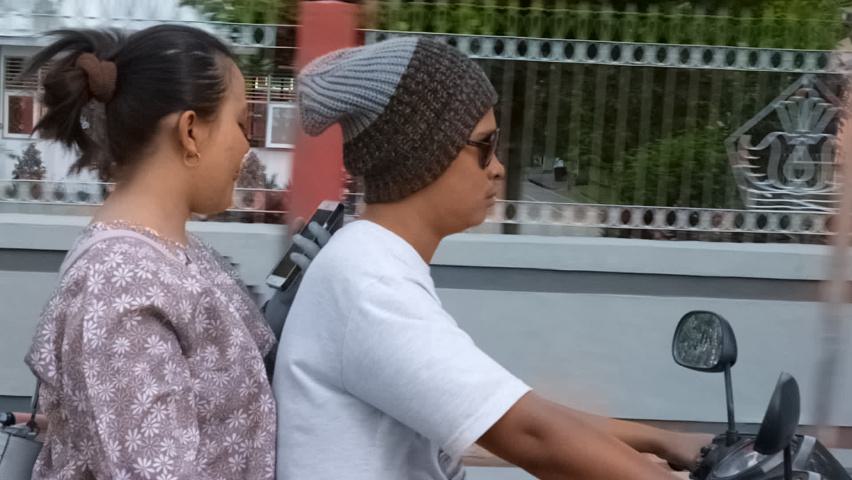}\end{subfigure} & \begin{subfigure}[b]{\myhighreswidth\textwidth}\includegraphics[width=\textwidth]{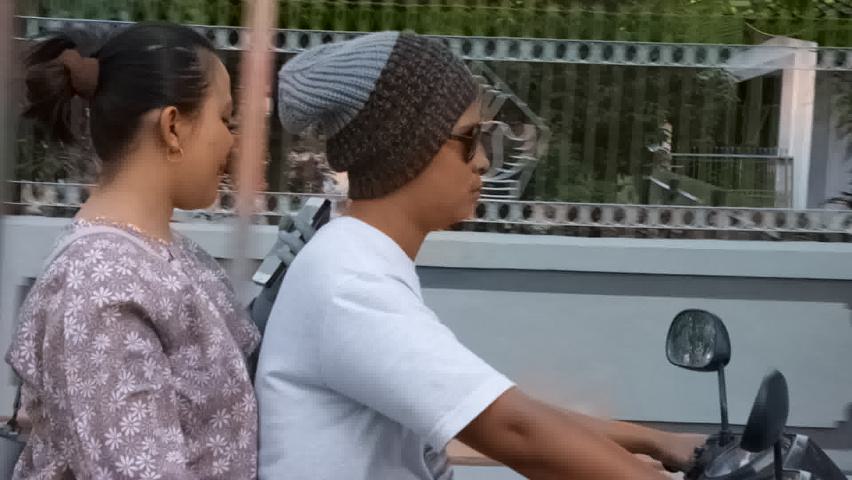}\end{subfigure} & \begin{subfigure}[b]{\myhighreswidth\textwidth}\includegraphics[width=\textwidth]{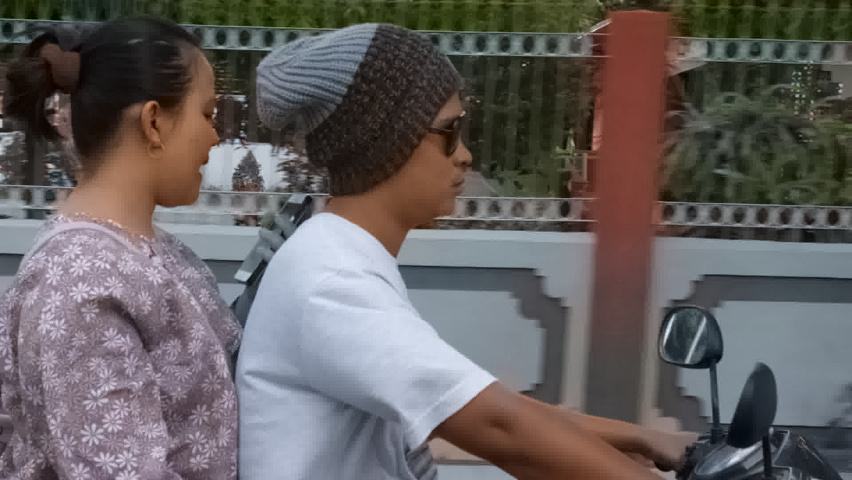}\end{subfigure} & \begin{subfigure}[b]{\myhighreswidth\textwidth}\includegraphics[width=\textwidth]{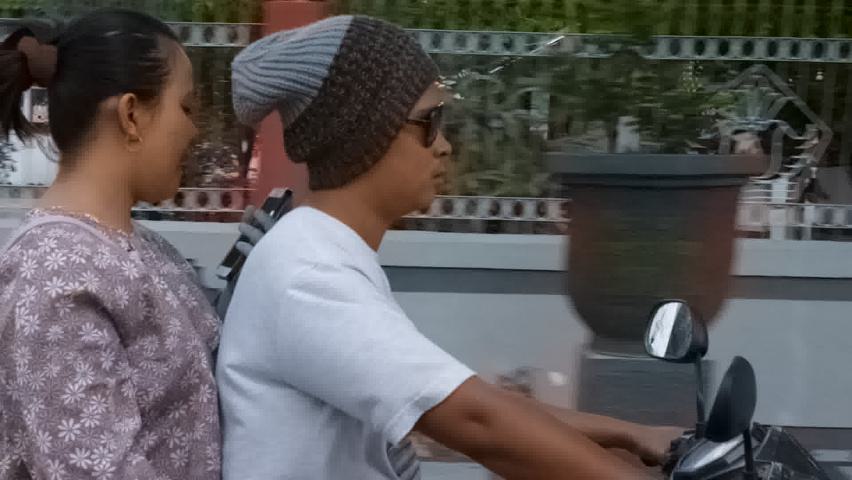}\end{subfigure}\\
\begin{subfigure}[b]{\myhighreswidth\textwidth}\includegraphics[width=\textwidth]{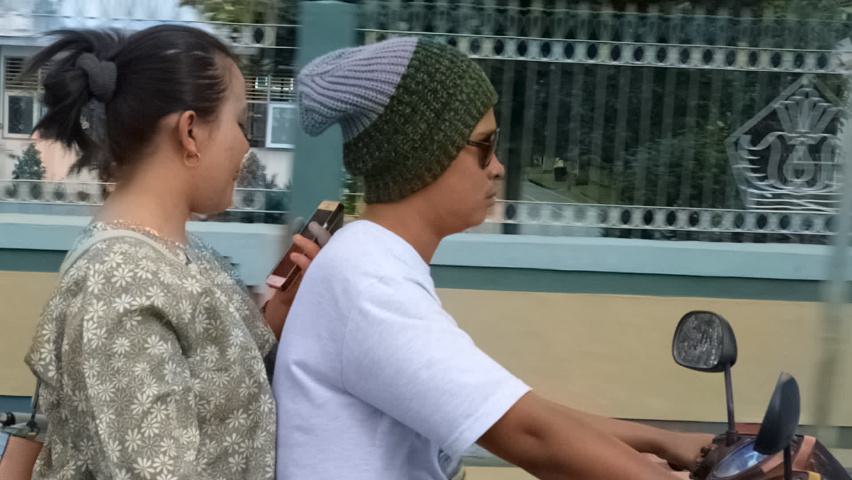}\end{subfigure} & \begin{subfigure}[b]{\myhighreswidth\textwidth}\includegraphics[width=\textwidth]{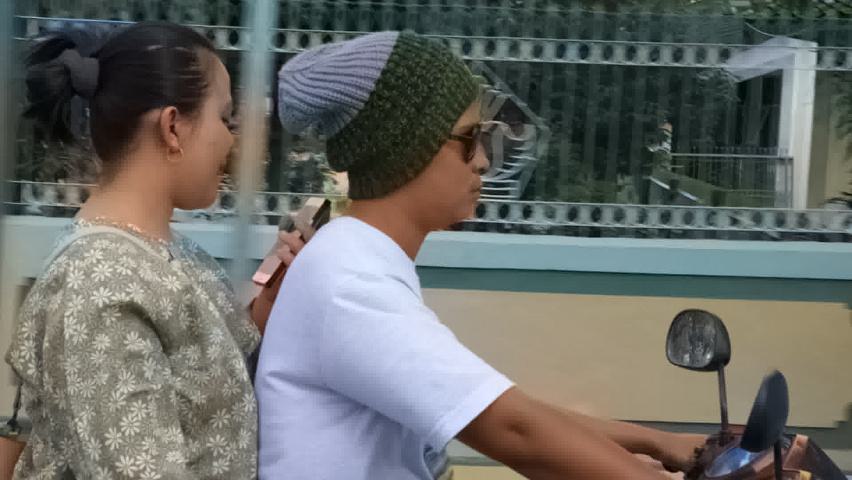}\end{subfigure} & \begin{subfigure}[b]{\myhighreswidth\textwidth}\includegraphics[width=\textwidth]{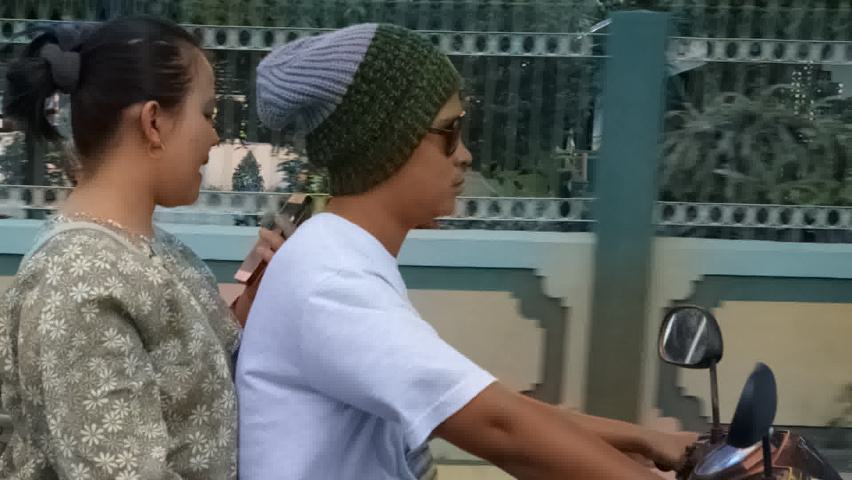}\end{subfigure} & \begin{subfigure}[b]{\myhighreswidth\textwidth}\includegraphics[width=\textwidth]{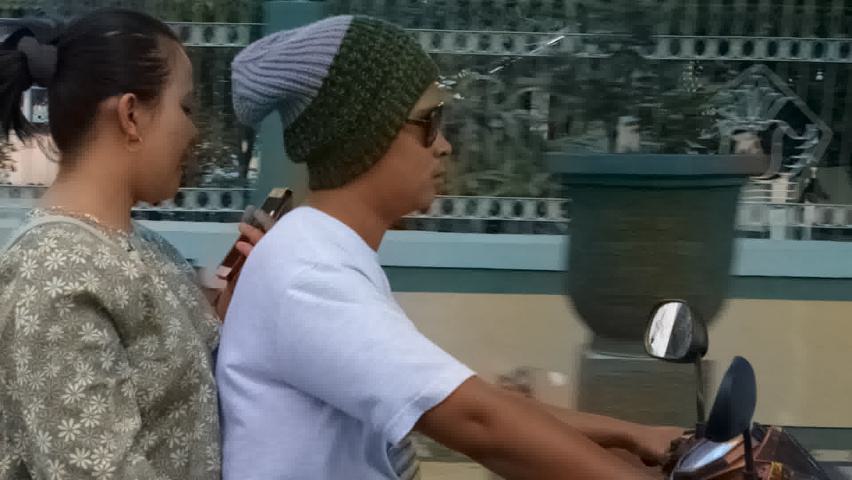}\end{subfigure}\\
\begin{subfigure}[b]{\myhighreswidth\textwidth}\includegraphics[width=\textwidth]{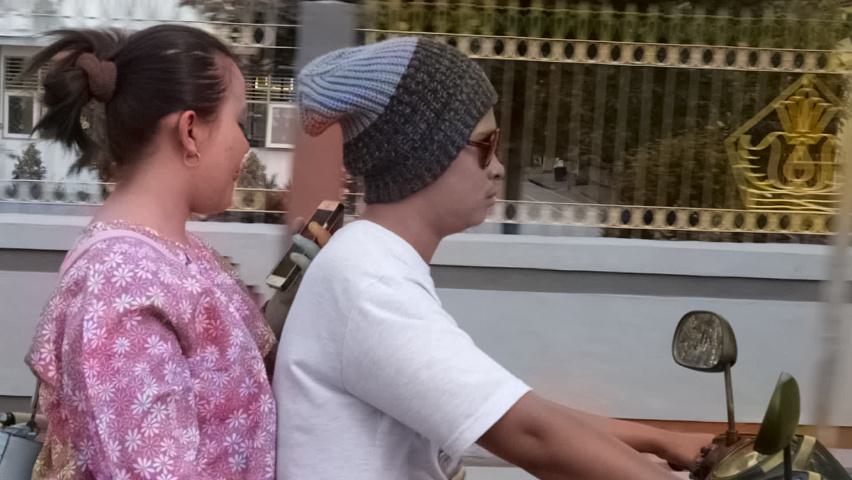}\end{subfigure} & \begin{subfigure}[b]{\myhighreswidth\textwidth}\includegraphics[width=\textwidth]{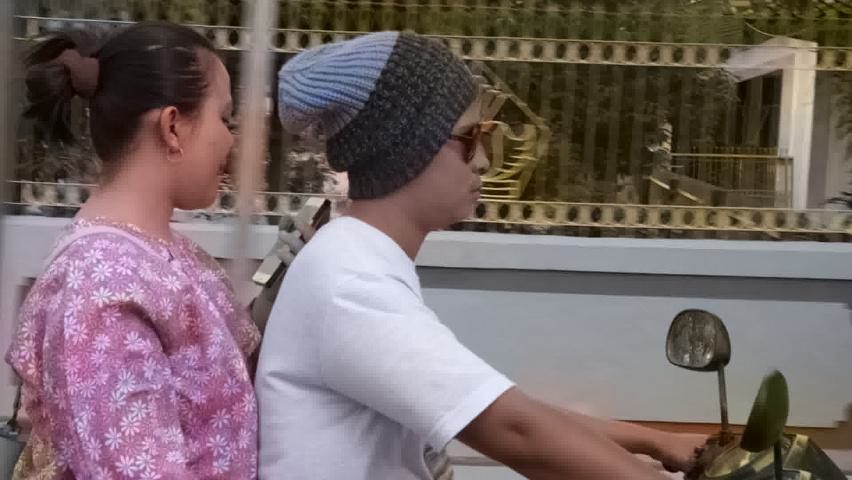}\end{subfigure} & \begin{subfigure}[b]{\myhighreswidth\textwidth}\includegraphics[width=\textwidth]{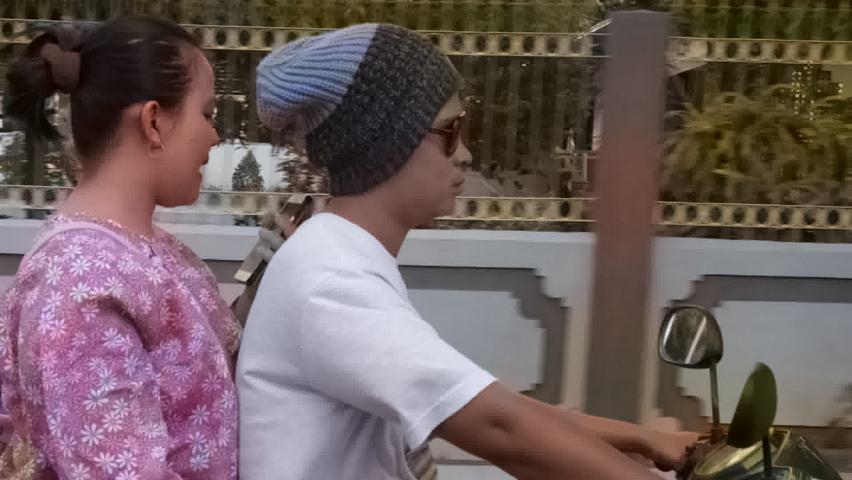}\end{subfigure} & \begin{subfigure}[b]{\myhighreswidth\textwidth}\includegraphics[width=\textwidth]{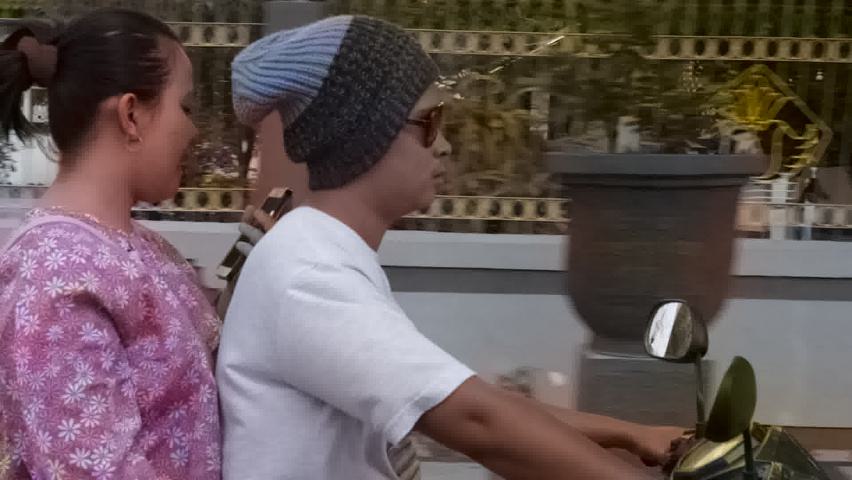}\end{subfigure}\\
&&&\\
\begin{subfigure}[b]{\myhighreswidth\textwidth}\includegraphics[width=\textwidth]{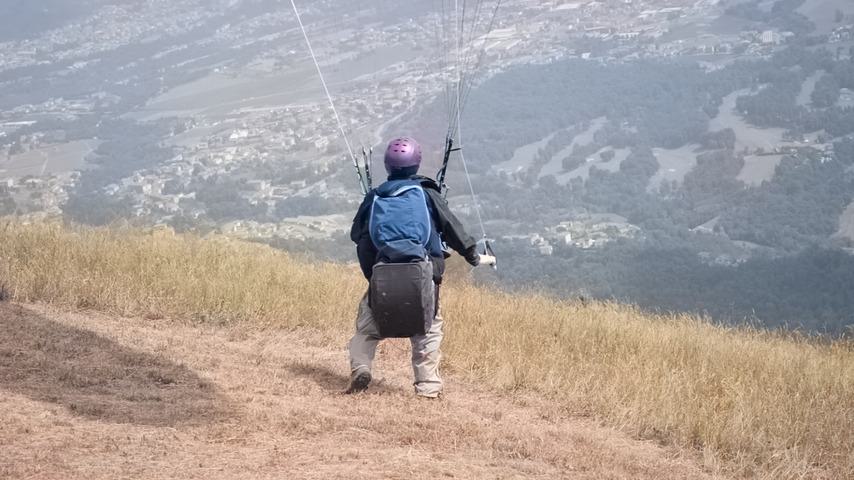}\end{subfigure} & \begin{subfigure}[b]{\myhighreswidth\textwidth}\includegraphics[width=\textwidth]{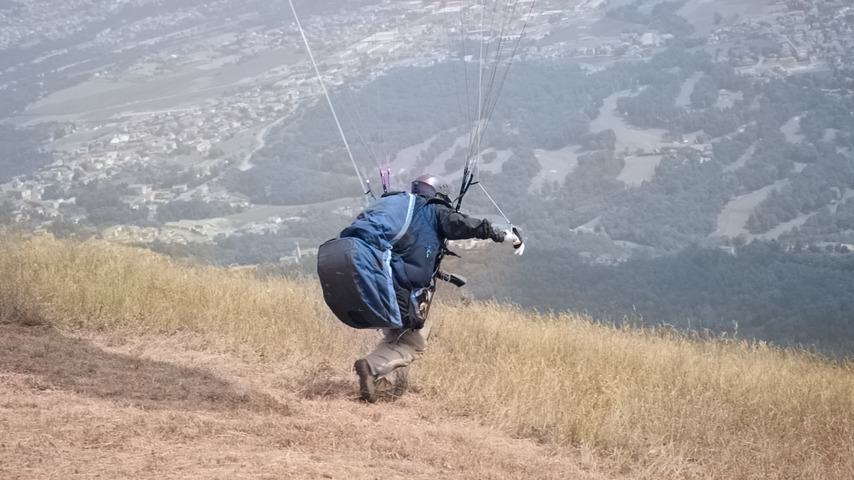}\end{subfigure} & \begin{subfigure}[b]{\myhighreswidth\textwidth}\includegraphics[width=\textwidth]{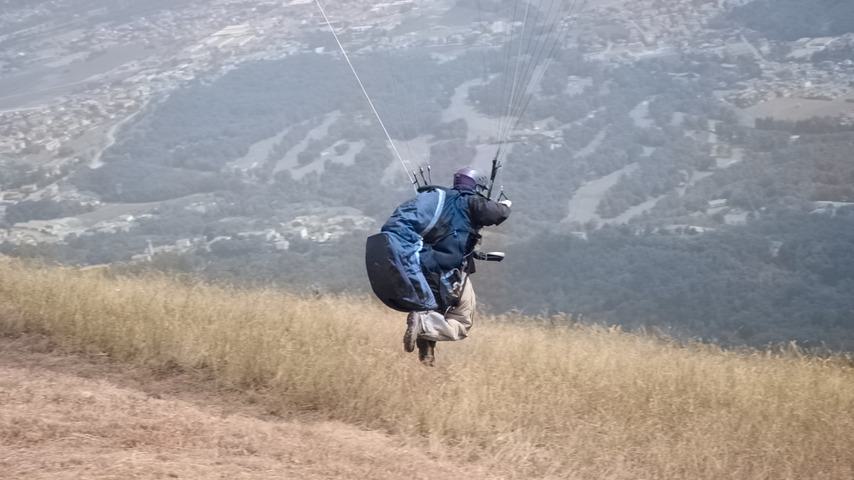}\end{subfigure} & \begin{subfigure}[b]{\myhighreswidth\textwidth}\includegraphics[width=\textwidth]{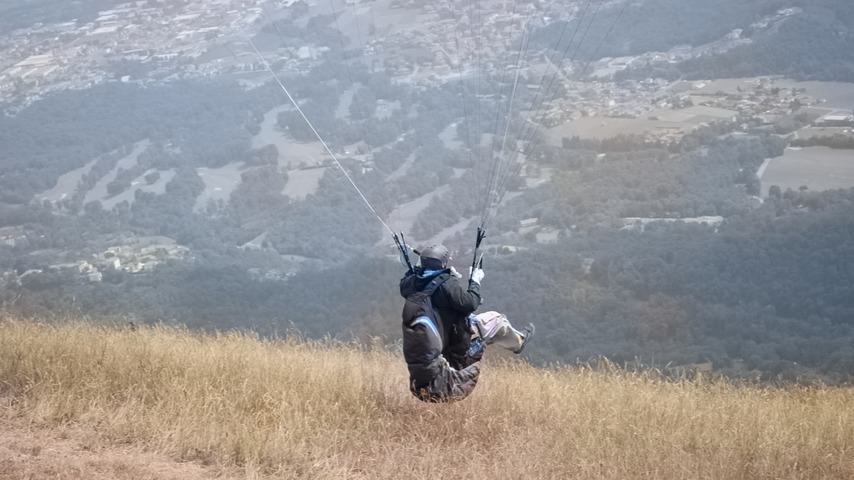}\end{subfigure}\\
\begin{subfigure}[b]{\myhighreswidth\textwidth}\includegraphics[width=\textwidth]{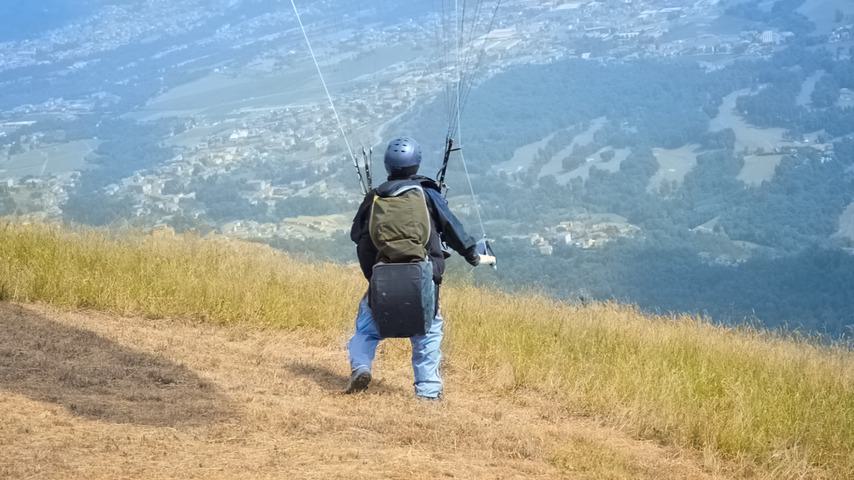}\end{subfigure} & \begin{subfigure}[b]{\myhighreswidth\textwidth}\includegraphics[width=\textwidth]{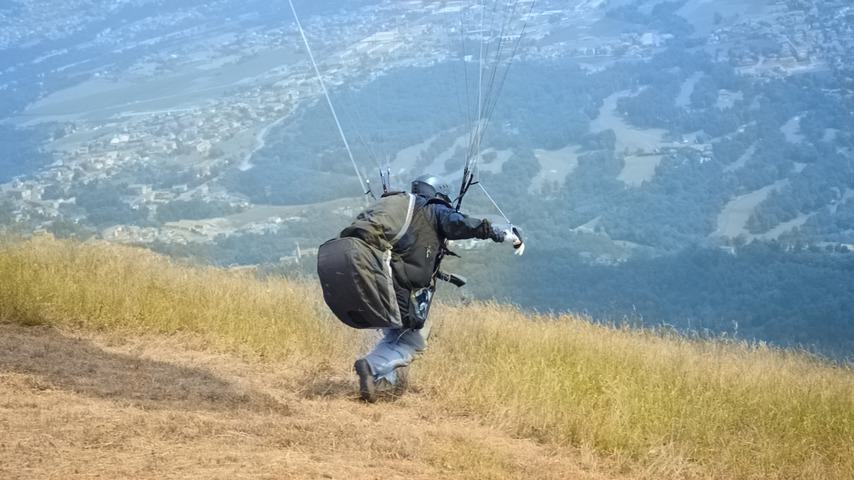}\end{subfigure} & \begin{subfigure}[b]{\myhighreswidth\textwidth}\includegraphics[width=\textwidth]{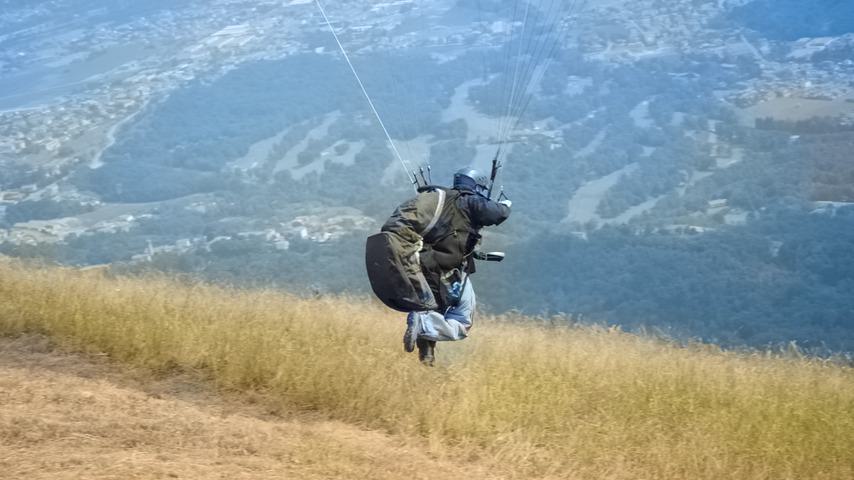}\end{subfigure} & \begin{subfigure}[b]{\myhighreswidth\textwidth}\includegraphics[width=\textwidth]{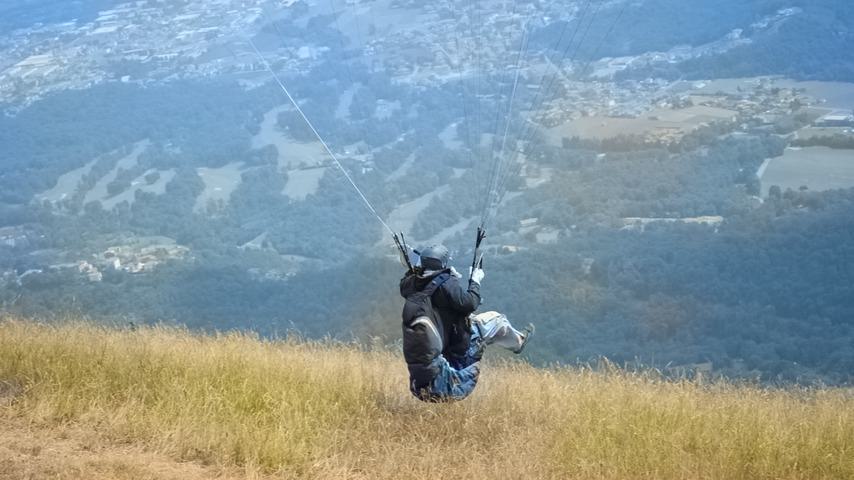}\end{subfigure}\\
\begin{subfigure}[b]{\myhighreswidth\textwidth}\includegraphics[width=\textwidth]{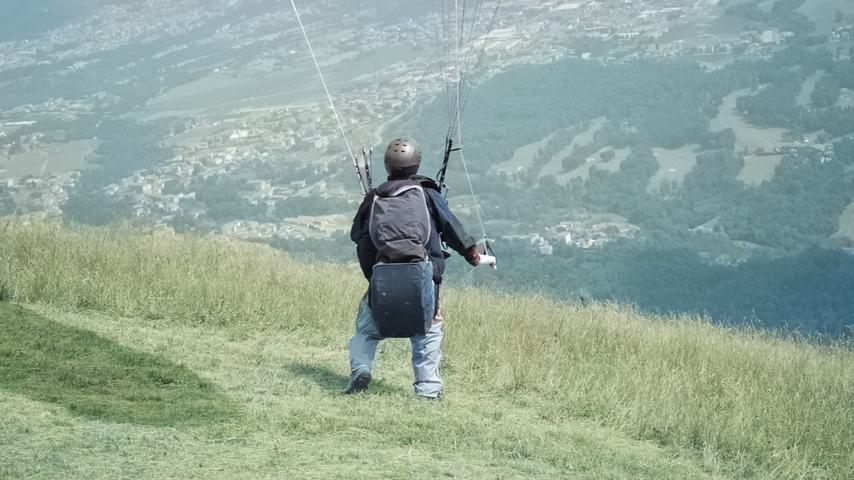}\end{subfigure} & \begin{subfigure}[b]{\myhighreswidth\textwidth}\includegraphics[width=\textwidth]{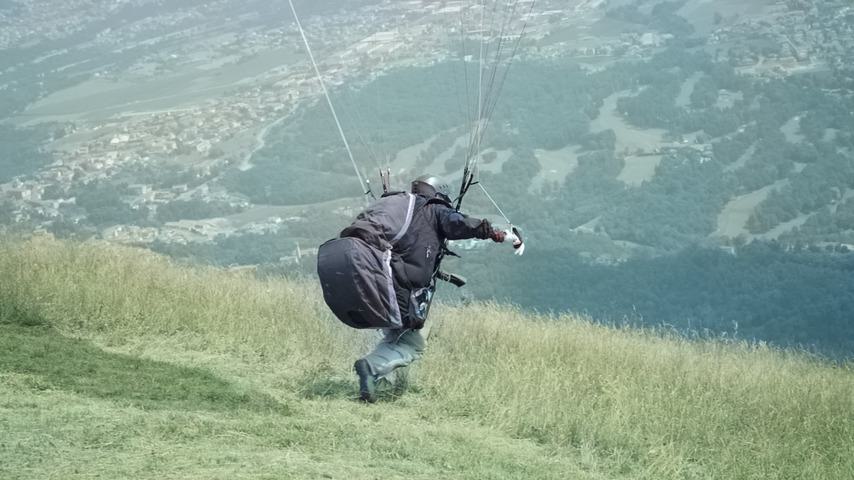}\end{subfigure} & \begin{subfigure}[b]{\myhighreswidth\textwidth}\includegraphics[width=\textwidth]{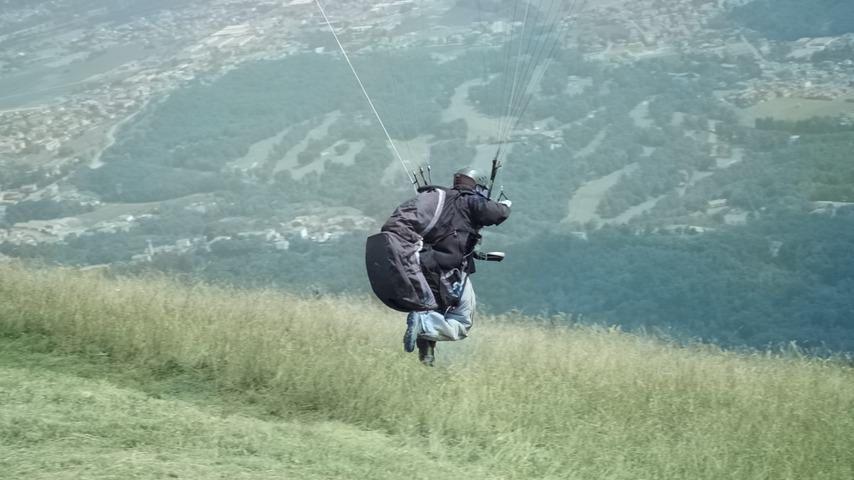}\end{subfigure} & \begin{subfigure}[b]{\myhighreswidth\textwidth}\includegraphics[width=\textwidth]{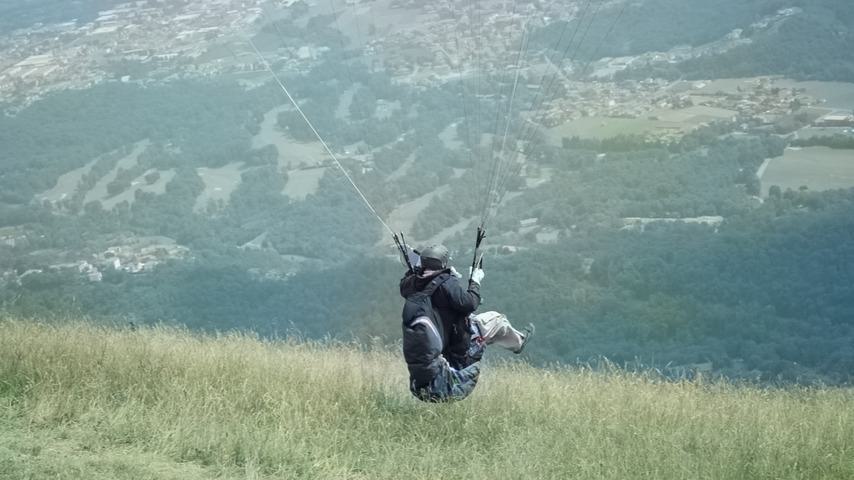}\end{subfigure}\\
&&&\\
\begin{subfigure}[b]{\myhighreswidth\textwidth}\includegraphics[width=\textwidth]{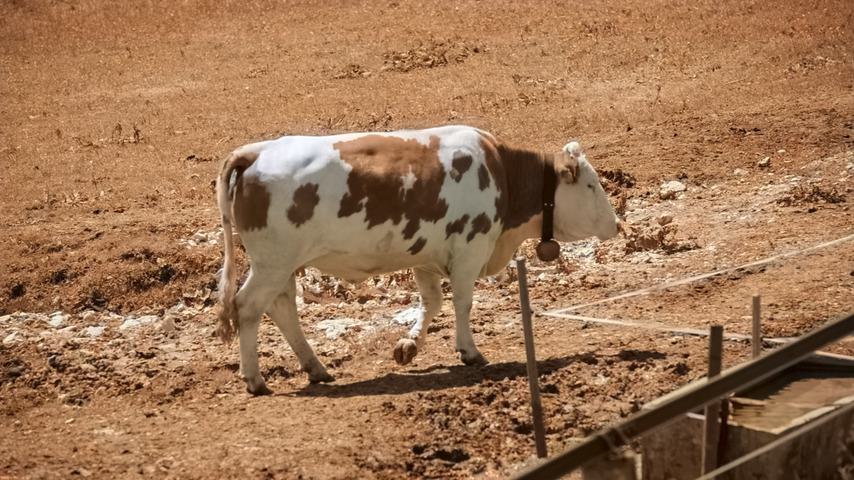}\end{subfigure} & \begin{subfigure}[b]{\myhighreswidth\textwidth}\includegraphics[width=\textwidth]{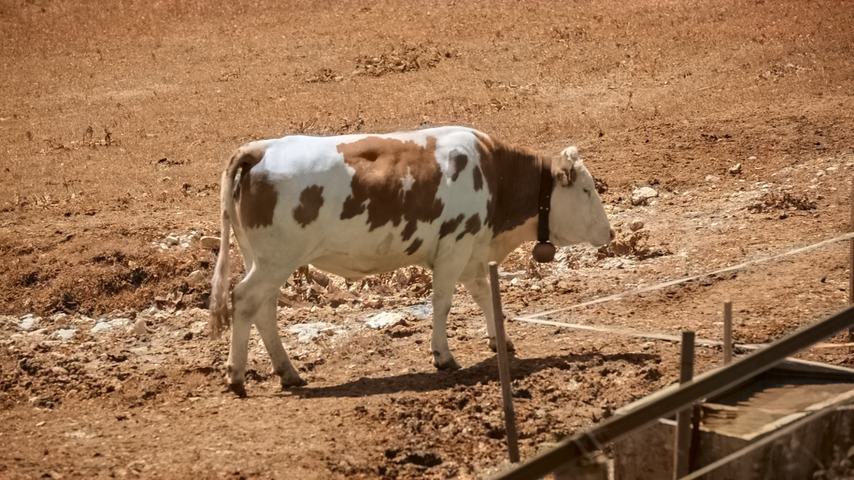}\end{subfigure} & \begin{subfigure}[b]{\myhighreswidth\textwidth}\includegraphics[width=\textwidth]{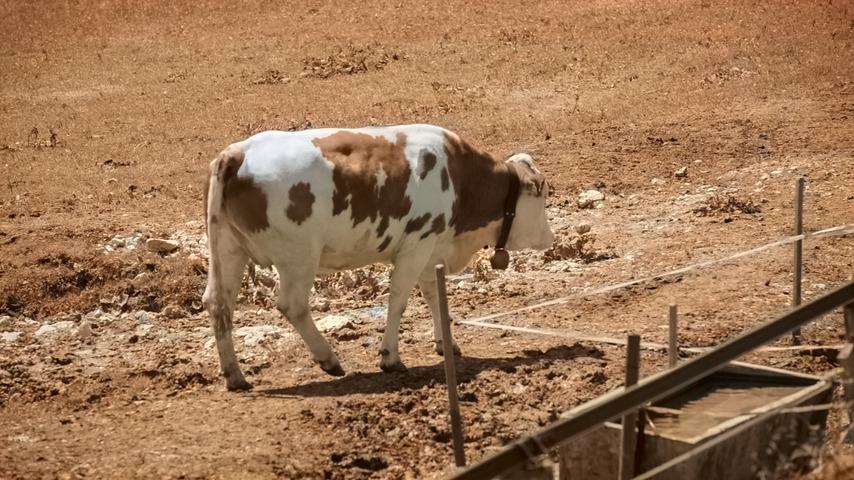}\end{subfigure} & \begin{subfigure}[b]{\myhighreswidth\textwidth}\includegraphics[width=\textwidth]{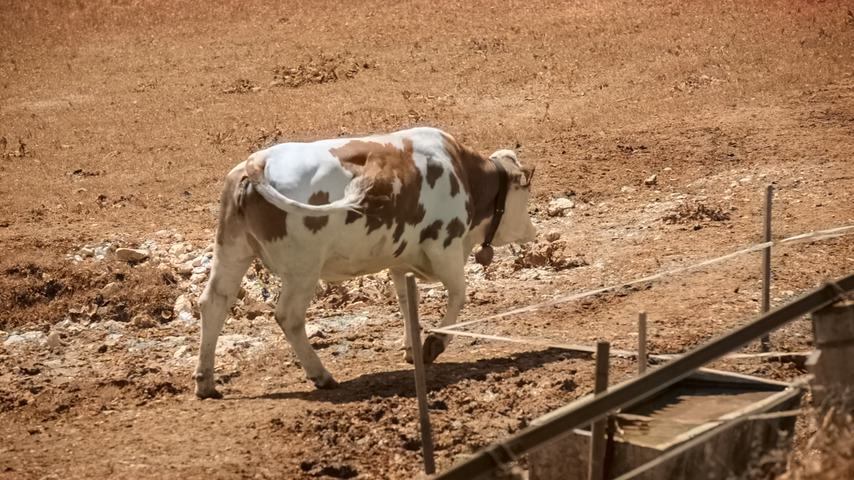}\end{subfigure}\\
\begin{subfigure}[b]{\myhighreswidth\textwidth}\includegraphics[width=\textwidth]{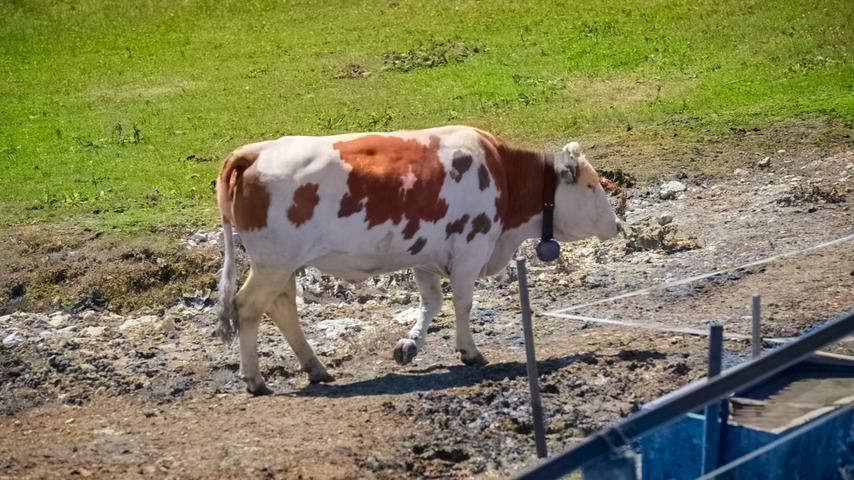}\end{subfigure} & \begin{subfigure}[b]{\myhighreswidth\textwidth}\includegraphics[width=\textwidth]{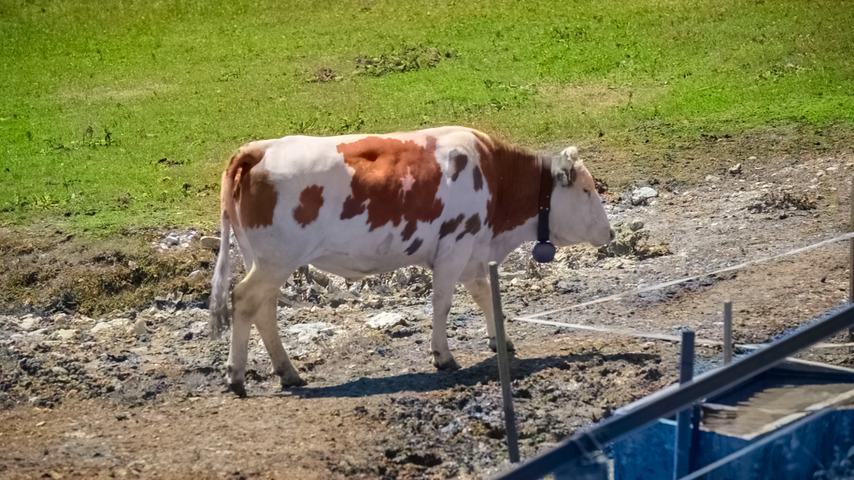}\end{subfigure} & \begin{subfigure}[b]{\myhighreswidth\textwidth}\includegraphics[width=\textwidth]{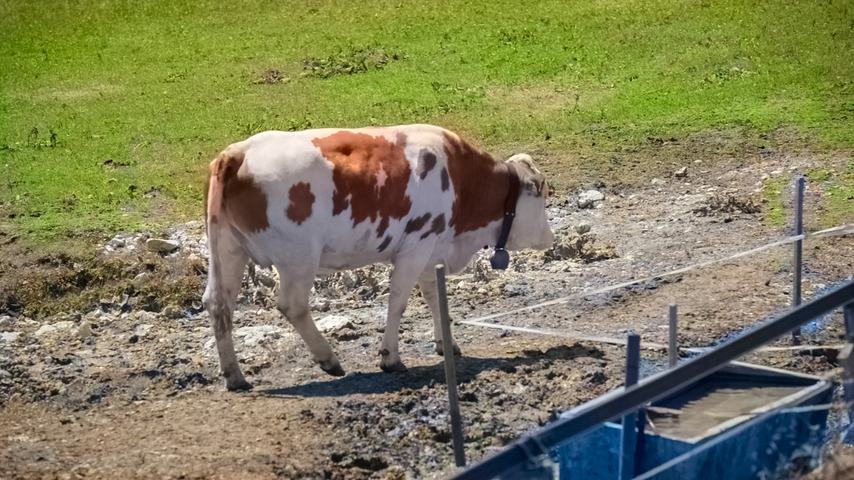}\end{subfigure} & \begin{subfigure}[b]{\myhighreswidth\textwidth}\includegraphics[width=\textwidth]{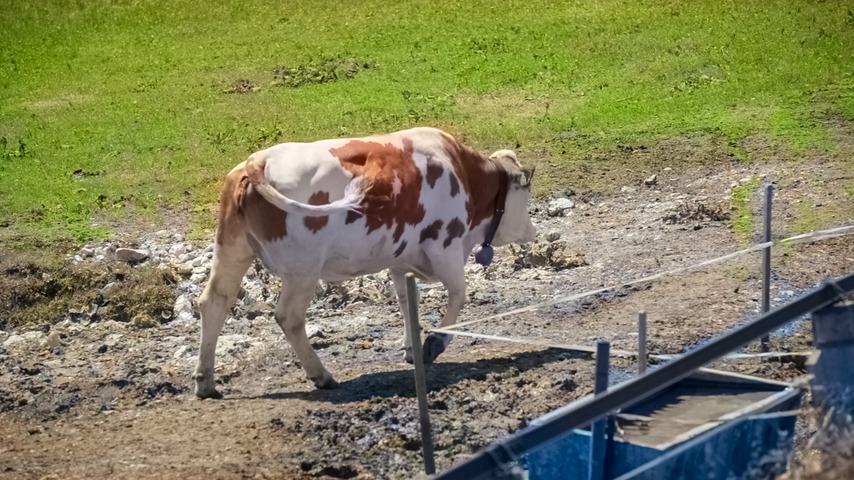}\end{subfigure}\\
\begin{subfigure}[b]{\myhighreswidth\textwidth}\includegraphics[width=\textwidth]{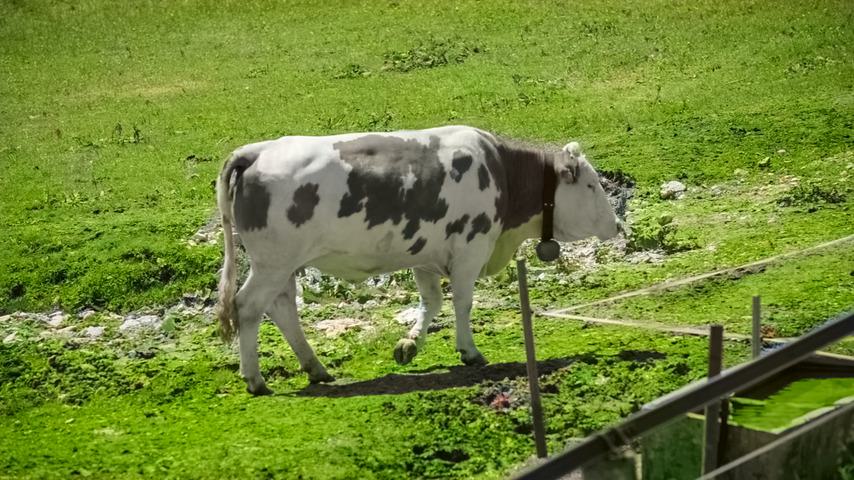}\end{subfigure} & \begin{subfigure}[b]{\myhighreswidth\textwidth}\includegraphics[width=\textwidth]{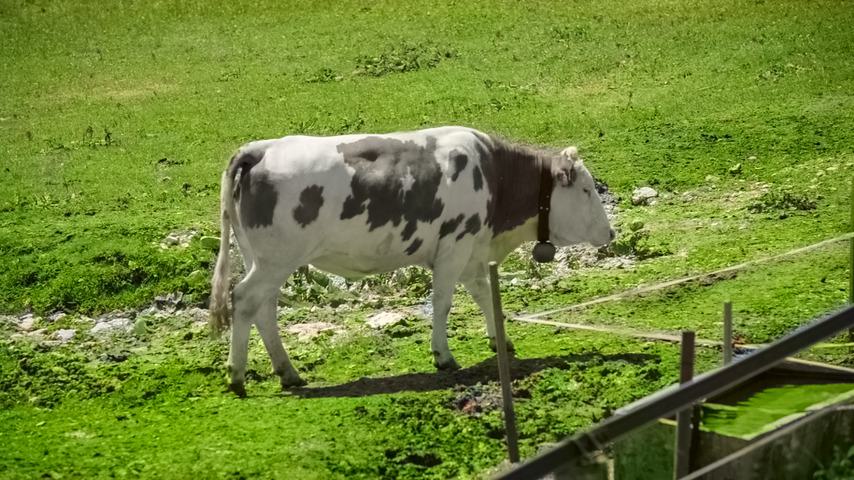}\end{subfigure} & \begin{subfigure}[b]{\myhighreswidth\textwidth}\includegraphics[width=\textwidth]{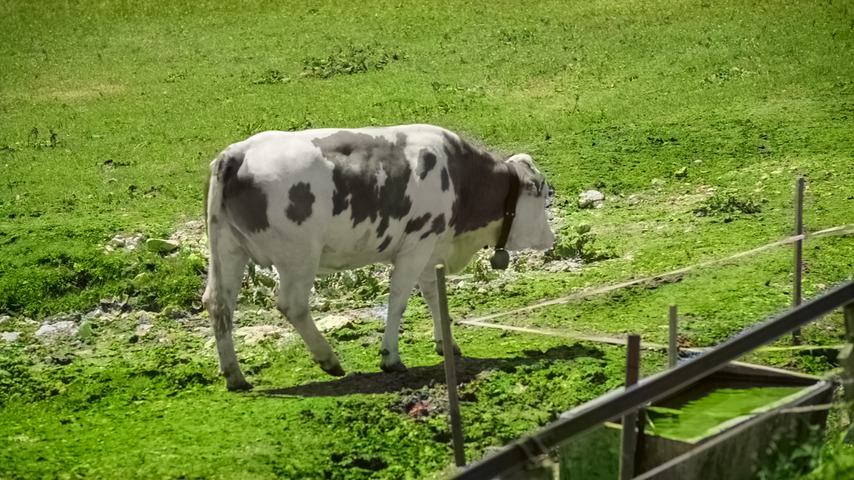}\end{subfigure} & \begin{subfigure}[b]{\myhighreswidth\textwidth}\includegraphics[width=\textwidth]{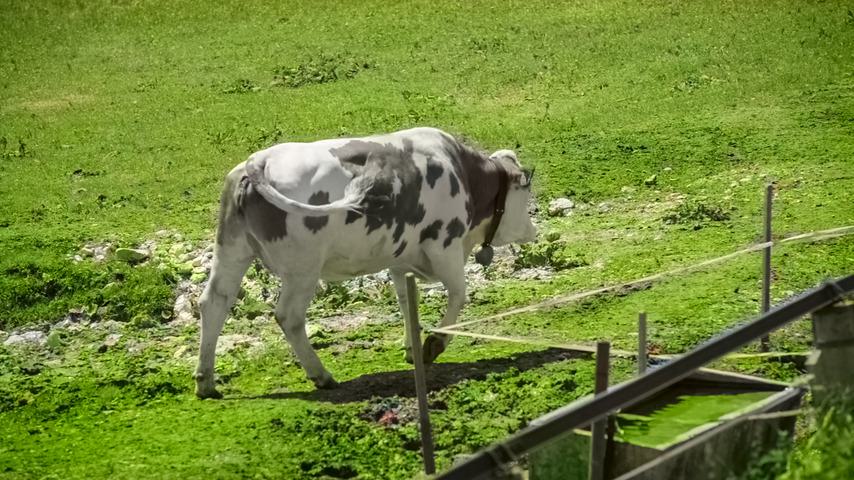}\end{subfigure}\\
&&&\\
T=0 & T=15 & T=30 & T=45 \\
\end{tabular}
\renewcommand{\arraystretch}{1.0}

\caption{Our method can generate diverse colorization results.}
\label{fig:diverse_results}

\end{figure}

\noindent\textbf{Comparisons.} We conducted a comparative analysis of our approach against two types of video colorization baselines: automatic video colorization and exemplar-based video colorization. In the case of automatic video colorization, we compared our method with four state of the art approaches: AutoColor \cite{AutomaticVideoColorization}, DeOldify \cite{DeOldify}, TCVC \cite{tcvc-color}, and VCGAN \cite{vcgan}. For the exemplar-based video colorization baselines, we compared our approach with DeepExemplar \cite{zhang2019deep} and DeepRemaster \cite{DeepRemaster}. As exemplar-based video colorization methods rely on reference images for color propagation, we utilized the state-of-the-art image colorization method UniColor \cite{unicolor} to generate the necessary references. To assess the quality of our approach, we conducted both qualitative and quantitative evaluations. The qualitative comparison is presented in Figure~\ref{fig:qualitative_eval}, while the quantitative results are reported in Table~\ref{tab:quantitative_eval}. The results demonstrate that our approach consistently achieves superior colorization across frames, producing more vivid colors compared to the baseline methods. Our proposed method has achieved state-of-the-art performance in both evaluations. Notably, we observed that providing textual descriptions can significantly enhance the colorization results in terms of perceptual quality and colorfulness. Furthermore, owing to the stochastic nature of the reverse diffusion process, our framework can natively generates diverse colorization results, as depicted in Figure~\ref{fig:diverse_results}.

\subsection{User Study}

In order to validate the effectiveness of our proposed method and to address the absence of a universally accepted evaluation standard for colorization techniques, we conducted a user study to obtain subjective assessments. This section presents the details of the user study and provides the results. 
To conduct the user study, we randomly selected 9 videos from the DAVIS validation dataset and 6 videos from the Videvo test dataset. For each video, we presented the colorization results generated by different methods in a grid, with the order of presentation randomized. The participants were then asked to rank the colorization results based on visual quality and temporal consistency respectively. Additionally, each participant was required to answer a validation question. We recruited participants with good vision and color recognition to ensure reliable evaluations. In total, 31 participants successfully completed the evaluation. The results of the user study are summarized in Table~\ref{tab:user-study}, revealing that our method was preferred by users in terms of both visual quality and temporal consistency.

\begin{table}[h]
\caption{The preferred rate of our framework against other methods in terms of visual quality (VQ) and temporal consistency (TC)}
\label{tab:user-study}
\resizebox{0.48\textwidth}{!}{
\setlength\tabcolsep{3pt}
\centering
\begin{tabular}{ccccccc}
  \toprule
  & AutoColor & Deoldify  & DeepExemplar & DeepRemaster & TCVC & VCGAN\\
  \midrule
VQ & 0.76 & 0.69 & 0.58 & 0.61 & 0.79 & 0.84 \\
TC & 0.62 & 0.81 & 0.57 & 0.69 & 0.55 & 0.68 \\
  \bottomrule
  \end{tabular}
}
\end{table}

\subsection{Ablation Study}
\begin{figure}[h]
    \centering
\def\myhighreswidth{0.09}
\def\myhighresoffset{0.00}
\centering
\renewcommand{\arraystretch}{0.3}
\setlength{\tabcolsep}{1pt}
\begin{tabular}[c]{ccccc}
        \begin{subfigure}[b]{\myhighreswidth\textwidth}\includegraphics[width=\textwidth]{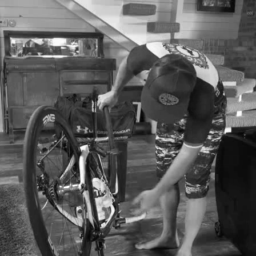}\end{subfigure}
        &
        \begin{subfigure}[b]{\myhighreswidth\textwidth}\includegraphics[width=\textwidth]{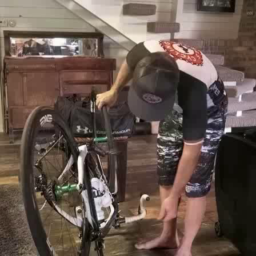}\end{subfigure}
        &
        \begin{subfigure}[b]{\myhighreswidth\textwidth}\includegraphics[width=\textwidth]{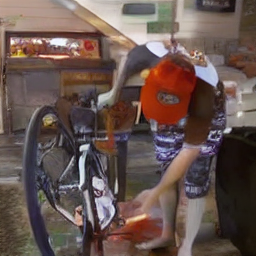}\end{subfigure}
        &
        \begin{subfigure}[b]{\myhighreswidth\textwidth}\includegraphics[width=\textwidth]{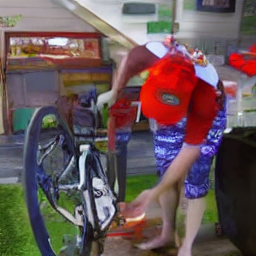}\end{subfigure}
        &
        \begin{subfigure}[b]{\myhighreswidth\textwidth}\includegraphics[width=\textwidth]{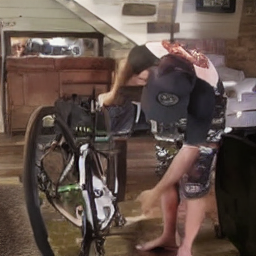}\end{subfigure}
        \\
        \begin{subfigure}[b]{\myhighreswidth\textwidth}\includegraphics[width=\textwidth]{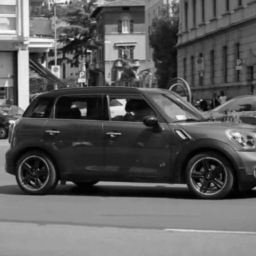}\end{subfigure}
        &
        \begin{subfigure}[b]{\myhighreswidth\textwidth}\includegraphics[width=\textwidth]{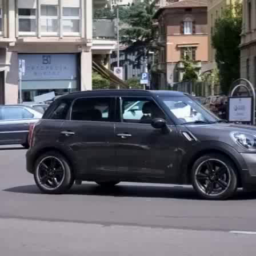}\end{subfigure}
        &
        \begin{subfigure}[b]{\myhighreswidth\textwidth}\includegraphics[width=\textwidth]{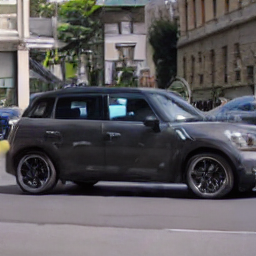}\end{subfigure}
        &
        \begin{subfigure}[b]{\myhighreswidth\textwidth}\includegraphics[width=\textwidth]{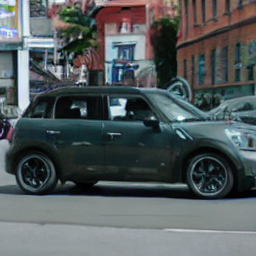}\end{subfigure}
        &
        \begin{subfigure}[b]{\myhighreswidth\textwidth}\includegraphics[width=\textwidth]{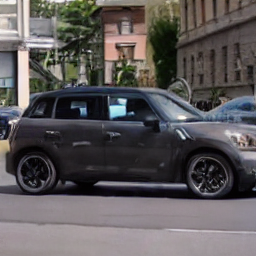}\end{subfigure}
        \\
        (a) & (b) & (c) & (d) & (e)
\end{tabular}
\renewcommand{\arraystretch}{1.0}
\caption{The effect of color propagation attention: (a) input frame; (b) reference frame; (c) w/o color propagation attention; (d) w/ UNet self-attention; (d) w/ color propagation attention}
\label{fig:ablation-attention}
\end{figure}
\begin{figure}[h]
    \centering
\def\myhighreswidth{0.09}
\def\myhighresoffset{0.00}
\centering
\renewcommand{\arraystretch}{1.0}
\setlength{\tabcolsep}{0pt}
\begin{tabular}[c]{cccccc}
        (a) &
        \begin{subfigure}[c]{\myhighreswidth\textwidth}\includegraphics[width=\textwidth]{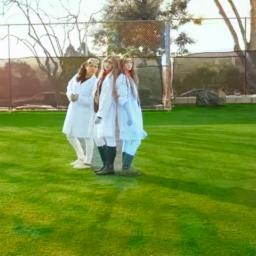}\end{subfigure}
        &
        \begin{subfigure}[c]{\myhighreswidth\textwidth}\includegraphics[width=\textwidth]{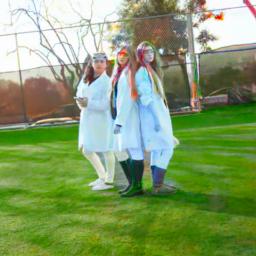}\end{subfigure}
        &
        \begin{subfigure}[c]{\myhighreswidth\textwidth}\includegraphics[width=\textwidth]{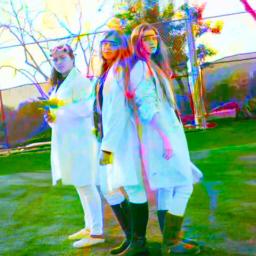}\end{subfigure}
        &
        \begin{subfigure}[c]{\myhighreswidth\textwidth}\includegraphics[width=\textwidth]{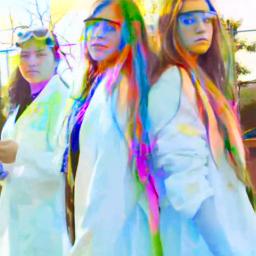}\end{subfigure}
        &
        \begin{subfigure}[c]{\myhighreswidth\textwidth}\includegraphics[width=\textwidth]{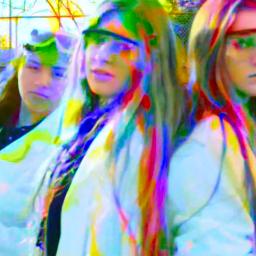}\end{subfigure}
        \\
        (b) &
        \begin{subfigure}[c]{\myhighreswidth\textwidth}\includegraphics[width=\textwidth]{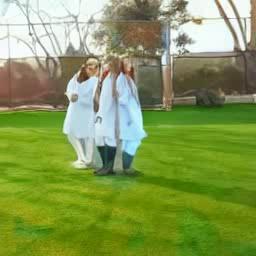}\end{subfigure}
        &
        \begin{subfigure}[c]{\myhighreswidth\textwidth}\includegraphics[width=\textwidth]{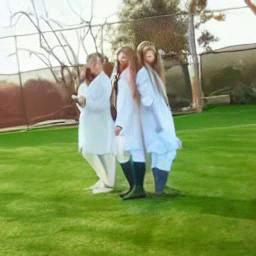}\end{subfigure}
        &
        \begin{subfigure}[c]{\myhighreswidth\textwidth}\includegraphics[width=\textwidth]{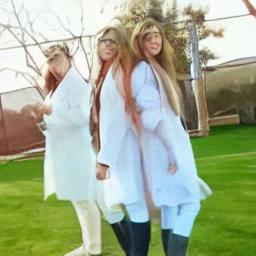}\end{subfigure}
        &
        \begin{subfigure}[c]{\myhighreswidth\textwidth}\includegraphics[width=\textwidth]{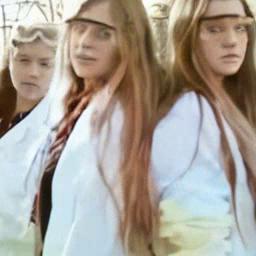}\end{subfigure}
        &
        \begin{subfigure}[c]{\myhighreswidth\textwidth}\includegraphics[width=\textwidth]{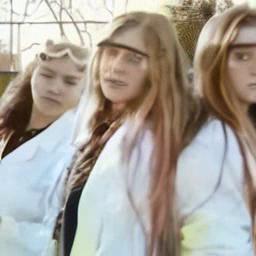}\end{subfigure}
        \\
        (c) &
        \begin{subfigure}[c]{\myhighreswidth\textwidth}\includegraphics[width=\textwidth]{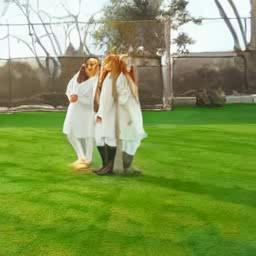}\end{subfigure}
        &
        \begin{subfigure}[c]{\myhighreswidth\textwidth}\includegraphics[width=\textwidth]{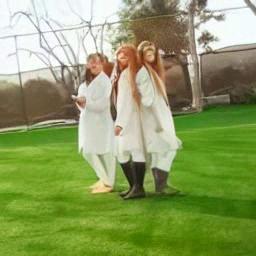}\end{subfigure}
        &
        \begin{subfigure}[c]{\myhighreswidth\textwidth}\includegraphics[width=\textwidth]{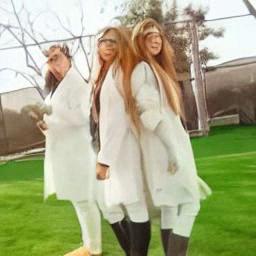}\end{subfigure}
        &
        \begin{subfigure}[c]{\myhighreswidth\textwidth}\includegraphics[width=\textwidth]{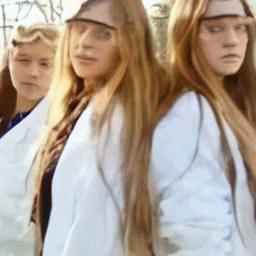}\end{subfigure}
        &
        \begin{subfigure}[c]{\myhighreswidth\textwidth}\includegraphics[width=\textwidth]{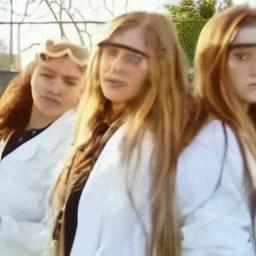}\end{subfigure}
        \\
        & T=0 & T=10 & T=20 & T=30 & T=40
        \\
\end{tabular}
\renewcommand{\arraystretch}{1.0}

\caption{The effect of alternated sampling strategy: (a) autoregressive sampling; (b) unidirectional color information propagation; (c) bidirectional color information propagation}
\label{fig:ablation-sampling}
\end{figure}

\noindent\textbf{The effect of color propagation attention and alternated sampling strategy.}
In our framework, the combination of Color Propagation Attention and Alternated Sampling Strategy allows us to transform a sequence of image diffusion processes into a coordinated video diffusion process. In this ablation study, we investigate the impact of color propagation attention and the alternated sampling strategy. We specifically train the following variants of our framework: a) accepting reference frame latent without color propagation attention; b) accepting reference frame latent via U-Net self-attention. Figure~\ref{fig:ablation-attention} demonstrates the effects of these variants. It can be observed that the color propagation attention is capable of generating consistent colorization based on a reference frame, while the other variants fail to do so. It is important to note that, for visual comparison purposes, we employ the original VQVAE decoder, excluding the influence of our proposed video colorization decoder. Furthermore, we compare the Alternated Sampling Strategy with the autoregressive generation strategy and the unidirectional sampling strategy. Figure~\ref{fig:ablation-sampling} presents the visual comparisons among these strategies. The alternated sampling strategy produces better results than others. 

\begin{figure}[h]
    \centering
\def\myhighreswidth{0.09}
\def\myhighresoffset{0.00}
\centering
\renewcommand{\arraystretch}{1.0}
\setlength{\tabcolsep}{0pt}
\begin{tabular}[c]{cccccc}
        (a) &
        \begin{subfigure}[c]{\myhighreswidth\textwidth}\includegraphics[width=\textwidth]{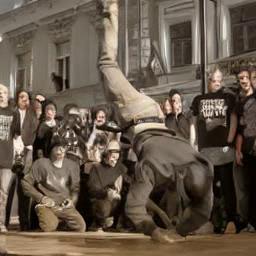}\end{subfigure}
        &
        \begin{subfigure}[c]{\myhighreswidth\textwidth}\includegraphics[width=\textwidth]{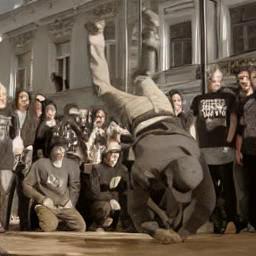}\end{subfigure}
        &
        \begin{subfigure}[c]{\myhighreswidth\textwidth}\includegraphics[width=\textwidth]{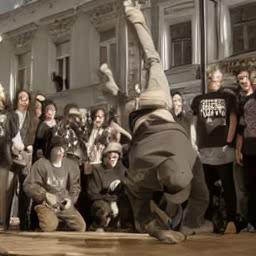}\end{subfigure}
        &
        \begin{subfigure}[c]{\myhighreswidth\textwidth}\includegraphics[width=\textwidth]{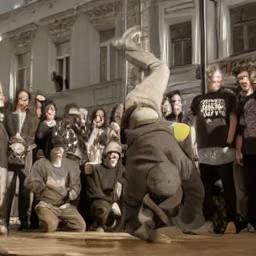}\end{subfigure}
        &
        \begin{subfigure}[c]{\myhighreswidth\textwidth}\includegraphics[width=\textwidth]{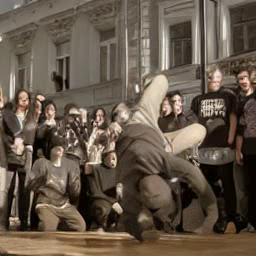}\end{subfigure}
        \\
        (b) &
        \begin{subfigure}[c]{\myhighreswidth\textwidth}\includegraphics[width=\textwidth]{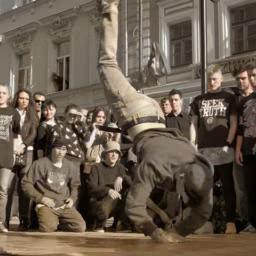}\end{subfigure}
        &
        \begin{subfigure}[c]{\myhighreswidth\textwidth}\includegraphics[width=\textwidth]{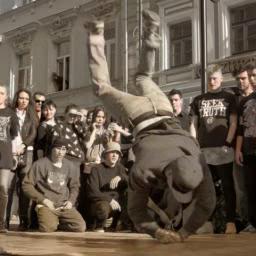}\end{subfigure}
        &
        \begin{subfigure}[c]{\myhighreswidth\textwidth}\includegraphics[width=\textwidth]{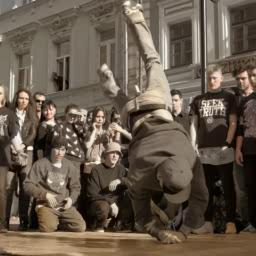}\end{subfigure}
        &
        \begin{subfigure}[c]{\myhighreswidth\textwidth}\includegraphics[width=\textwidth]{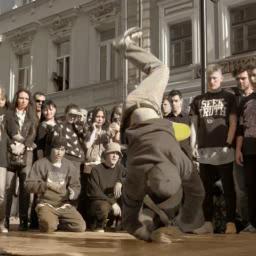}\end{subfigure}
        &
        \begin{subfigure}[c]{\myhighreswidth\textwidth}\includegraphics[width=\textwidth]{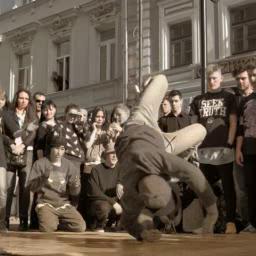}\end{subfigure}
        \\
        (c) &
        \begin{subfigure}[c]{\myhighreswidth\textwidth}\includegraphics[width=\textwidth]{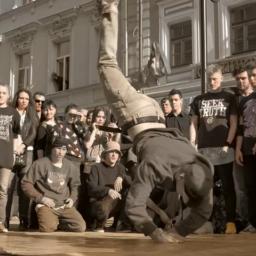}\end{subfigure}
        &
        \begin{subfigure}[c]{\myhighreswidth\textwidth}\includegraphics[width=\textwidth]{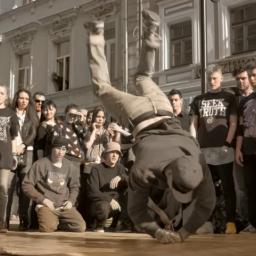}\end{subfigure}
        &
        \begin{subfigure}[c]{\myhighreswidth\textwidth}\includegraphics[width=\textwidth]{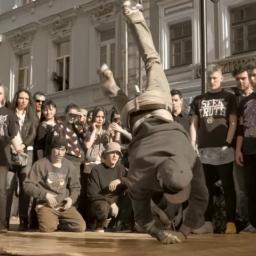}\end{subfigure}
        &
        \begin{subfigure}[c]{\myhighreswidth\textwidth}\includegraphics[width=\textwidth]{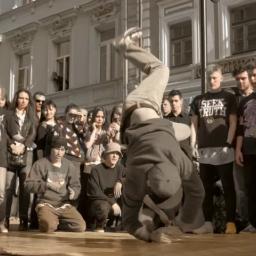}\end{subfigure}
        &
        \begin{subfigure}[c]{\myhighreswidth\textwidth}\includegraphics[width=\textwidth]{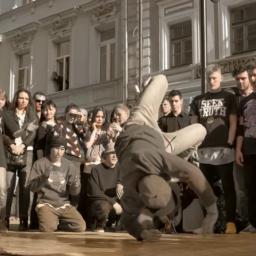}\end{subfigure}
        \\
        & T=0 & T=2 & T=4 & T=8 & T=10
        \\
\end{tabular}
\renewcommand{\arraystretch}{1.0}
\caption{The effect of \ColorizationDecoder: (a) original VQVAE decoder (w/ ograyscale feature and temporal convolutional layers); (b) ours (w/o temporal convolutional layers); (c) ours }
\label{fig:vae-out}
\end{figure}

\noindent\textbf{The \ColorizationDecoder.}
To verify the effect of our \ColorizationDecoder~ in reconstructing videos with grayscale features and temporal coherence, we compare it with the original VQVAE decoder used in Stable Diffusion. The results of this comparison are presented in Table~\ref{tab:ablation_decoder}, indicating that our \ColorizationDecoder~achieves the highest performance when benefiting from grayscale features and pseudo-3D enhancement Furthermore, Figure~\ref{fig:vae-out} provides a visual comparison of the decoder outputs, supporting our findings. It is evident from the visual results that our \ColorizationDecoder~is capable of producing the sharpest and most visually appealing results.

\begin{table}[h]
\caption{Quantitative comparisons of reconstruction quality by different decoders on our evalutation datasets}
\label{tab:ablation_decoder}
\centering
\resizebox{0.5\textwidth}{!}{
\begin{tabular}{c|ccc|ccc}
  \toprule
  \multirow{2}*{Model}& \multicolumn{3}{c|}{DAVIS} & \multicolumn{3}{c}{Videvo}\\
  & PSNR $\uparrow$ & SSIM $\uparrow$ & LPIPS $\downarrow$&PSNR $\uparrow$ & SSIM $\uparrow$ & LPIPS $\downarrow$\\
  \midrule
  \texttt{kl-f8}  & 24.29 & 0.6996 & 0.1678 & 26.10 & 0.7734 & 0.1539 \\
  Ours            & 35.74 & 0.9769 & 0.0356 & 37.02 & 0.9790 & 0.0354 \\
  \bottomrule
  \end{tabular}
  }
\end{table}



\subsection{Limitations and Discussions}

\noindent\textbf{Runtime performance.} The diffusion model utilized in our framework requires an time-consuming iterative sampling process, which may take minutes to complete on CPU devices (around 100 seconds per frame on Intel Xeon 6326 with 50 sampling steps and a spatial resolution $256\times256$). Hence, it is imperative that users should utilize GPUs when running our of pipeline (around 6 seconds per frame on NVIDIA RTX 3090 with 50 sampling steps with a spatial resolution of $256\times256$). Besides, the alternated sampling strategy requires a certain number of sampling step to ensure temporal coherence and color information propagation. Consequently, efficient diffusion samplers or diffusion model distillation techniques are incompatible with our framework.

\noindent\textbf{Potential biases.} We have leveraged a pre-trained model based on miniSD \cite{mini-sd} (a fine-tuned version of Stable Diffusion v1.4 \cite{latent-diffusion}), which was trained on the LAION dataset \cite{laion-dataset}, a dataset known to have social and cultural biases. It remains unclear how these biases can manifest themselves in the color space and potentially influence the outputs of our framework.

\section{Conclusion}
In this paper, we propose an effective video colorization pipeline based on the pre-trained T2I stable diffusion models for realistic and diverse video colorization. As evidenced by extensive evaluation, our method achieves state-of-the-art performance for video colorization.

\clearpage
{\small
\bibliographystyle{ieee_fullname}
\bibliography{bibliography}
}

\end{document}